\documentclass{article}

\usepackage{microtype}
\usepackage{graphicx}
\usepackage{subfigure}
\usepackage{booktabs} % for professional tables

\usepackage{hyperref}

\usepackage[accepted]{icml2023}

\usepackage{amsmath}
\usepackage{amssymb}
\usepackage{mathtools}
\usepackage{amsthm}
\newcommand{\segframework}{MSEG-VCUQ}
\DeclareUnicodeCharacter{2212}{-}
\usepackage[utf8]{inputenc} % allow utf-8 input
\usepackage[T1]{fontenc}    % use 8-bit T1 fonts
\usepackage{hyperref}       % hyperlinks
\usepackage{url}            % simple URL typesetting
\usepackage{booktabs}       % professional-quality tables
\usepackage{amsfonts}       % blackboard math symbols
\usepackage{nicefrac}       % compact symbols for 1/2, etc.
\usepackage{microtype}      % microtypography
\usepackage{lipsum}
\usepackage{fancyhdr}       % header
\usepackage{graphicx}       % graphics
\graphicspath{{media/}}     % organize your images and other figures under media/ folder

\usepackage[capitalize,noabbrev]{cleveref}

\theoremstyle{plain}

\theoremstyle{definition}

\theoremstyle{remark}

\usepackage[textsize=tiny]{todonotes}

\icmltitlerunning{MSEG-VCUQ}

\begin{document}

\twocolumn[
\icmltitle{\textbf{\segframework{}}: \textbf{M}ultimodal \textbf{SEG}mentation with Enhanced \textbf{V}ision Foundation Models, \textbf{C}onvolutional Neural Networks, and \textbf{U}ncertainty \textbf{Q}uantification for High-Speed Video Phase Detection Data}

\icmlsetsymbol{equal}{*}

\begin{icmlauthorlist}
\icmlauthor{Chika Maduabuchi}{mit}
\icmlauthor{Ericmoore Jossou}{mit}
\icmlauthor{Matteo Bucci}{mit}
\end{icmlauthorlist}

\icmlaffiliation{mit}{Massachusetts Institute of Technology, Cambridge, MA, USA}

\icmlcorrespondingauthor{Chika Maduabuchi}{chika691@mit.edu}

\icmlkeywords{Machine Learning, ICML}

\vskip 0.3in
]
\printAffiliationsAndNotice{}

\begin{abstract}
High-speed video (HSV) phase detection (PD) segmentation is crucial for monitoring vapor, liquid, and microlayer phases in industrial processes. While CNN-based models like U-Net have shown success in simplified shadowgraphy-based two-phase flow (TPF) analysis, their application to complex HSV PD tasks remains unexplored, and vision foundation models (VFMs) have yet to address the complexities of either shadowgraphy-based or PD TPF video segmentation. Existing uncertainty quantification (UQ) methods lack pixel-level reliability for critical metrics like contact line density and dry area fraction, and the absence of large-scale, multimodal experimental datasets tailored to PD segmentation further impedes progress. To address these gaps, we propose \emph{\segframework{}}. This hybrid framework integrates U-Net CNNs with the transformer-based \emph{Segment Anything Model (SAM)} to achieve enhanced segmentation accuracy and cross-modality generalization. Our approach incorporates systematic UQ for robust error assessment and introduces the first open-source multimodal HSV PD datasets. Empirical results demonstrate that \emph{\segframework{}} outperforms baseline CNNs and VFMs, enabling scalable and reliable PD segmentation for real-world boiling dynamics.
\end{abstract}

\section{Introduction}

Boiling heat transfer is a complex phenomenon integral to various industrial and engineering applications, including nuclear reactors, electronics cooling, and chemical processing systems, where efficient thermal management is paramount \cite{RICHENDERFER201835, doi:10.1080/01457632.2023.2191441}. This process is governed by intricate mechanisms such as evaporation, quenching, and single-phase convection, all contributing to the overall heat flux \cite{BUCCI2016115}. Predictive mechanistic models aim to quantify these contributions using key parameters like nucleation site density, bubble departure diameter, contact line density, and dry area fraction. These metrics are crucial for understanding boiling dynamics but depend heavily on high-resolution data obtained through advanced diagnostics such as infrared thermography \cite{doi:10.1080/01457632.2023.2191441, 10.1063/5.0135110} and high-speed video (HSV) phase detection (PD) imaging \cite{KOSSOLAPOV2021103522, Kossolapov_2024}.

\begin{figure}[ht!]
    \centering
    \includegraphics[width=\linewidth]{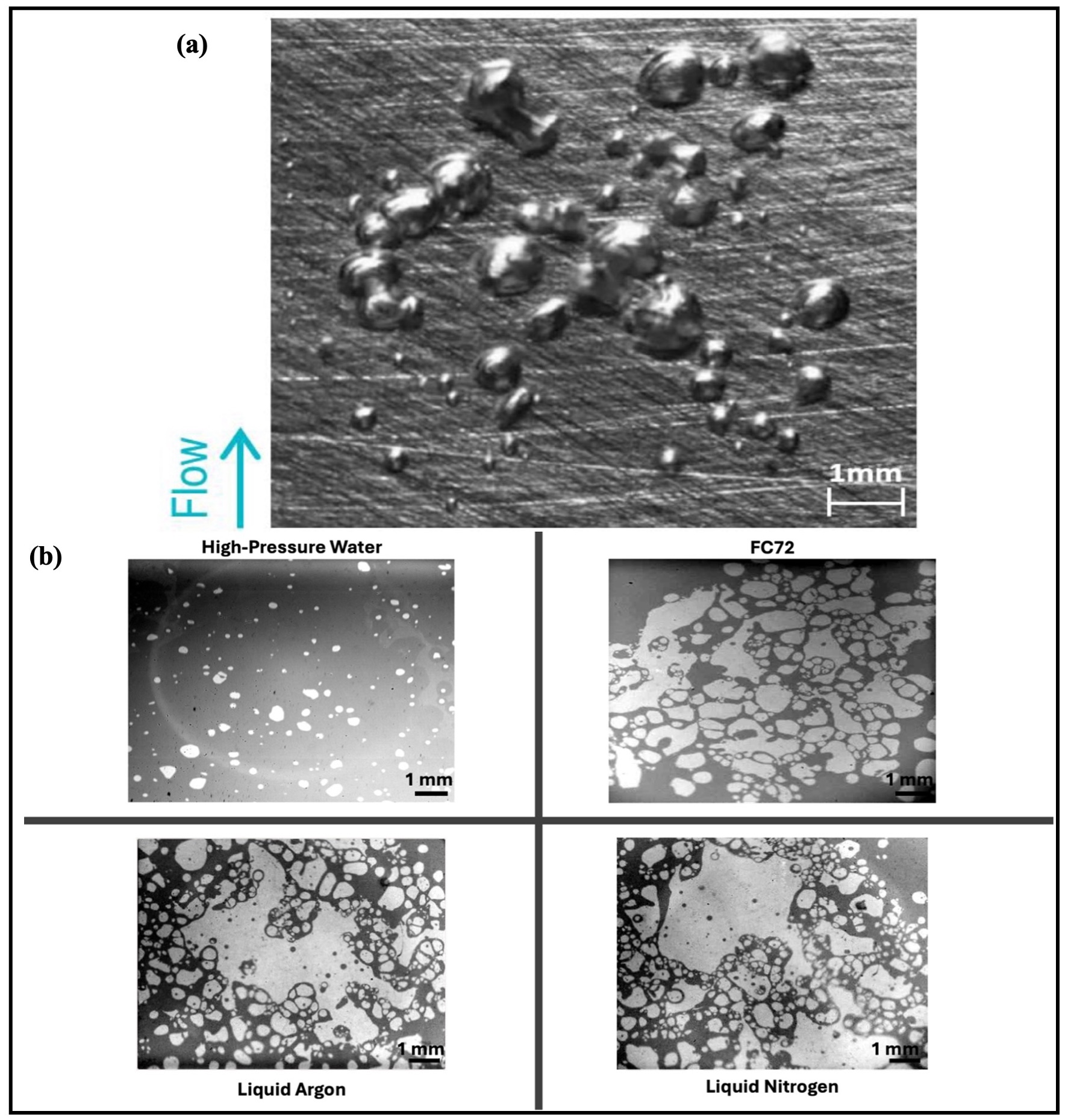}
    \caption{(a) Sample of Front-Lit Shadowgraphy images from two-phase flow. (b) Sample Phase-Detection Images were used in this study.}
    \label{fig:shadowgraphy_images}
\end{figure}

PD images, such as those shown in Figure~\ref{fig:shadowgraphy_images}b, provide crucial insights by differentiating between liquid, vapor, and microlayer phases on a boiling surface \cite{KOSSOLAPOV2021103522}. Yet, traditional methods for analyzing these images, whether manual or semi-automated, require significant time and expertise, limiting scalability for larger multimodal datasets. Convolutional neural networks (CNNs), especially U-Net \cite{SOIBAM2023104589, MALAKHOV2023104402}, have successfully segmented two-phase flow (TPF) images by learning complex features in data, thus providing a promising solution to automate HSV PD segmentation. Unlike the shadowgraphy images typically used for TPF studies (Figure~\ref{fig:shadowgraphy_images}a), PD images feature complex bubble structures and diverse liquid-vapor interactions. These unique attributes create challenges for accurate segmentation, particularly in identifying the bubble footprints and contact lines.

HSV PD data offers a valuable lens into the boiling process by delineating liquid, vapor, and microlayer phases. However, these HSV datasets' manual or semi-automated analysis can be labor-intensive and subject to subjectivity, limiting scalability for large multimodal datasets. Traditional segmentation techniques, including edge detection and thresholding, struggle with overlapping bubbles and dynamic flow patterns, often requiring manual adjustments across experiments. Recently, CNN architectures like U-Net have become the standard for HSV tasks due to their ability to learn intricate visual patterns and deliver precise segmentation results \cite{Zhang2020, Eppel2020, BRZOZOWSKI2022108003}. Nevertheless, these CNN architectures, though powerful, exhibit limited generalizability when applied to multimodal datasets comprising varying experimental conditions, posing a significant barrier to their adoption in scientific PD tasks \cite{SEONG2023104336, RAVICHANDRAN2023110879, PASSONI2024104871}.

This limitation is pronounced in HSV PD analysis, where dynamic physical processes and domain-specific visual characteristics demand flexible yet robust segmentation models. The emergence of vision foundation models, like the Segment Anything Model (SAM), has demonstrated promising generalization capabilities across various segmentation tasks by leveraging transformer-based architectures \cite{Zhang2023}. Despite their potential, adapting these models to scientific HSV PD tasks remains underexplored, with most applications focused on natural images rather than PD in boiling experiments \cite{e24070942}.

To overcome the limitations of traditional CNN-based models in HSV PD segmentation, this work presents \segframework{}. At the core of \segframework{} is VideoSAM, a hybrid segmentation model that integrates CNNs with the transformer-based architecture of SAM to enhance PD segmentation in multimodal HSV data. VideoSAM utilizes U-Net for preliminary mask generation, effectively capturing primary liquid-vapor boundaries and refines these masks through SAM’s attention-driven feature extraction. This combined approach enables VideoSAM to segment complex fluid behaviors and dynamic boiling conditions with higher precision across diverse multimodal datasets, including water, FC-72, nitrogen, and argon.

The open-source dataset curated for this study encompasses a broad spectrum of PD modalities and dynamic fluid behaviors, establishing a strong foundation for training and evaluating VideoSAM. Additionally, \segframework{} incorporates an uncertainty quantification (UQ) module to assess the reliability of boiling metrics such as dry area fraction and contact line density, which are essential for accurately modeling boiling dynamics. Experimental results demonstrate that VideoSAM consistently outperforms traditional custom models, including U-Net, especially in complex scenarios characterized by varied bubble formations and intricate boundary delineation. These findings highlight the model’s versatility and scalability, positioning \segframework{} as a comprehensive solution for HSV PD segmentation in multimodal boiling data for advanced scientific and engineering applications in autonomous experiments.

\section{Related Works}

\subsection{Research Gaps}  
The development of image segmentation techniques for TPF analysis has shown significant progress, transitioning from classical image processing methods~\cite{4721870, s17061448, ZHOU2020103277, RICHENDERFER201835, JIN2021121517, HANAFIZADEH2011327, Singh_2009} to advanced deep learning models~\cite{PASSONI2024104871, MALAKHOV2023104402, Kim2021, SOIBAM2023104589, SEONG2023104336, RAVICHANDRAN2023110879, SUH2024100309}. Despite these advancements, several critical gaps remain unaddressed, limiting the scalability and precision of current methods for PD tasks in TPF systems.  

First, while CNNs such as U-Net~\cite{10.1007/978-3-319-24574-4_28} have demonstrated success in segmenting simpler shadowgraphy-based TPF images~\cite{JUNGST2024106314, Choi2022}, their potential remains entirely unexplored for the more intricate PD tasks. Similarly, scalable VFMs like SAM~\cite{Kirillov_2023_ICCV}, despite achieving groundbreaking segmentation performance across diverse domains, have yet to be applied to either shadowgraphy-based TPF segmentation or the complex HSV PD challenges. This dual gap highlights a significant opportunity to leverage the robust feature extraction capabilities of CNNs and the scalability, adaptability, and fine-grained precision of VFMs to address real-world HSV PD segmentation complexities, unlocking transformative advancements in the field.
  
Moreover, there is a notable lack of open-source multimodal HSV datasets tailored specifically for PD segmentation. Existing datasets~\cite{10.1115/HT2022-85582, Hu2023, NEURIPS2023_01726ae0} suffer from significant limitations. First, they are derived predominantly from simplistic front/back-lit shadowgraphy techniques or theoretical simulations, making them inadequate for real-world experimental PD conditions. Second, these datasets are often unimodal, limited in scale, and lacking the diversity and complexity required to develop scalable, generalizable models. Finally, this absence of representative, large-scale experimental datasets restricts the potential for developing robust VFMs tailored to PD segmentation tasks.
 
Furthermore, existing UQ studies~\cite{10.1007/978-3-030-59861-7_9, MUNIA2025102719, WAHID2024110542, LIU20191096} primarily focus on general-purpose applications, with limited attention to HSV PD analysis. Moreover, these approaches rarely incorporate real experimental metrics such as contact line density and dry area fraction at the pixel level---two key parameters for accurate and reliable TPF segmentation. This omission hinders the development of robust models capable of quantifying uncertainties inherent in PD segmentation.

Therefore, this paper builds upon prior work in~\cite{maduabuchi2024videosamlargevisionfoundation, Maduabuchi2024} by highlighting the evolution of image processing techniques, the role of deep learning, and the challenges posed by existing datasets and UQ methods in Appendices~\ref{sec:traditional_methods}--\ref{sec:uncertainty_quantification}. By leveraging scalable VFMs, robust uncertainty quantification, and newly curated open-source experimental datasets, our work establishes a transformative framework for advancing HSV PD segmentation.

\subsection{Technical Contributions}

To address the critical limitations in existing TPF segmentation frameworks, our work introduces a comprehensive and scalable approach tailored specifically for HSV PD segmentation. The key contributions are as follows:

\subsubsection{Transfer Learning for PD Analysis}
We bridge the gap in pixel-level segmentation for PD by adapting transfer learning from biological cell imaging domains to HSV datasets. This ensures that custom U-Net models can effectively capture the fine distinctions between vapor, liquid, and microlayer phases, overcoming the object-level boundary limitations of traditional CNNs like U-Net and Mask R-CNN. By leveraging domain knowledge from analogous tasks, our models achieve superior performance under complex, multimodal PD datasets encompassing diverse boiling conditions and fluctuating heat flux scenarios.

\subsubsection{Integrated Uncertainty Quantification}
We introduce a robust UQ pipeline that systematically quantifies errors in pixel-level segmentation metrics, including contact line density and dry area fraction. Unlike existing general-purpose UQ methods, our framework directly addresses the variability inherent in experimental PD data, ensuring reliable and interpretable outputs. Open-source scripts for quantifying discretization errors further enhance reproducibility and accuracy in assessing phase transitions and boiling heat transfer dynamics.

\subsubsection{Open-Source Multimodal PD Datasets.}
To overcome the scarcity of representative datasets for PD tasks, we curate and release the first large-scale, multimodal HSV datasets tailored for PD segmentation. These datasets encompass diverse experimental conditions and boiling regimes, significantly surpassing existing unimodal and simulation-driven datasets~\cite{10.1115/HT2022-85582, Hu2023, NEURIPS2023_01726ae0}. This contribution provides a foundation for developing scalable, generalizable segmentation models.

\subsubsection{Extension to Vision Foundation Models}
We demonstrate, for the first time, the applicability of scalable VFMs, such as SAM~\cite{Kirillov_2023_ICCV}, to PD segmentation. By fine-tuning SAM on our curated PD datasets, we bridge the scalability gap, enabling VFMs to generalize effectively across diverse modalities and experimental conditions. This advancement ensures seamless adaptation to new PD scenarios, addressing the specificity constraints of earlier CNN-based approaches.

\subsubsection{Holistic Framework for PD Analysis}
Our integrated framework—spanning segmentation, uncertainty quantification, dataset development, and cross-dataset adaptability—provides an open-source, reproducible solution for advancing PD segmentation research. By equipping the community with datasets, tools, and scalable models, we establish a versatile platform for addressing the complexities of boiling heat transfer, fostering innovation across experimental and computational research contexts.

\section{Methodology}

\begin{figure}[ht!]
    \centering
    \includegraphics[width=\linewidth]{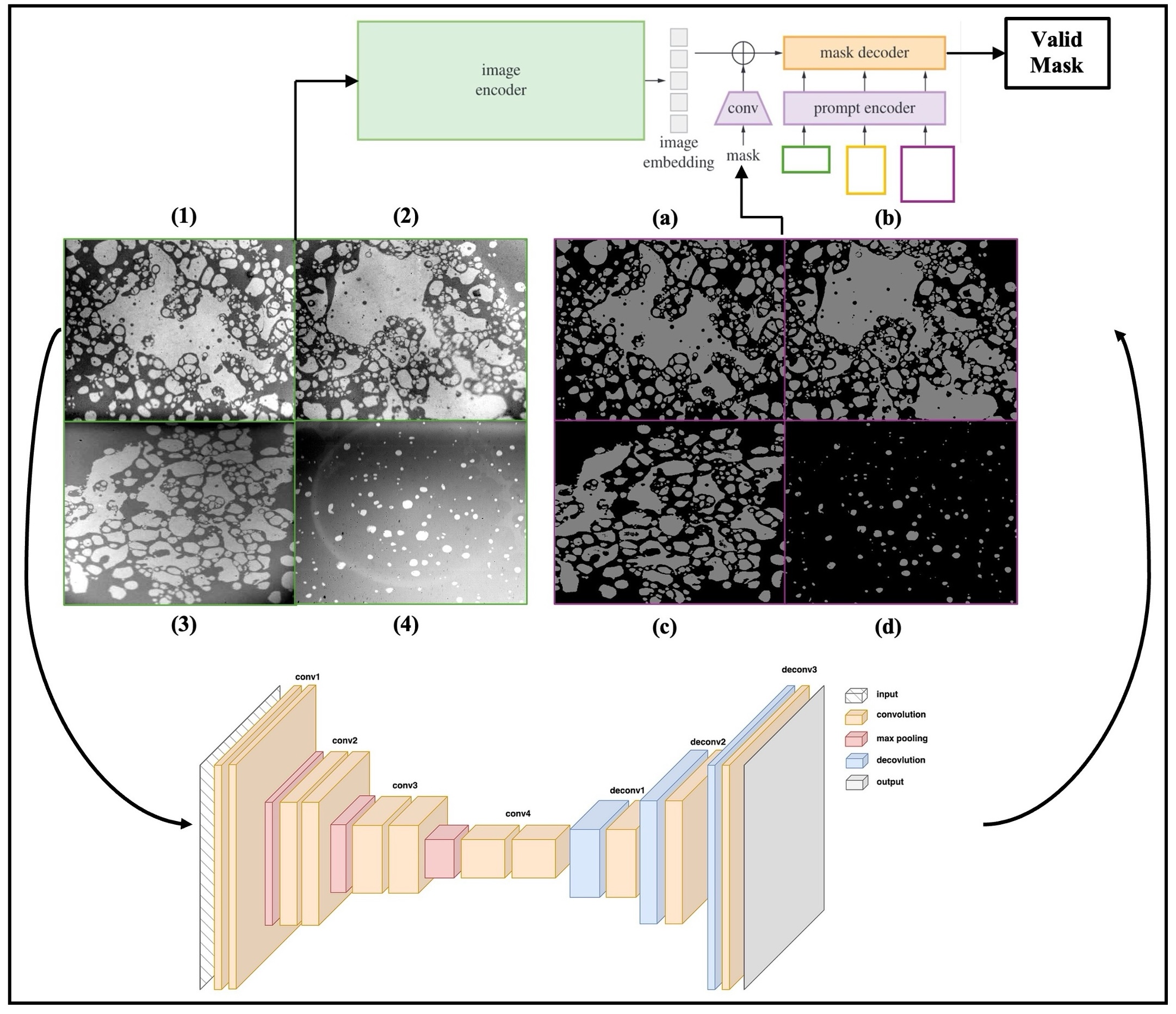}
    \caption{Illustration of the integrated process within the VideoSAM architecture. Initially, fine-tuned U-Net models produce segmentation masks tailored to each fluid modality, capturing primary liquid-vapor boundaries. These masks are then paired with their corresponding images and processed by the VideoSAM transformer. The model refines segmentation outputs through its image encoder and mask decoder, leveraging SAM’s pre-trained components to achieve consistent and high-accuracy HSV segmentation across various experimental conditions.}
    \label{fig:arch}
\end{figure}

\paragraph {Model Architecture}  
The \segframework{} architecture is a hybrid model that integrates U-Net with VideoSAM to enhance phase detection (PD) segmentation from high-speed video (HSV) data, as shown in Figure~\ref{fig:arch}. This design leverages U-Net’s strengths in generating initial, detailed segmentation masks, which capture primary liquid-vapor boundaries. To enhance generalization, these initial masks are refined using VideoSAM’s advanced transformer-based attention mechanisms and feature extraction capabilities. This dual-stage process allows \segframework{} to address the inherent complexities of boiling phenomena, including overlapping bubbles, diverse fluid modalities, and challenging lighting conditions, achieving pixel-level precision across various experimental datasets. Further architectural details, equations, and illustrations of the workflow are provided in Appendix~\ref{sec:model_architecture}.

\paragraph{Data Collection and Processing}  
To establish a robust and generalizable segmentation framework, we curated a multimodal high-speed video (HSV) dataset that captures a wide range of boiling conditions. The dataset includes multiple fluids and experimental setups, ensuring diversity in boiling regimes, fluid properties, and imaging conditions. A systematic preprocessing pipeline was implemented to enhance data quality and optimize model performance. This pipeline includes grayscale conversion, patchification, and semi-automated annotation of boiling video frames. A detailed explanation of the data acquisition, experimental configurations, and annotation strategies is provided in Appendix~\ref{sec:data_processing}.  

The experimental dataset comprises video frame-mask pairs collected under both Saturated Pool Boiling (SPB) and Flow Boiling (FB) at 500 kg/m$^2$s conditions. These scenarios encompass fluids with distinct thermal and physical properties recorded at high heat fluxes and pressures. Table~\ref{tab:experimental_conditions} summarizes the key experimental conditions, including the boiling modality, heat flux (\textit{q}), and the number of frames available for each fluid.

\begin{table}[ht]
\caption{Summary of Experimental Conditions for Data Collection}
\label{tab:experimental_conditions}
\centering
\begin{tabular}{lccc}
\toprule
\textbf{Modality} & \textbf{Condition} & \textbf{\textit{q} (kW/m$^2$)} & \textbf{Frames} \\
\midrule
Argon     & SPB & 120 & 6,000 \\
Nitrogen  & SPB & 120 & 6,000 \\
FC-72     & SPB & 170 & 6,000 \\
Water     & FB & 3,000 & 7,500 \\
\bottomrule
\end{tabular}
\end{table}

This multimodal dataset is the foundation for training and validating segmentation models, ensuring their applicability across different fluids and boiling scenarios.
\paragraph{U-Net CNN}  
The U-Net model is the backbone for preliminary segmentation, leveraging transfer learning and a feature-rich encoder-decoder architecture. We fine-tune custom U-Net for each fluid modality, with key components such as skip connections and upsampling enhancing its ability to capture bubble structures. Technical details on the U-Net architecture, transfer learning strategy, and the feature transfer pipeline can be found in Appendix~\ref{sec:unet_cnn}.

\paragraph{VideoSAM Model}  
VideoSAM extends the SAM architecture for HSV PD segmentation by integrating CNN-based initial segmentation with transformer-based refinement. The model utilizes a flexible prompt encoder, image encoder, and mask decoder to achieve pixel-level precision. Training and inference pipelines are optimized for multimodal datasets, ensuring robust segmentation across experimental conditions. Appendix~\ref{sec:videosam_model} provides full implementation details, equations, and illustrations.

\paragraph{Uncertainty Quantification}  
To assess the reliability of PD segmentation outputs, we employ a rigorous uncertainty quantification (UQ) framework that evaluates discretization errors in pixel-based measurements. We estimate errors in metrics such as contact line density and dry area fraction through iterative simulations and weighted frequency analysis. The methodology, including theoretical formulations and algorithmic details, is discussed in Appendix~\ref{sec:uq_boiling}.

\section{Results and Discussions}

\begin{figure*}[ht!]
    \centering
    \includegraphics[width=1\textwidth]{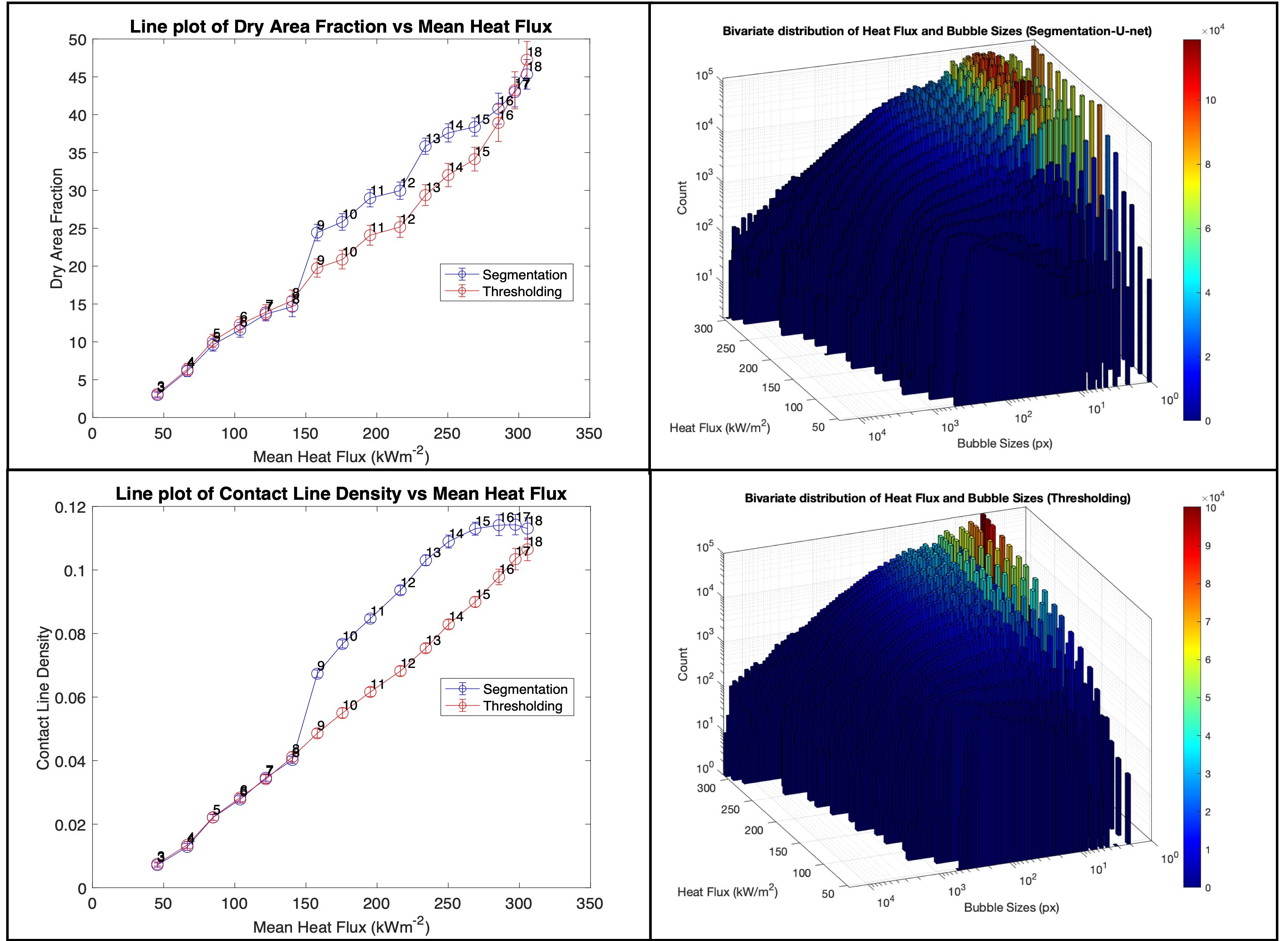}
    \caption{Variation of Dry Area Fraction and Contact Line Density with Increasing Mean Heat Flux (left plots) and 3D Histogram of Heat Flux vs. Bubble Sizes Distribution (right plots) Using Segmentation (U-Net CNN) and Thresholding Techniques.}
    \label{fig:combined_boiling_metrics}
\end{figure*}

\paragraph{Comparative Analysis of Dry Area Fraction and Contact Line Density}  
The analysis of boiling metrics, shown in Figure~\ref{fig:combined_boiling_metrics}, reveals clear trends in dry area fraction and contact line density as functions of heat flux under saturated pool boiling conditions. Both segmentation methods—U-Net CNN and adaptive thresholding—indicate an upward trend consistent with expected boiling dynamics where vapor coverage increases with rising heat flux. Both techniques yield similar results at lower heat flux levels; however, significant deviations appear as the heat flux exceeds 140 kW/m². The U-Net consistently reports higher dry area fractions and contact line densities, particularly between 140-158 kW/m², where smaller, closely packed vapor bubbles dominate the interface. This difference underscores the U-Net’s sensitivity to finer vapor structures, often overlooked by the thresholding technique. Such accuracy in detecting small-scale features is crucial for understanding the dynamic behavior of the liquid-vapor interface and improving the prediction of heat transfer performance. Detailed statistical analyses are provided in Appendix~\ref{sec:comparative_analysis} for a more granular comparison of the segmentation errors.

\paragraph{3D Histogram Analysis of Heat Flux vs. Bubble Size Distribution}  
The right-side plots of Figure~\ref{fig:combined_boiling_metrics} illustrate the bivariate distribution of bubble sizes across varying heat flux levels, highlighting the comparative performance of U-Net CNN and thresholding. At low heat flux, the bubble size distribution skews toward smaller bubbles, with the U-Net effectively capturing a broad range of bubble sizes. As the heat flux increases, the histogram shifts toward larger bubbles, reflecting the onset of vigorous boiling. The U-Net segmentation highlights a smooth progression of bubble size growth, with a notable concentration of medium-sized bubbles at intermediate flux levels. In contrast, the thresholding method struggles to detect smaller bubbles, particularly at lower flux levels, resulting in an incomplete distribution skewed toward larger sizes. The ability of the U-Net CNN to resolve fine-scale details contributes to a more comprehensive understanding of bubble dynamics across heat flux regimes, which is critical for optimizing thermal management systems. Full histogram comparisons and additional heat flux levels are detailed in Appendix~\ref{sec:3D_histogram}.

\paragraph{Comparison of Bubble Size Distribution and Visualizations}  
Figure~\ref{fig:bubble_size_comparison} compares bubble size distributions for frames extracted from videos 8 and 9. The raw camera outputs, displayed as grayscale images, provide a baseline for assessing segmentation performance. The U-Net results, color-coded by bubble size (blue for small, green for medium, yellow for large, and orange for the largest), exhibit a diverse size range with clear distinctions between individual bubbles, even in densely packed regions. This fine-grained segmentation confirms the U-Net’s robustness in capturing intricate bubble dynamics, particularly in active boiling phases where bubble coalescence and fragmentation occur simultaneously. In comparison, the thresholding method highlights larger bubbles effectively but often merges or overlooks smaller bubbles, reducing segmentation accuracy. This difference is particularly pronounced in regions with closely packed bubbles, where U-Net consistently outperforms thresholding. These results emphasize the value of integrating U-Net segmentation into boiling experiments requiring precise bubble size characterization. Detailed visualizations of additional frames are included in Appendix~\ref{sec:bubble_sizes}

\paragraph{Validation with Ground Truth Annotations}  
To assess the accuracy of the U-Net CNN model, we compared its results against ground truth annotations manually provided by five independent domain experts (Figure~\ref{fig:ground_truth_benchmark}). The expert-annotated dry area fraction and contact line density values clustered around 0.17 and 0.05, respectively, with minimal inter-user variability. The U-Net segmentation aligns closely with the ground truth, achieving deviations below 2\% for dry area fraction and under 8\% for contact line density. In contrast, the thresholding method underestimates both metrics, particularly in frames with smaller bubbles or complex bubble interactions. The observed underestimation arises from thresholding’s limitations in detecting fine-scale details, leading to gaps in bubble identification. Error quantification (detailed in Appendix~\ref{sec:ground_benchmark}) confirms that U-Net segmentation provides a more reliable approximation of ground-truth metrics, underscoring its suitability for high-precision boiling analysis.

\paragraph{Statistical Comparison of Segmentation Across Modalities}  
The robustness of the U-Net CNN model was further validated through a statistical comparison of boiling metrics across argon, nitrogen, and FC-72 modalities, as illustrated in Figure~\ref{fig:statistical_analysis}. In argon and nitrogen, both the U-Net and thresholding methods yielded comparable results, with dry area fractions ranging between 0.46-0.48 and contact line densities between 0.06-0.07. This consistency highlights the segmentation methods' efficacy in relatively simpler boiling regimes. However, the FC-72 dataset revealed significant discrepancies between the two methods. The U-Net results showed a more clustered distribution of dry area fraction and contact line density, while thresholding exhibited greater variability and a higher incidence of outliers. These deviations align with FC-72's complex boiling patterns, characterized by smaller, densely packed bubbles that challenge the thresholding technique. Probability density functions, cumulative distribution functions, and box plots (included in Appendix~\ref{sec:stat_comparison}) further confirm the U-Net’s superior performance in complex boiling scenarios, solidifying its role as a robust segmentation tool.

\paragraph{Qualitative and Quantitative Performance across Modalities}  
High-pressure water boiling experiments (10 bar, 3000 kW/m², and 40 bar, 3400 kW/m²) were used to evaluate the transferability and adaptability of the U-Net CNN model (Figure~\ref{fig:perimeter_visualization}). Initially trained on liquid nitrogen data, the U-Net exhibited limitations when applied to water boiling, particularly in detecting large bubbles where hollow centers and false positives appeared. Fine-tuning the model with annotated high-pressure water frames significantly improved segmentation accuracy, as indicated by a close match between the fine-tuned U-Net results and expert-verified ground truth perimeters. These findings underscore the importance of domain-specific tuning for accurate segmentation in varying experimental conditions. This iterative approach—combining pre-trained models with fine-tuning—demonstrates the U-Net’s adaptability for diverse boiling environments, providing a foundation for robust analysis in high-pressure applications. Further results showing the qualitative perimeter visualization results across various modalities are provided in Appendix~\ref{sec:perimeter_visuals}. Meanwhile, the multi-modal quantitative results are presented here in Appendix~\ref{sec:quantitative_results}.

\begin{figure*}[ht!]
    \centering
    \includegraphics[width=\textwidth]{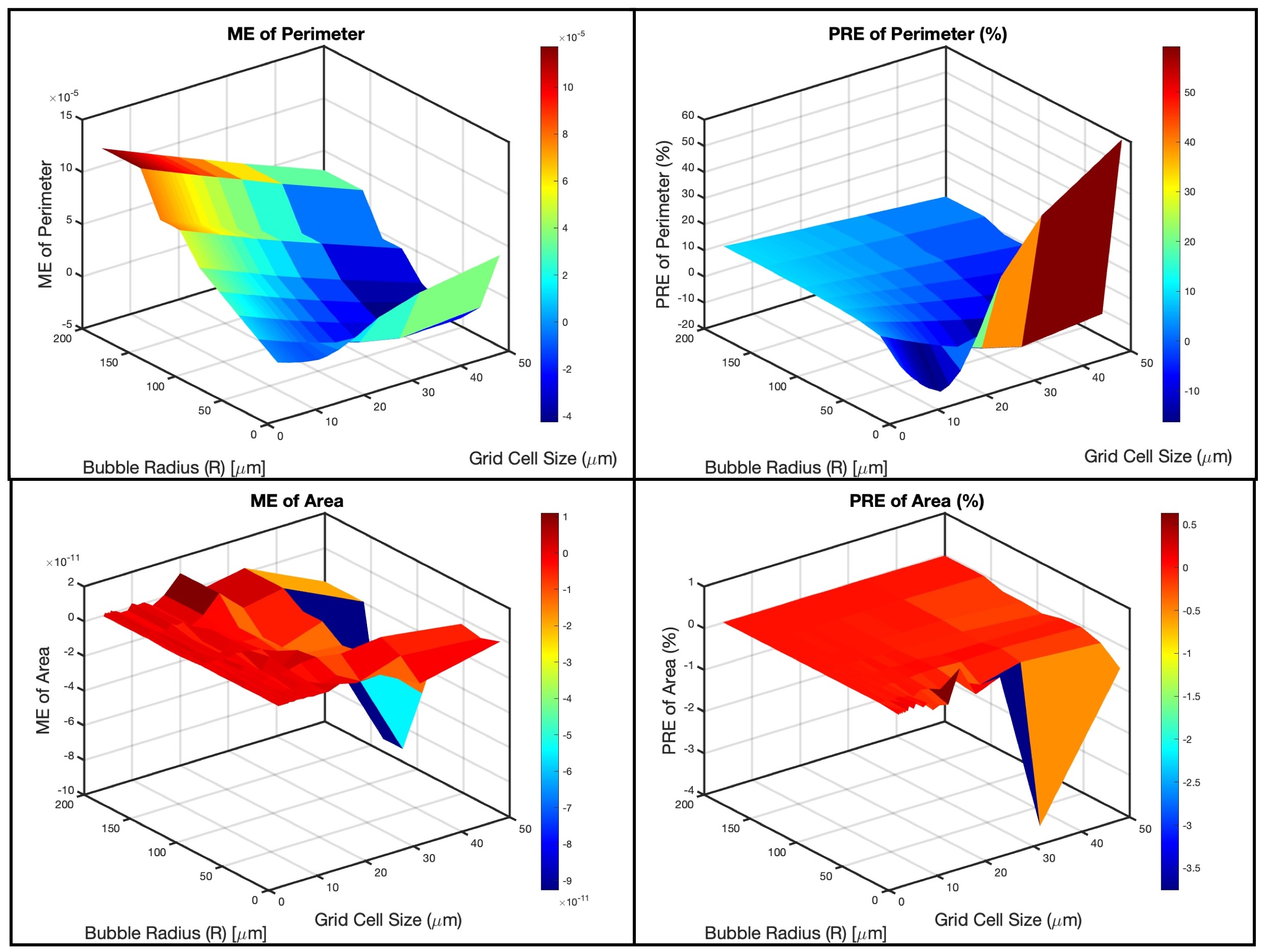}
    \caption{Mean Error (ME) and Percentage Relative Error (PRE) of Perimeter and Area Variations with Bubble Radius and Grid Cell Size.}
    \label{fig:radii_grid_impacts}
\end{figure*}

\paragraph{Uncertainty Quantification in Performance Parameters}  
To analyze the impact of discretization on boiling metrics, we investigated the effects of grid resolution and bubble size on dry area fraction and contact line density. These parameters were evaluated using numerical experiments, where grid resolutions and bubble radii were systematically varied. The methodology, detailed in Appendix~\ref{sec:uncertainty_quantification_results}, compares the discretized measurements of bubble perimeter and area with their theoretical values, quantified using Percentage Relative Error (PRE) and Mean Error (ME) metrics. Figure~\ref{fig:radii_grid_impacts} illustrates that perimeter measurements are particularly sensitive to grid size, transitioning from overestimation at fine resolutions to underestimation at coarser ones. In contrast, area measurements exhibit greater resilience, stabilizing more effectively across bubble radii and grid sizes.

\paragraph{Experimental Data Validation}  
A representative frame from the liquid argon dataset was selected and analyzed at a resolution of 12.6 $\mu$m/px under saturated boiling conditions to validate these numerical findings. The results, presented in Appendix ~\ref{sec:convergence_test}, show the distributions of bubble perimeter, area, and radius (Figure~\ref{fig:argondistributions_convergence}). The probability distributions confirm that smaller bubbles dominate the sample, while a logarithmic scale highlights the presence of occasional larger bubbles. Convergence tests for PRE, performed over iteration milestones (5K to 20K), demonstrate that area calculations require more iterations to stabilize than the perimeter. At 20K iterations, numerical errors converge to negligible levels, balancing accuracy with computational efficiency.

\paragraph{Grid Resolution and Bubble Size Influence}  
The impact of grid resolution and bubble size on measurement accuracy was further analyzed in Appendix ~\ref{sec:grid_resolution}. Figure~\ref{fig:radii_grid_impacts} demonstrates the trends in ME and PRE for perimeter and area calculations. The results highlight a non-linear relationship between grid resolution and bubble characteristics, with perimeter errors exhibiting a distinct transition from overestimation to underestimation. Area errors remain predominantly negative, with minimal sensitivity to grid changes. These findings emphasize the need for fine grid resolutions to minimize the mistakes in perimeter estimation, particularly for smaller bubbles.

\paragraph{Erosion and Dilation Effects on Boundary Accuracy}  
We also examined the effects of boundary modifications, such as erosion and dilation, on measurement uncertainties. Appendix ~\ref{sec:erosion_dilation}, and Figure~\ref{fig:erosion_dilation} visually illustrate these effects. Dilation reduces edge-related artifacts, decreasing perimeter PRE from 13.8\% to 6.6\% (Table~\ref{tab:boundary_modifications}). This smoothing effect ensures closer alignment between perimeter estimates and ground truth values. While area PRE remains stable, the findings underscore the importance of selecting appropriate boundary conditions in digital image processing to refine bubble segmentation accuracy.

\paragraph{Uncertainty Quantification Across Modalities}  
The same uncertainty analysis was extended to FC-72, liquid argon (LAr), liquid nitrogen (LN2), and high-pressure water datasets, as detailed in Appendix ~\ref{sec:modality_uq}. Table~\ref{tab:uncertainty_modalities} summarizes the results, showing consistent PRE values for area measurements across all modalities. However, perimeter uncertainties vary significantly, with LN2 exhibiting the highest PRE (-8.4\%) due to pronounced boundary irregularities. These discrepancies highlight the greater sensitivity of perimeter calculations to segmentation accuracy and pixel resolution, particularly in datasets with complex boiling dynamics.

\begin{table}[ht!]
    \centering
    \caption{Uncertainty Quantification for Each Modality}
    \label{tab:uncertainty_modalities}
    \begin{tabular}{lccc}
        \toprule
        \textbf{Modality} & $\alpha$ & $\beta$ & $\gamma$ \\
        \midrule
        FC-72 & -0.05 & -6.8 & 2.5e-06 \\
        LAr   & -0.05 & -6.5 & 1.2e-06 \\
        LN2   & -0.06 & -8.4 & -2.4e-06 \\
        Water & -0.03 & -1.0 & 1.9e-05 \\
        \bottomrule
    \end{tabular}
    \vspace{2mm} % Space before the definitions
    \footnotesize{
    
    \textbf{Symbol Definitions:} \\
    $\alpha$: Weighted Avg PRE Area (\%) \\
    $\beta$: Weighted Avg PRE Perimeter (\%) \\
    $\gamma$: Weighted Avg ME Perimeter ($\mu$m)
    }
\end{table}

\paragraph{VideoSAM Segmentation Across Modalities}  
We evaluated the zero-shot generalization performance of VideoSAM across nitrogen, FC-72, and water datasets. Figure~\ref{fig:videosam_results}, detailed in Appendix ~\ref{sec:videosam_multimodal}, compares VideoSAM’s results with SAM and ground truth masks. VideoSAM demonstrates robust performance in nitrogen and FC-72 datasets, effectively capturing intricate bubble boundaries and outperforming SAM in terms of accuracy, IoU, and Dice scores. For instance, in the nitrogen dataset, VideoSAM achieves an IoU of 0.8317, compared to SAM’s 0.6702. However, its performance deteriorates in the water dataset, where low object variability and background noise undermine segmentation accuracy. These results highlight VideoSAM’s strength in handling dynamic environments with high bubble density and complex structures.

\paragraph{Multimodal Segmentation Performance}  
To further benchmark VideoSAM against U-Net and SAM, segmentation accuracy was compared across water, FC-72, nitrogen, and argon datasets. Table~\ref{tab:performance_metrics} and Figure~\ref{fig:videosam_results} (Appendix ~\ref{sec:videosam_multimodal}) present the results for IoU and F1 scores. VideoSAM consistently outperforms other models in complex environments like nitrogen and FC-72, achieving IoU values of 0.83 and 0.80, respectively. In contrast, U-Net performs best in simpler datasets, such as water, with an IoU of 0.5619, underscoring its adaptability to low-contrast, uniform backgrounds. These results confirm the trade-offs between model generalization capacity and specialization, with VideoSAM excelling in dense, irregular boiling structures and U-Net delivering superior results for uniform datasets.

\begin{table}[ht!]
    \centering
    \caption{Comparison of IoU and F1 Score for U-Net, VideoSAM, and SAM across Modalities.}
    \label{tab:performance_metrics}
    \begin{tabular}{lccc}
        \hline
        Modality & Model & IoU & F1 Score \\
        \hline
        Water    & U-Net       & \textbf{0.5619} & \textbf{0.7191} \\
                 & SAM         & 0.0620 & 0.1165 \\
                 & VideoSAM    & 0.1894 & 0.3143 \\
        FC-72    & U-Net       & 0.7244 & 0.8400 \\
                 & SAM         & 0.5721 & 0.7278 \\
                 & VideoSAM    & \textbf{0.7997} & \textbf{0.8885} \\
        Nitrogen & U-Net       & 0.7547 & 0.8602 \\
                 & SAM         & 0.6702 & 0.8025 \\
                 & VideoSAM    & \textbf{0.8317} & \textbf{0.9080} \\
        Argon    & U-Net       & 0.7815 & 0.8773 \\
                 & SAM         & 0.6464 & 0.7852 \\
                 & VideoSAM    & \textbf{0.8384} & \textbf{0.9120} \\
        \hline
    \end{tabular}
\end{table}

\section{Conclusions}

\segframework{} developed and validated advanced methodologies for HSV PD segmentation, emphasizing model performance and uncertainty quantification across diverse modalities. We utilized the newly developed VideoSAM model incorporating U-Net CNNs and transformer-based SAM architecture to examine multimodal HSV PD segmentation for the first time.

Our findings demonstrate that the novel application of U-Net outperformed adaptive thresholding in accurately identifying bubble structures, especially under complex HSV PD modalities. VideoSAM further improved segmentation quality and generalization ability, particularly in dense modalities like FC-72 and Nitrogen, achieving high Intersection over Union (IoU) and F1 scores. However, in simpler modalities like Water, VideoSAM exhibited some limitations, indicating potential overfitting to complex scenarios. Future work should explore enhancements, such as hybrid architectures and multi-scale feature aggregation, to expand VideoSAM's adaptability.

We also conducted an uncertainty quantification study on pixelation effects for boiling metrics, revealing higher relative uncertainty in contact line density compared to dry area fraction. Applying dilation techniques reduced perimeter overestimation, supporting robust boundary representation.

Additionally, we contribute an open-source HSV PD segmentation dataset, providing a valuable resource for advancing HSV PD analysis in dynamic modalities. Expanding real-time segmentation and integrating temporal dependencies are essential next steps, especially for applications requiring dynamic monitoring.

This research offers a framework combining VideoSAM's advanced capabilities and uncertainty quantification across various modalities. Future studies could further refine generalizable segmentation models, uncertainty quantification methods, and domain-specific fine-tuning techniques such as LoRA, VPT, and SSF, advancing HSV PD segmentation for autonomous experimentation applications.

% \bibliography{example_paper}
% \bibliographystyle{icml2023}

%%%%%%%%%%%%%%%%%%%%%%%%%%%%%%%%%%%%%%%%%%%%%%%%%%%%%%%%%%%%%%%%%%%%%%%%%%%%%%%
%%%%%%%%%%%%%%%%%%%%%%%%%%%%%%%%%%%%%%%%%%%%%%%%%%%%%%%%%%%%%%%%%%%%%%%%%%%%%%%
% APPENDIX
%%%%%%%%%%%%%%%%%%%%%%%%%%%%%%%%%%%%%%%%%%%%%%%%%%%%%%%%%%%%%%%%%%%%%%%%%%%%%%%
%%%%%%%%%%%%%%%%%%%%%%%%%%%%%%%%%%%%%%%%%%%%%%%%%%%%%%%%%%%%%%%%%%%%%%%%%%%%%%%
\newpage
\appendix
\onecolumn

\section*{Appendix}
\section{Related Works}
\subsection{Traditional Image Processing Methods in Two-Phase Flow Analysis}
\label{sec:traditional_methods}

Before adopting deep learning, various image-processing algorithms were employed to track bubbles in TPF scenarios. Early approaches focused on classical computer vision techniques, including Canny edge detection enhanced with Gaussian smoothing to mitigate noise. Wenyin et al. \cite{4721870} leveraged these methods to improve the segmentation of gas-liquid interfaces from HSV data. Their approach utilized gradient non-maximum suppression and dual thresholding, extracting bubble contours with notable precision. However, these algorithms encountered challenges when applied to complex scenarios involving overlapping bubble boundaries, highlighting the need for further improvements.

Building on these earlier efforts, Paz et al. \cite{s17061448} proposed machine vision algorithms that extended segmentation capabilities for bubbles in subcooled TPF boiling phenomena. Their work incorporated side and front views of the boiling process, improving the adaptability of segmentation across different experimental setups. Dynamic thresholding and edge detection allowed the extraction of bubble outlines and trajectories, facilitating more detailed flow analysis. However, setup calibration and parameterization remained key challenges, requiring manual tuning to achieve optimal performance under diverse conditions.

Zhou and Niu \cite{ZHOU2020103277} advanced bubble detection with a multi-frame image processing algorithm in another significant contribution. Their approach utilized binarization and a predictor-corrector method to handle overlapping bubble images better. This method significantly enhanced bubble size and velocity measurements in experiments featuring dense bubbly plumes. Despite these improvements, the algorithm struggled to distinguish individual bubbles when significant overlap occurred, treating merged bubbles as single entities. This limitation underscored the need for more adaptive segmentation methods capable of performing well in varied flow conditions.

Further refinement came from Richenderfer et al. \cite{RICHENDERFER201835}, who developed an experimental methodology using HSV and infrared thermometry to measure key boiling parameters, including nucleation site density and bubble dynamics. Their method incorporated a combination of gray threshold filters and watershed segmentation to partition heat flux accurately, providing insights into the contribution of microlayer evaporation. Although effective, traditional thresholding techniques faced limitations under high heat flux conditions, where they often failed to segment bubbles accurately, prompting the need for more advanced methodologies.

These early efforts \cite{JIN2021121517, HANAFIZADEH2011327, Singh_2009} illustrate the potential of traditional image processing methods for segmenting bubble behavior in HSV data. They also expose their inherent shortcomings, particularly in the face of variable lighting conditions, overlapping bubbles, and fluid-specific properties. The absence of a universally applicable segmentation framework necessitated the shift towards deep learning architectures, capable of adapting to the complex, dynamic nature of boiling phenomena through learned features and robust generalization.

Though effective under controlled conditions, traditional algorithms faltered when confronted with complex flow patterns, overlapping interfaces, and fluctuating thermal dynamics. These early methods, including edge detection and threshold-based segmentation, lacked generalizability across varying experimental setups, requiring manual adjustments for each scenario. This inconsistency underscored the need for more flexible and adaptive frameworks. 

\subsection{Deep Learning Architectures for HSV Two-Phase Flow Segmentation}
\label{sec:dl_methods}

In recent years, deep learning methods have gained prominence in addressing the challenges posed by complex flow segmentation tasks, surpassing traditional image processing approaches. This section discusses the development and refinement of deep learning models, focusing on CNNs like U-Net and Mask R-CNN, and how they have evolved to meet the demands of precise segmentation and quantification of flow characteristics in HSV.

Traditional CNN-based segmentation tools such as U-Net \cite{10.1007/978-3-319-24574-4_28} and Mask R-CNN \cite{8237584} have become the backbone of bubble segmentation. These models were initially employed to detect objects such as bubbles, droplets, and particles within heated or cooled fluids, often under varying conditions. Studies by Passoni et al. \cite{PASSONI2024104871} and Malakhov et al. \cite{MALAKHOV2023104402} highlight the success of these CNNs, with U-Net excelling in detecting smaller structures but struggling with generalization when faced with unseen flow conditions. While computationally heavier, Mask R-CNN achieved commendable instance segmentation by leveraging the COCO dataset pre-trained weights \cite{10.1007/978-3-319-10602-1_48}, allowing more detailed shape extraction of overlapping bubbles.

Transfer learning in these models has emerged as a critical step to overcome the limitations of small datasets. For instance, Kim and Park \cite{Kim2021} enhanced Mask R-CNN’s capabilities by pre-training on biological cell imagery, demonstrating that domain alignment plays a pivotal role in model performance. They achieved an average precision of 0.981, reducing mask extraction times compared to manual techniques. Similarly, Soibam et al. \cite{SOIBAM2023104589} utilized transfer learning to develop a robust YOLOv7 model tailored for subcooled boiling scenarios. Their work demonstrates that combining deep learning with transfer learning enables models to achieve higher Intersection over Union (IoU) scores (88\%) than traditional algorithms.

Despite these advancements, several models still need help. Seong et al. \cite{SEONG2023104336} identified that traditional U-Net struggled to capture intricate phase boundaries while effectively segmenting specific bubbles, limiting its utility across diverse conditions. Ravichandran et al. \cite{RAVICHANDRAN2023110879} addressed this by developing a 3D U-Net that incorporated infrared (IR) thermometry data for more accurate detection of dry patches. However, their model required extensive customization for new experimental setups, reducing scalability.

Suh et al. \cite{SUH2024100309} introduced the Vision-Inspired Online Nuclei Tracking System (VISION-iT), which integrated Mask R-CNN with post-processing tools to capture nucleation events in HSV data. Although this system offers precise tracking, it demands significant fine-tuning to adapt to new datasets, hindering its general applicability across diverse experimental frameworks.

Introducing deep learning architectures, such as CNNs like U-Net \cite{10.1007/978-3-319-24574-4_28} and Mask R-CNN \cite{8237584}, significantly addressed these challenges. These models captured intricate phase boundaries and dynamic behaviors in TPF images. However, even CNN-based approaches encountered limitations in PD scenarios, demanding pixel-level precision to distinguish vapor, liquid, and microlayer phases effectively. Transfer learning emerged as a promising solution, enabling models trained on analogous domains—such as biological imaging—to enhance performance on TPF datasets. Yet, the task-specific nature of these CNNs restricted their adaptability, especially under new experimental conditions or with previously unseen modalities, leaving room for further exploration in multimodal HSV PD datasets.
In response to these gaps, large vision foundation models like the Segment Anything Model (SAM) have gained traction due to their potential to generalize across diverse datasets with minimal fine-tuning. These models, built on extensive pretraining across various domains, offer a new horizon in segmentation, particularly in complex scientific applications like PD. 

\subsection{Vision Foundation Models for Segmentation}
\label{sec:vfm_methods}

Recent strides in computer vision have led to the development of large vision foundation models, such as SAM, designed to enhance generalization and performance across diverse segmentation tasks. These models surpass traditional CNN-based approaches by leveraging extensive pretraining on large-scale datasets, capturing global and local features, and excelling in learning long-range dependencies critical for complex segmentation tasks \cite{9052469}. Unlike CNNs, which primarily focus on local spatial features, models like Swin Transformer introduce hierarchical attention mechanisms that enhance feature extraction across multiple scales \cite{9710580}. Prompt-based learning frameworks like SAM offer rapid adaptation to new segmentation challenges, ensuring flexibility and robustness across varied domains with minimal fine-tuning \cite{Kirillov_2023_ICCV}. This architectural adaptability makes them superior for diverse segmentation tasks \cite{9878483}.

SAM, for example, demonstrates versatility by excelling in natural image segmentation and achieving breakthroughs in medical imaging \cite{HUANG2024103061, MAZUROWSKI2023102918, Ma2024}, remote sensing \cite{Ji2024, OSCO2023103540, 10522788}, and video tracking \cite{10436161, cheng2023segmenttrack, yang2023track}. Other prominent models include SEEM for multimodal segmentation \cite{NEURIPS2023_3ef61f7e}, Mask2Former with masked attention mechanisms for universal segmentation \cite{9878483}, HRNet with high-resolution representations \cite{9052469}, and Swin Transformer for hierarchical feature extraction \cite{9710580}. These innovations address scalability limitations in smaller CNN models, paving the way for their integration into more complex scientific tasks like phase detection and fluid dynamics.

\subsection{Existing Datasets in Boiling Phenomena}
\label{sec:boiling_datasets}

Traditional datasets, such as the Boiling Dataset, \cite{10.1115/HT2022-85582, Hu2023} focus on broader TPF phenomena, targeting tasks like classification and flow regime identification. However, these datasets fall short regarding PD tasks that demand pixel-level precision. TPF datasets often lack high-resolution frame-mask pairs, limiting their applicability in training advanced deep-learning models for precise segmentation, such as those used in HSV imagery. This gap hinders accurately segmenting vapor, liquid, and microlayer phases.

To address these challenges, we introduce a novel PD dataset tailored for HSV segmentation. This dataset bridges the gap by offering high-resolution frame-mask pairs with detailed annotations for liquid, vapor, and microlayer phases. Its open-source nature encourages further research and enables the fine-tuning of state-of-the-art models for new experimental setups and various fluid types. Our contribution advances PD segmentation and provides a benchmark that fosters innovation in both scientific and industrial applications of HSV PD segmentation.

\subsection{Uncertainty Quantification in Boiling Phenomena Segmentation}
\label{sec:uncertainty_quantification}

UQ is essential for ensuring accurate and reliable measurements in scientific experiments, particularly in complex systems such as HSV PD tasks. In this domain, the precision of segmented outputs, including parameters like contact line density and dry area fraction, directly impacts the understanding of fundamental mechanisms. However, traditional segmentation approaches often need to pay more attention to variability and discretization errors inherent in pixel-based measurements, limiting their reliability under different experimental conditions. UQ methods provide a systematic way to address these issues by evaluating and mitigating the uncertainties associated with such measurements.

Several studies emphasize the necessity of robust UQ techniques in fluid dynamics. Ravichandran et al. \cite{RAVICHANDRAN2023110879} explored UQ in the context of deep learning models for boiling heat transfer, highlighting how uncertainties in segmented data can skew estimations of thermal performance, ultimately influencing system-level insights. Seong et al. \cite{SEONG2023104336} further reinforced the importance of UQ, particularly in HSV datasets, where overlapping objects like bubbles pose significant challenges to segmentation accuracy. Without proper uncertainty estimation, these models are prone to delivering erroneous conclusions, especially in PD tasks that require pixel-level precision. These findings align with earlier works that called for greater emphasis on UQ to enhance robustness in medical image segmentation \cite{10.1007/978-3-030-59861-7_9, MUNIA2025102719}, radiotherapy \cite{WAHID2024110542}, and bubble segmentation applications \cite{LIU20191096}.

Our study introduces a PD-focused UQ framework, building upon these prior efforts to address the specific challenges HSV PD segmentation poses. Unlike general-purpose UQ methodologies, our approach leverages advanced tools to quantify how grid resolution and bubble size influence segmentation outcomes. This approach offers enhanced precision and ensures more reliable estimations, even under varying heat flux and experimental conditions. The framework identifies how resolution-dependent errors propagate across different modalities by comparing theoretical and discretized values of key metrics like area and perimeter. The weighted frequency analysis further refines the error quantification by assessing how errors vary with bubble size-frequency distributions. This approach offers enhanced precision and ensures more reliable estimations, even under varying heat flux and experimental conditions.

Our UQ framework represents a step forward by integrating simulation-based and real-world data validations, bridging gaps identified in existing studies. In particular, the framework’s ability to adjust error estimates based on real experimental measurements distinguishes it from prior works that relied solely on theoretical metrics. Moreover, the open-source nature of our UQ tools encourages further research and collaboration, addressing the scarcity of specialized resources for PD segmentation. By enhancing the reliability of boiling heat transfer analysis, our method sets a foundation for future explorations, fostering more accurate scientific insights and driving advancements in phase detection research.

\section{Methodology}
\subsection{Model Architecture}
\label{sec:model_architecture}
Our hybrid approach begins with U-Net, a well-established encoder-decoder architecture. The U-Net is designed to capture hierarchical spatial information through a symmetric structure consisting of a downsampling (encoder) path and an upsampling (decoder) path. The encoder applies multiple convolutional layers with ReLU activations and max-pooling operations to extract low-level and high-level features. Mathematically, the output of a convolutional layer in the encoder is given by:

\begin{equation}
f_{ij}^l = \sigma \left( \sum_{m,n} W_{mn}^l \cdot X_{i+m, j+n}^{(l-1)} + b^l \right)
\end{equation}

where \( f_{ij}^l \) is the feature map at layer \( l \), \( \sigma \) represents the ReLU activation function, \( W^l \) are the convolutional weights, \( X^{(l-1)} \) is the input feature map from the previous layer, and \( b^l \) is the bias term. Max pooling reduces the spatial dimensions, allowing the network to aggregate contextual information effectively:

\begin{equation}
p(x) = \max_{i,j \in N(x)} X_{ij}
\end{equation}

This aggregation improves the receptive field without increasing computational cost. The decoder mirrors the encoder’s structure, employing transposed convolutions for upsampling. The output at each upsampling step is mathematically expressed as:

\begin{equation}
g_{ij}^l = \sum_{m,n} W_{mn}^l \cdot h_{i-m, j-n}^{(l-1)}
\end{equation}

where \( h^{(l-1)} \) denotes the input feature map from the preceding layer. The decoder also incorporates skip connections, which concatenate corresponding feature maps from the encoder to enrich the upsampled output with high-level contextual and low-level spatial information.

Given the task-specific nature of U-Net and the distinct characteristics of PD data for different fluids, the model undergoes transfer learning. Pre-trained weights from cellular segmentation tasks are fine-tuned on fluid-specific data modalities (Argon, Nitrogen, FC-72, Water) to adapt to the unique visual patterns of boiling phenomena. This is achieved by freezing the initial layers of the U-Net and fine-tuning deeper layers. The optimization process during fine-tuning is governed by the following:

\begin{equation}
W^*, b^* = \arg \min_{W,b} L(Y, \hat{Y})
\end{equation}

where \( L(Y, \hat{Y}) \) is the cross-entropy loss between the ground truth labels \( Y \) and the predicted outputs \( \hat{Y} \). This segmentation phase uses a binary cross-entropy (BCE) loss, particularly effective for cases like phase detection with binary masks:

\begin{equation}
L_{BCE} = -\frac{1}{N} \sum_{i=1}^N \left[ y_i \log(\hat{y}_i) + (1 - y_i) \log(1 - \hat{y}_i) \right]
\label{eq:bce_loss}
\end{equation}

where \( N \) is the total number of pixels, \( y_i \) is the ground truth label, and \( \hat{y}_i \) is the predicted probability for pixel \( i \).

\begin{figure}[ht!]
    \centering
    \includegraphics[width=0.5\textwidth]{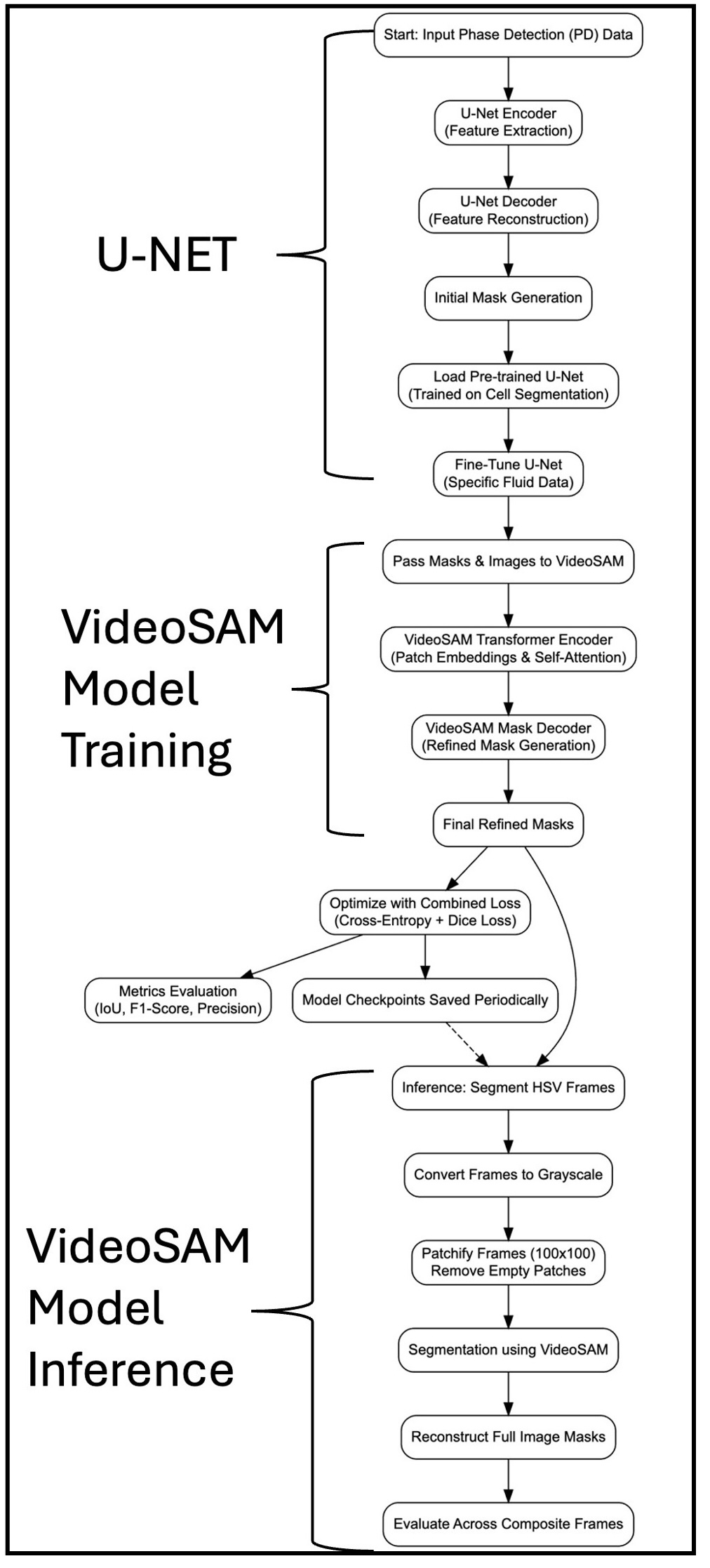}
    \caption{Integrated Segmentation Workflow for Phase Detection Data Using Hybrid U-Net and VideoSAM Models.}
    \label{fig:workflow}
\end{figure}

Once the U-Net generates preliminary segmentation masks, these are passed into the SAM-based VideoSAM framework for refinement as illustrated in Figure~\ref{fig:workflow}. VideoSAM integrates a transformer-based image encoder with a mask decoder, leveraging the pre-trained weights from the SAM architecture. The image encoder maps each pixel into a latent space, capturing intricate relationships among pixels, while the mask decoder refines segmentation boundaries through masked attention. The transformer encoder processes the input image as a sequence of patches, each represented by a fixed-dimensional vector:

\begin{equation}
z_0 = [X_{\text{patch}}^1 E; X_{\text{patch}}^2 E; \ldots; X_{\text{patch}}^N E] + E_{\text{pos}}
\end{equation}

where \( E \) is the embedding matrix and \( E_{\text{pos}} \) encodes positional information. These embeddings are passed through multiple transformer layers, each composed of multi-head self-attention (MHSA) and feed-forward networks (FFN):

\begin{equation}
MHSA(Q, K, V) = \text{softmax} \left( \frac{QK^T}{\sqrt{d}} \right) V
\end{equation}

Here, \( Q \), \( K \), and \( V \) are the query, key, and value matrices, respectively, and \( d \) is the dimensionality of the embeddings. The output from the transformer is fed into the mask decoder to produce the final refined segmentation.

The loss function for VideoSAM incorporates both the cross-entropy loss and the Dice loss to optimize boundary alignment and overlap:

\begin{equation}
L_{Dice} = 1 - \frac{2 \sum_i y_i \hat{y}_i}{\sum_i y_i + \sum_i \hat{y}_i}
\end{equation}

The combined loss function is:

\begin{equation}
L = L_{BCE} + L_{Dice}
\end{equation}

This dual optimization ensures that VideoSAM achieves accurate segmentation and aligns well with the nuanced spatial distribution of bubbles in PD data.

\subsection{Data Collection and Processing}
\label{sec:data_processing}
To ensure comprehensive coverage of boiling dynamics, 250 random frames were sampled from each modality, amounting to a dataset of 1,000 frames for this study. The dataset was partitioned into 80:20 training-validation splits to maintain diversity across modalities. Although this work focuses on frame-level segmentation, the model architecture allows future expansion into temporal dynamics to capture bubble evolution across sequences.

The raw HSV frames were converted to grayscale and normalized to enhance feature visibility by subtracting reference frames and adjusting contrast, minimizing background noise for improved segmentation performance. Ground truth segmentation masks were developed through a semi-automated pipeline that combined U-Net-based initial segmentation with expert manual refinement using ImageJ software \cite{Schindelin2012} and adaptive thresholding algorithms \cite{CHAVAGNAT2021121294}. These preliminary masks were refined to meet accuracy standards while minimizing manual labor.

To enhance model performance and manage memory efficiently, the images and masks were patchified into 100x100 pixel grids, with patches without mask data discarded. The remaining patches were resized to 256x256 pixels and normalized to binary values (0 or 1) to ensure compatibility with the model.

This meticulous data collection and processing pipeline ensures the robustness and generalizability of the VideoSAM model. It facilitates high-accuracy segmentation across different boiling conditions. Combining multimodal datasets, efficient annotation strategies, and rigorous preprocessing guarantees that the resulting model can handle simple and complex fluid behaviors. This framework emphasizes adaptability and precision and is a scalable solution for future HSV studies.

\subsection{U-Net CNN}
\label{sec:unet_cnn}
\subsubsection{Model Architecture}

The U-Net CNN architecture is fundamental in \segframework{}, particularly for segmenting bubble footprints in HSV PD. Its encoder-decoder structure allows the model to capture local and global spatial features, making it well-suited for segmenting complex patterns in boiling phenomena. This section details the architecture, transfer learning strategies, and the preprocessing pipeline for adapting U-Net to multiple boiling fluids.

The U-Net architecture consists of two primary components: the contracting (encoder) path and the expanding (decoder) path. The encoder reduces the spatial dimensions through convolutional operations followed by max-pooling, extracting increasingly abstract features. The mathematical operation of each convolutional layer is defined by Eq.~(\ref{eq:conv_layer}). 

The decoder path restores the spatial resolution through up-convolutions (deconvolutions) and feature concatenation from the encoder. This concatenation step integrates high-level contextual information with localized details, ensuring precise segmentation of bubble boundaries:
\begin{equation}
H_{\text{cat}}^l = [H_{\text{enc}}^l; G_{\text{dec}}^l]
\label{eq:conv_layer}
\end{equation}
where \( H_{\text{enc}}^l \) and \( G_{\text{dec}}^l \) are the encoder and decoder feature maps, respectively. This combination improves the model’s capacity to handle fine-grained segmentation, especially in phase-detection applications where precise identification of bubble interfaces is crucial.

\subsubsection{Transfer Learning}

Transfer learning enhanced U-Net’s adaptability, leveraging a pre-trained model from a biological segmentation dataset. This approach involved freezing the encoder layers, which capture general features such as edges, corners, and textures, while fine-tuning the deeper decoder layers to specialize them for the specific PD segmentation task.

Several critical hyperparameters governed the fine-tuning process to ensure optimal performance. The learning rate \( \eta \) was set within a range of \(10^{-3}\) to \(10^{-4}\), selected through empirical testing to balance convergence speed with update stability. A small batch size of 8 accommodates the high memory demands of high-resolution images and ensures smoother gradient updates. The number of epochs was adjusted dynamically, often ranging between 20 to 50, depending on the complexity and variability of the fluid modality under study.

The weight updates during fine-tuning were performed using the Adam optimizer without weight decay to preserve learned features while making small adjustments, defined mathematically as:
\begin{equation}
\Delta W_l = -\eta \frac{\partial L_{\text{BCE}}}{\partial W_l}, \quad \Delta B_l = -\eta \frac{\partial L_{\text{BCE}}}{\partial B_l}
\label{eq:weight_updates}
\end{equation}
where \( L_{\text{BCE}} \) is the binary cross-entropy loss for pixel-wise classification as defined in Eq.~(\ref{eq:bce_loss}). 

Custom U-Net models were trained for each modality since the dataset spans multiple modalities, including argon, nitrogen, FC-72, and high-pressure water. The poor generalization of a universal model across all modalities drove this decision. Each model was fine-tuned using transfer learning on a small set of annotated frames, typically three for training and two for validation.

\subsubsection{Feature Transfer and Data Annotation Pipeline}

\begin{figure}[ht!]
    \centering
    \includegraphics[width=\linewidth]{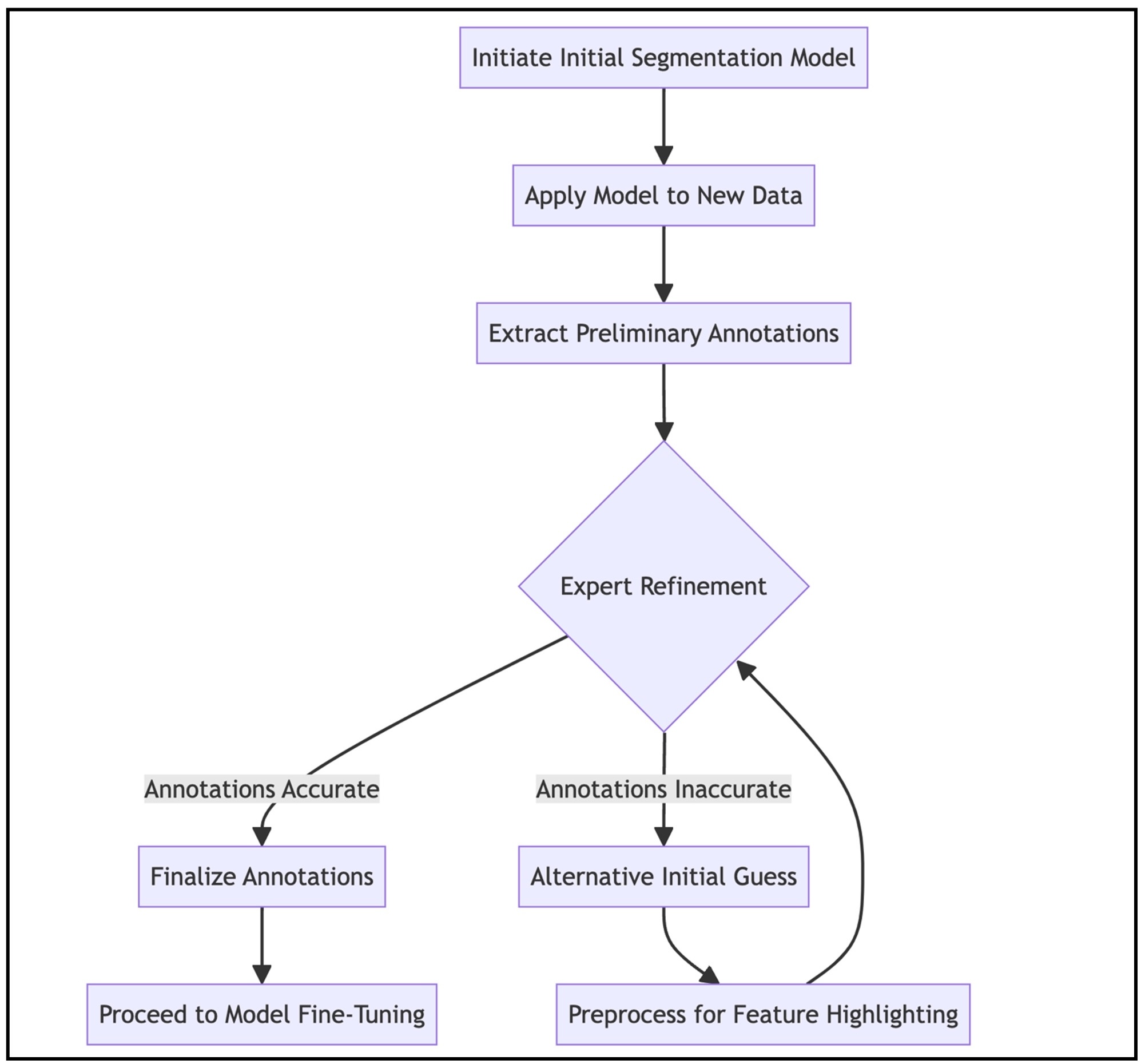} % Replace with actual image path
    \caption{Feature Transfer to Facilitate Creating Annotations for Transfer Learning.}
    \label{fig:feature_transfer}
\end{figure}

We present the feature transfer and annotation pipeline to streamline the manual annotation process, significantly reducing time and effort. This process builds on pre-trained U-Net models, which serve as the backbone for initial segmentation and employ feature transfer techniques for refining these annotations. As shown in Table~\ref{tab:feature_transfer} and visualized in Figure~\ref{fig:feature_transfer}, this structured algorithm integrates zero-shot segmentation, iterative refinement, and feature enhancement to prepare high-quality data for training the U-Net models on PD tasks.

The algorithm begins with zero-shot segmentation using a pre-trained U-Net model to generate preliminary masks. Domain experts refine these masks to ensure precision, followed by feature highlighting through preprocessing techniques like contrast normalization. In cases where the initial model fails, basic image processing techniques are employed to create alternative guesses. This comprehensive process ensures that annotations align with the physical phenomena of the boiling experiments, providing a robust foundation for model fine-tuning.

\begin{table}[ht!]
\centering
\caption{Feature Transfer and Annotation Algorithm}
\label{tab:feature_transfer}
\begin{tabular}{|c|l|p{7cm}|}
\hline
\textbf{Step} & \textbf{Action} & \textbf{Description} \\
\hline
1 & Initiate Zero-Shot Segmentation & Load a pre-existing U-Net model trained on a relevant segmentation task to generate preliminary masks. \\
\hline
2 & Apply the Model to New Data & Use the pre-trained U-Net to perform initial segmentation on HSV frames from PD experiments. \\
\hline
3 & Extract Preliminary Annotations & Utilize the “mask” Fiji macro to extract ROI features from the predicted masks and transfer them to the target HSV frames. \\
\hline
4 & Expert Refinement & Domain experts review and refine the preliminary annotations to ensure accuracy and completeness. \\
\hline
5 & Alternative Initial Guess (if needed) & When initial segmentation is inaccurate, apply basic image processing techniques to create initial guesses for ROI regions. \\
\hline
6 & Preprocess for Feature Highlighting & Normalizing images by subtracting reference frames and adjusting the contrast to enhance feature visibility. \\
\hline
7 & Finalize Annotations & Experts finalize the refined annotations to be used in the fine-tuning process. \\
\hline
8 & Proceed to Model Fine-Tuning & Use the annotated samples to fine-tune the U-Net model for optimized performance in PD segmentation. \\
\hline
\end{tabular}
\end{table}

\subsection{VideoSAM Model}
\label{sec:videosam_model}

\subsubsection{Model Architecture}
VideoSAM is built on the SAM architecture, extending its design to address the specific challenges of HSV PD segmentation. This hybrid model leverages CNNs and transformer-based components to enhance segmentation accuracy for PD tasks involving complex bubble dynamics across diverse modalities. The core architecture of VideoSAM integrates a U-Net CNN for initial mask generation with the Vision Transformer (ViT-H) backbone from SAM. This configuration combines CNNs' detailed feature extraction with the transformers' long-range dependency-capturing ability.

The Image Encoder in VideoSAM, based on the ViT-H model, processes each video frame by dividing it into fixed-size patches. These patches are flattened and transformed into vectors through a linear embedding layer, providing a structured input for the transformer. The ViT-H encoder then applies self-attention mechanisms to these embeddings, allowing the model to prioritize relevant image regions adaptively, which is essential in capturing complex fluid interfaces. This transformer-based encoder excels at capturing global context and fine-grained details across varied boiling phenomena, such as overlapping bubbles and dynamic phase interactions, which are challenging for traditional CNN-based encoders alone.

The Prompt Encoder in VideoSAM is pivotal in adapting the model to various segmentation needs by accommodating user-defined prompts such as points, bounding boxes, and text inputs. This flexibility enables VideoSAM to focus on specific regions of interest (ROIs) across frames in high-speed video data, where the segmentation targets may vary significantly. During training, bounding box prompts are derived directly from ground truth masks, allowing the model to learn detailed segmentation of bubble regions. When ground truth prompts are unavailable, such as during inference, VideoSAM employs a grid-based bounding box strategy to ensure comprehensive image coverage, ensuring that all potential segmentation targets are addressed. By generating compact representations of these prompts, the prompt encoder allows VideoSAM to effectively combine this guidance with image features, ensuring accurate and refined segmentation for complex and simpler fluid dynamics. This approach enhances the model’s adaptability across various modalities, improving segmentation precision in detailed and broad segmentation tasks. The following system of equations represents the bounding box and patch generation process:

\begin{equation}
x_{\text{min}} = \min \{ x | \text{mask}(x, y) > 0 \} 
\end{equation}
\begin{equation}
x_{\text{max}} = \max \{ x | \text{mask}(x, y) > 0 \} 
\end{equation}
\begin{equation}
y_{\text{min}} = \min \{ y | \text{mask}(x, y) > 0 \} 
\end{equation}
\begin{equation}
y_{\text{max}} = \max \{ y | \text{mask}(x, y) > 0 \} 
\end{equation}
\begin{equation}
\text{patch}(i,j) = \text{image}[(i:i+P),(j:j+P)]
\end{equation}

where $x_{\text{min}}$, $x_{\text{max}}$, $y_{\text{min}}$, $y_{\text{max}}$, and $\text{patch}(i, j)$ represent the minimum x-coordinate of the bounding box, maximum x-coordinate of the bounding box, minimum y-coordinate of the bounding box, maximum y-coordinate of the bounding box, and patch extraction function, with $i$ and $j$ as starting coordinates, and $P$ as the patch size, respectively.

The Mask Decoder in VideoSAM is a lightweight transformer designed to refine segmentation masks by integrating image features from the ViT-H encoder with prompt embeddings. Unlike traditional convolutional decoders, this transformer-based component utilizes self-attention mechanisms, allowing VideoSAM to weigh different input parts dynamically. This adaptability is crucial for capturing the varied bubble sizes, contours, and complex dynamics inherent in HSV data. The self-attention framework enables the model to focus on critical regions precisely. At the same time, a feed-forward neural network layer introduces non-linearity, enhancing the model’s ability to capture intricate boundary details. The mask decoder is fine-tuned specifically for HSV PD segmentation tasks, ensuring it can effectively handle boiling bubbles' multi-scale and dynamic nature. During this process, the parameters of the vision encoder and prompt encoder remain frozen, leveraging SAM’s pre-trained feature extraction capabilities. This approach balances computational efficiency with accuracy, enabling VideoSAM to deliver pixel-level precision in real-time or near-real-time applications. It is well-suited for high-speed, dynamic experimental environments.

To address segmentation challenges further, VideoSAM employs a combined loss function that integrates Dice Coefficient Loss with Cross-Entropy Loss. This mixed loss approach ensures that VideoSAM accurately segments both foreground (bubble regions) and background areas, balancing precise boundary delineation with robustness against class imbalance. Overall, VideoSAM capitalizes on the detailed feature extraction of CNNs and SAM’s transformer-based attention mechanisms to deliver a robust and versatile solution for HSV segmentation in boiling heat transfer studies.

\begin{figure}[ht!]
    \centering
    \includegraphics[width=\linewidth]{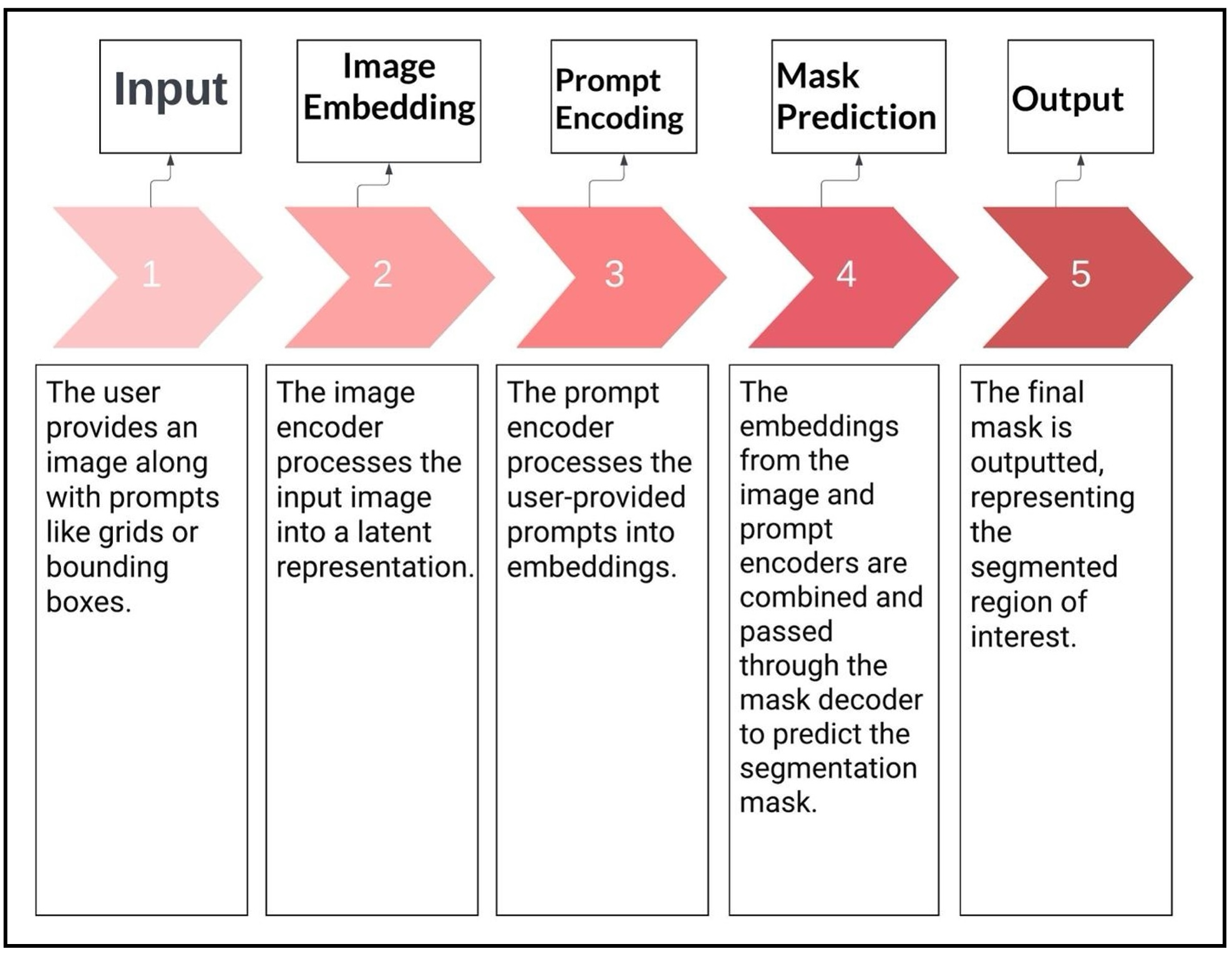} % Replace with actual image path
    \caption{Interactive Segmentation Process of VideoSAM.}
    \label{fig:videosam_workflow}
\end{figure}

The final segmentation mask is generated through a sigmoid function with a single patch probability of 0.5 for binary segmentation, producing binary masks where each patch represents the probability of belonging to the segmented region of interest. This method ensures reliable binary segmentation, balancing flexibility and efficiency across fluid environments and conditions. By freezing the image and prompt encoder layers (which retain SAM’s foundational feature extraction), VideoSAM allows the mask decoder to adapt specifically to HSV data, ensuring that the model remains versatile and finely tuned to the nuances of boiling heat transfer analysis. The overall segmentation workflow of VideoSAM is depicted in Figure~\ref{fig:videosam_workflow}.

\subsubsection{Model Training and Inference}
The development of VideoSAM focused on creating a training and inference pipeline tailored for accurate segmentation across diverse HSV datasets. During training, the pre-trained vision and prompt encoder layers from the \texttt{facebook/sam-vit-base} model were frozen to preserve foundational features. At the same time, fine-tuning was applied to the mask decoder to optimize it specifically for HSV segmentation tasks. Data handling was managed by a custom \texttt{SAMDataset} class, which generated bounding box prompts based on mask annotations, allowing for precise supervision. This data pipeline, organized within PyTorch \texttt{DataLoader} objects, ensured efficient batch processing, enhancing computational performance during training.

Training employed the Adam optimizer with a learning rate of $1 \times 10^{-5}$ and no weight decay, combined with a mixed loss function of Dice Coefficient and Cross-Entropy losses. This choice of loss functions was critical to maintaining high overlap accuracy for bubbles (Dice) while addressing class imbalance (Cross-Entropy). The \texttt{GradScaler} functionality enabled mixed precision training, optimizing training speed and memory usage. Gradient clipping was applied to stabilize training, preventing excessive updates in backpropagation. After each epoch, performance metrics, including Intersection over Union (IoU), precision, and recall, were computed on the validation set. \texttt{ReduceLROnPlateau} was implemented to dynamically adjust the learning rate based on validation loss, further refining the model’s convergence. Throughout the training, all metrics, training, and validation losses were logged, and model checkpoints were saved at regular intervals, allowing recovery from interruptions and facilitating model selection based on peak performance.

The inference pipeline evaluated VideoSAM’s segmentation capabilities across various data modalities, emphasizing mask extraction and performance assessment. High-speed video frames were converted to grayscale and normalized before segmentation to improve feature visibility and reduce noise. For spatial coverage, the \texttt{SAMInferenceDataset} class applied grid-based bounding boxes to segment images into smaller patches, which were processed independently by both VideoSAM and the base SAM model. The resulting patches were then reassembled to reconstruct full-frame segmentation masks.

For temporal consistency in dynamic environments, the inference approach extended segmentation across time, especially for composite frames or sequences, ensuring an accurate evaluation of evolving bubble structures. VideoSAM’s segmentation performance was quantified by comparing predicted masks with ground truth, using metrics like F1 Score, IoU, and precision. These evaluation metrics were aggregated across sequences, providing a comprehensive view of performance trends, including mean, minimum, maximum, and standard deviation, offering insight into VideoSAM’s reliability and robustness.

To address scenarios lacking ground truth prompts, VideoSAM employed grid-based prompting during inference. This approach generated a comprehensive grid of prompts across each frame, ensuring thorough segmentation even when specific region annotations were unavailable, thereby enhancing VideoSAM’s adaptability across varied testing scenarios. Additionally, the \texttt{MONAI} library's \texttt{DiceCELoss} function facilitated a balanced loss computation by integrating both dice and cross-entropy components, ensuring precision in segmentation, particularly for complex bubble shapes and sizes encountered in boiling videos. A detailed, multi-metric evaluation comparing VideoSAM to SAM validated the model's effectiveness, demonstrating superior segmentation performance across all key metrics, particularly in complex datasets. This rigorous testing underscored VideoSAM’s advanced capabilities, establishing it as a high-performance model for HSV PD segmentation.

\subsubsection{Experimental Setup}
This study comprises three primary experiments designed to evaluate VideoSAM's segmentation performance across multiple data modalities and benchmark its effectiveness against other models. These experiments explore the model’s generalization capabilities, performance across diverse conditions, and comparison with custom CNN-based segmentation models.

\paragraph{Experiment 1: Zero-Shot Generalization Across Modalities} 
The first experiment assessed VideoSAM’s ability to generalize to unseen data modalities, effectively testing its zero-shot learning capacity. Here, VideoSAM was trained solely on HSV PD frames of the Argon data modality. After training, the model was tested on other distinct data modalities, including Nitrogen, FC-72, and Water, which were not encountered during training. This experiment evaluated the model’s ability to segment novel data modalities without additional fine-tuning accurately. The model's performance was assessed using standard segmentation metrics, specifically Intersection over Union (IoU) and F1 Score. Visual inspections complemented quantitative assessments, enabling a nuanced understanding of segmentation quality. The results highlighted VideoSAM’s capacity to outperform the baseline SAM across various modalities, showing superior generalization without prior exposure to the new datasets.

\paragraph{Experiment 2: Performance Across Multiple Modalities}
The second experiment aimed to validate VideoSAM’s robustness in handling multiple data modalities by training the model on a combination of four datasets—Argon, Nitrogen, FC-72, and Water. This comprehensive training approach exposed VideoSAM to diverse experimental conditions, allowing it to understand modality-specific segmentation requirements better. Following training, the model was evaluated on unseen data from each modality to determine its adaptability to new samples within the trained categories. Consistent performance was expected across all datasets, particularly for complex experimental conditions. IoU and F1 Score metrics were used for quantitative assessment, and the findings demonstrated that VideoSAM maintained high segmentation accuracy across all modalities, effectively capturing the intricate dynamics of each experimental setup. VideoSAM outperformed SAM in all cases, underscoring its robustness and adaptability.

\paragraph{Experiment 3: Comparison with Custom U-Net CNN}
In the final experiment, the performance of VideoSAM was benchmarked against a custom-trained U-Net CNN, which is a well-regarded architecture for HSV TPF segmentation. U-Net, known for its effectiveness in segmenting cellular structures, was a suitable baseline due to the structural similarities between cellular segmentation and bubble footprints in HSV data. VideoSAM and U-Net CNN were trained on the same four data modalities (Argon, Nitrogen, FC-72, and Water) and evaluated on IoU and F1 Score metrics. Due to its transformer-based architecture, the hypothesis was that VideoSAM would excel in handling more complex data modalities (e.g., Argon, FC-72, and Nitrogen). At the same time, U-Net would be more effective for simpler patterns, such as those in the Water dataset. The results supported this hypothesis: VideoSAM outperformed U-Net in the complex fluid environments, capturing the intricate dynamics of each modality, while U-Net showed slightly better performance in the simpler Water dataset, emphasizing its strength in basic segmentation tasks.

\subsection{Model Evaluation}
Evaluating the model's performance requires rigorous testing beyond the training data to ensure reliability and adaptability. In this section, we assess the fine-tuned model's ability to accurately segment unseen images and validate its predictions against ground truths manually annotated by experts. This validation involved five randomly selected images from the HSV dataset, ensuring diverse bubble structures were represented. The segmented outputs were visually inspected and aligned against the original images to check for perimeter and feature alignment discrepancies. If misalignments were detected, further fine-tuning was applied, while satisfactory visual alignment allowed for deeper quantitative evaluation using key performance metrics.

\subsubsection{Boiling Performance Metrics}
\textbf{Dry Area Fraction ($\theta_{dry}$)}: Measures the proportion of the surface area unexposed to liquid, offering insights into heat transfer efficiency. It is calculated as:
\begin{equation}
    \theta_{dry} = 1 - \frac{\text{Wet Pixels}}{\text{Total Pixels}}
\end{equation}

\textbf{Contact Line Density ($\rho_{cl}$)}: Represents the extent of the interface between liquid and dry regions, which is crucial for understanding evaporation zones. It is calculated using:
\begin{equation}
    \rho_{cl} = \frac{\text{Contact Line Length}}{\text{Total Pixels}}
\end{equation}

The algorithm used for computing the boiling performance metrics is presented in Table \ref{table:boiling-metric-computation}.

\begin{table}[ht!]
\centering
\caption{Comprehensive Algorithm for Boiling Metric Computation}
\label{table:boiling-metric-computation}
\begin{tabular}{|c|l|}
\hline
\textbf{Input} & Binary mask (\texttt{binaryMask}), where 1 represents dry pixels, and 0 represents wet pixels. \\
\textbf{Output} & $\theta_{dry}$: Ratio of dry pixels to total pixels, $\rho_{cl}$: Contact line length ratio \\
\hline
1 & Read binary mask \texttt{binaryMask}. \\
2 & Compute total number of pixels: \texttt{totalPixels} $\leftarrow$ \texttt{numel(binaryMask)}. \\
3 & Calculate number of dry pixels: \texttt{dryPixels} $\leftarrow$ $\sum$(\texttt{binaryMask}). \\
4 & Compute dry pixel ratio: $\theta_{dry} \leftarrow \frac{\texttt{dryPixels}}{\texttt{totalPixels}}$. \\
5 & Generate inverted mask: \texttt{invertedBinaryMask} $\leftarrow$ $1 - \texttt{binaryMask}$. \\
6 & Apply distance transform: \texttt{distances} $\leftarrow$ \texttt{bwdist(invertedBinaryMask)}. \\
7 & Extract contact line length: \texttt{contactLineLength} $\leftarrow$ \texttt{count(distances = 1)}. \\
8 & Compute contact line length ratio: $\rho_{cl} \leftarrow \frac{\texttt{contactLineLength}}{\texttt{totalPixels}}$. \\
9 & Return $\theta_{dry}$ and $\rho_{cl}$. \\
\hline
\end{tabular}
\end{table}

\subsubsection{Machine Learning Metrics}
The segmentation quality was further evaluated using machine learning metrics derived from the confusion matrix, presented in Table \ref{table:ml-metrics}.

\begin{table}[ht!]
\centering
\caption{Machine Learning Performance Metrics and Their Mathematical Definitions}
\label{table:ml-metrics}
\begin{tabular}{|c|c|}
\hline
\textbf{Metric} & \textbf{Formula} \\
\hline
Accuracy & $A = \frac{TP + TN}{TP + TN + FP + FN}$ \\
Precision & $P = \frac{TP}{TP + FP}$ \\
Recall & $R = \frac{TP}{TP + FN}$ \\
Specificity & $S = \frac{TN}{TN + FP}$ \\
F1 Score & $F_1 = 2 \times \frac{P \times R}{P + R} = \frac{2 \times TP}{2TP + FP + FN}$ \\
IoU (Jaccard Index) & $IoU = \frac{TP}{TP + FP + FN}$ \\
Matthews Correlation Coefficient (MCC) & $MCC = \frac{(TP \times TN) - (FP \times FN)}{\sqrt{(TP + FP)(TP + FN)(TN + FP)(TN + FN)}}$ \\
\hline
\end{tabular}
\end{table}

\subsection{Uncertainty Quantification}
\label{sec:uq_boiling}
To ensure accurate estimation of boiling dynamics, particularly contact line density and dry area fraction, it is essential to quantify the uncertainty in pixel-based measurements. These measurements, though effective, inherently contain discretization errors due to the limitations of representing continuous shapes (such as bubbles) on a discrete pixel grid. To systematically address these uncertainties, we employed a series of computational techniques, including dilation and erosion operations, to handle over- and underestimations of bubble regions. This methodology simulates the effects of varying bubble radii and grid resolutions on the calculated area and perimeter.

\begin{figure}[ht!]
    \centering
    \includegraphics[width=0.8\textwidth]{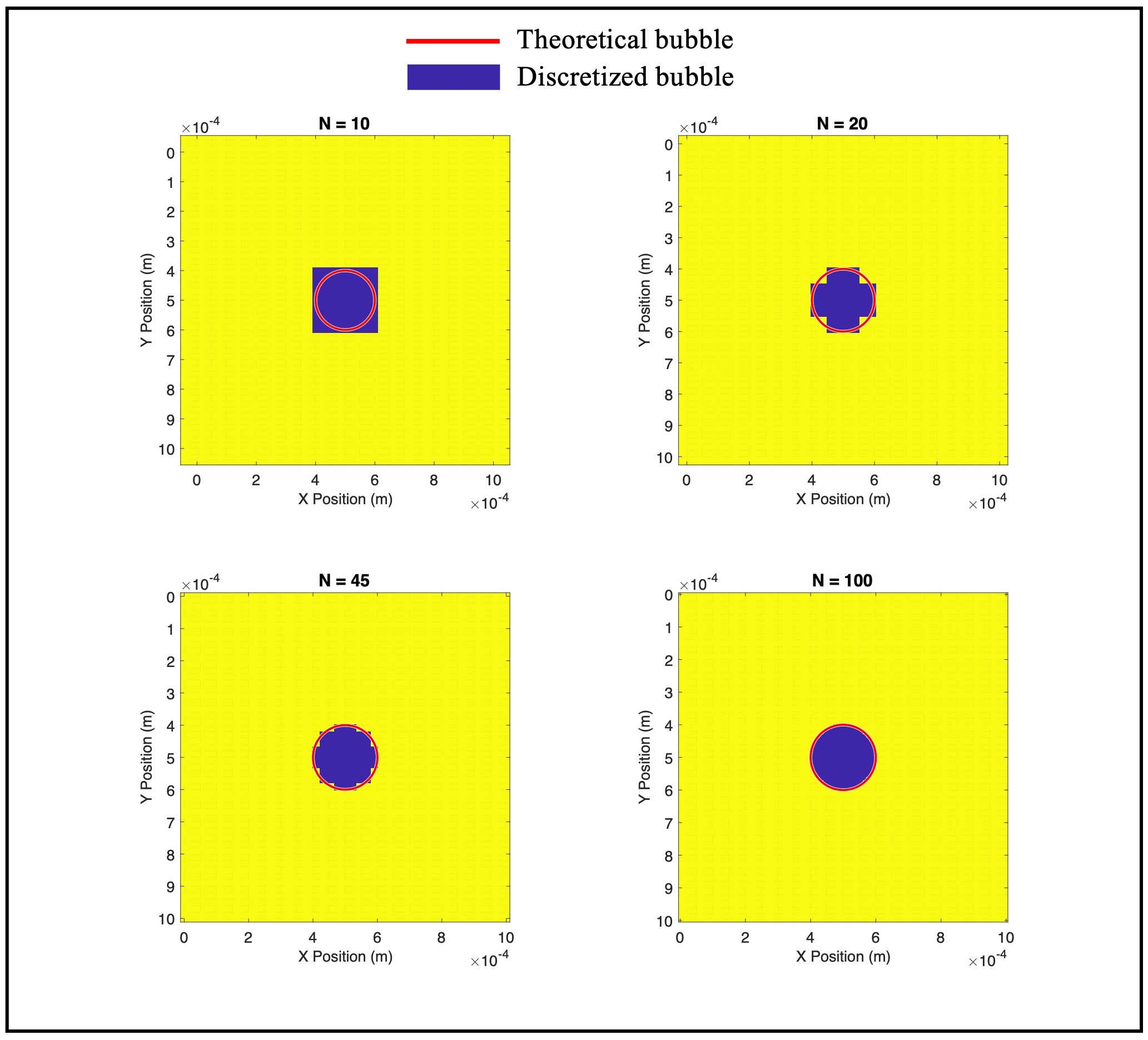}
    \caption{Theoretical and Discretized Bubble for a Fixed Bubble Radius (100 microns) and Varying Grid Resolution \(N\)}
    \label{fig:bubble_convergence}
\end{figure}

The algorithm performs iterative simulations to compare a bubble's theoretical area and perimeter (modeled as a perfect circle) against the discretized values obtained from the pixel data. Across a range of grid resolutions (denoted as \(N\)) and bubble radii \(R\), the percentage relative error (PRE) and mean error (ME) were calculated to quantify the deviations between the theoretical and numerical measurements. These metrics are critical for understanding how resolution impacts segmentation accuracy and provide insights into the trade-offs between computational efficiency and measurement precision. Figure \ref{fig:bubble_convergence} illustrates how the discretized representation of a bubble converges toward its theoretical circular shape as the grid resolution increases.

Experimental bubble data were extracted to validate these calculations, and their dimensions (radius, area, and perimeter) were measured to calibrate the error estimates. This calibration aligns the simulations with real-world data, enhancing the reliability of uncertainty estimates. A weighted frequency analysis was applied to refine error quantification further, accounting for the varying occurrence frequencies of bubbles of different sizes. Thus, a comprehensive view of the distribution of errors was provided.

The iterative simulations were repeated 500 times for each pair of grid resolution and bubble radius to ensure robust statistical significance, minimizing the impact of random bubble placements. The final error matrices generated from these simulations are a reference for correcting measurement errors under specific experimental conditions, further detailed in Table \ref{table:error-quantification-algorithm}. This rigorous approach to uncertainty quantification ensures that the segmentation process remains accurate and reliable across various data modalities, providing critical insights for computer vision applications.

\begin{table}[ht!]
\centering
\caption{Algorithm for Enhanced Bubble Discretization and Error Quantification}
\label{table:error-quantification-algorithm}
\begin{tabular}{|l|l|}
\hline
\textbf{Input} & Domain length \(L\), \texttt{NumSimulations}, \texttt{N\_values}, \texttt{R\_values} \\
\textbf{Output} & Results with PRE and ME for area and perimeter \\
\hline
1 & Initialize simulation parameters \\
2 & \textbf{for each} \(N\) \textbf{in} \texttt{N\_values} \textbf{do} \\
3 & \quad \textbf{for each} \(R\) \textbf{in} \texttt{R\_values} \textbf{do} \\
4 & \quad \quad Initialize \(A_{\text{theoretical}}\), \(P_{\text{theoretical}}\) for \(R\) \\
5 & \quad \quad \textbf{for} \(k = 1\) \textbf{to} \texttt{NumSimulations} \textbf{do} \\
6 & \quad \quad \quad Randomly position bubble within domain \\
7 & \quad \quad \quad Calculate \texttt{Dis}, define \texttt{bub} (\texttt{Dis} $<$ \(R\)) \\
8 & \quad \quad \quad Compute \(A_{\text{bub}}\), \(P_{\text{bub}}\) using pixel count \\
9 & \quad \quad \textbf{end for} \\
10 & \quad \quad Compute PRE and ME for the area, perimeter \\
11 & \quad \quad Store errors in Results \\
12 & \quad \textbf{end for} \\
13 & \textbf{end for} \\
14 & Load and process Results for visualization \\
15 & Extract error matrices from Results \\
16 & Visualize errors (Histograms, Surface plots) \\
17 & Estimate errors for experimental data: \\
18 & \quad Match experimental data with simulation parameters \\
19 & \quad Read off errors for specific experimental conditions \\
20 & Perform weighted frequency analysis on errors \\
21 & Output refined error estimations \\
\hline
\end{tabular}
\end{table}

\section{Results and Discussions}
\subsection{U-Net CNN Results}
\subsubsection{Comparative Analysis of Dry Area Fraction and Contact Line Density}
\label{sec:comparative_analysis}

In Figure \ref{fig:combined_boiling_metrics}, we present the analysis of boiling metrics derived from HSV frames of liquid nitrogen in saturated pool boiling conditions (93.5 K at 70 psia). This dataset encompasses 16 distinct heat flux levels, each containing approximately 2000 frames, resulting in a comprehensive dataset of around 32,000 frames. The metrics analyzed are the dry area fraction and contact line density, crucial indicators of boiling dynamics and heat transfer performance. These metrics were extracted using two segmentation methods: the U-Net CNN model and an adaptive thresholding technique, allowing a comparative analysis of their effectiveness in capturing boiling features under increasing thermal loads.

The top left plot illustrates the trend in dry area fraction as a function of mean heat flux. Both segmentation techniques show an upward trend, indicating that vapor covers a larger portion of the surface as the heat flux increases. This behavior aligns with pool boiling dynamics, where increased thermal input intensifies vapor formation, expanding the dry area fraction. The U-Net CNN consistently provides slightly higher values for the dry area fraction than adaptive thresholding, particularly from heat flux levels around 140-158 kW/m² onwards. This deviation suggests that the U-Net CNN captures smaller bubbles and finer vapor structures more effectively than thresholding, especially as the boiling regime transitions to more dynamic states with increased heat input.

Similarly, the bottom left plot in Figure \ref{fig:combined_boiling_metrics} shows contact line density versus mean heat flux. Both methods indicate an increase in contact line density with rising heat flux, highlighting the intensification of the liquid-vapor interface, a hallmark of enhanced boiling activity. However, the U-Net CNN consistently provides higher contact line density values, particularly at higher flux levels, underscoring its superior sensitivity to intricate details in the boiling interface. This divergence becomes prominent around heat flux steps 8-9 (approximately 140-158 kW/m²), where the boiling process introduces smaller, closely packed bubbles, challenging thresholding to detect accurately.

\subsubsection{3D Histogram Analysis of Heat Flux vs. Bubble Sizes Distribution}
\label{sec:3D_histogram}

The plots on the right side of Figure \ref{fig:combined_boiling_metrics} present a detailed 3D histogram analysis, showcasing the bivariate distribution of bubble sizes at different heat flux levels obtained using two distinct image processing techniques: segmentation with U-Net CNN and adaptive thresholding. This analysis provides insights into the boiling dynamics within liquid nitrogen, illustrating how bubble sizes evolve with increasing heat flux.

The histogram reveals a comprehensive distribution of bubble sizes across a broad range of heat fluxes for the segmentation method using the U-Net CNN. The distribution is skewed towards smaller bubble sizes at lower heat flux values. This indicates a lower bubble-size region. As the heat flux increases, the histogram shifts toward larger bubble sizes, suggesting a progression toward producing larger vapor bubbles under higher thermal conditions. This trend is typically associated with intensified boiling activity. The color gradient in the histogram emphasizes the prevalence of medium-sized bubbles, which appear most frequently, underscoring the U-Net CNN’s ability to detect a diverse range of bubble sizes accurately.

In contrast, the thresholding histogram shows a similar trend but with some differences. Although larger bubbles are prevalent at higher heat flux levels, the number of smaller bubbles appears reduced compared to the U-Net CNN results. This indicates potential limitations in thresholding for detecting finer bubbles, especially at lower heat fluxes. This results in a less nuanced distribution, particularly in the lower bubble-size regions, which may impact bubble characterization accuracy, as thresholding might overlook smaller bubbles effectively captured by the U-Net CNN.

Both histograms exhibit a peak at the higher end of the heat flux range, potentially indicating a transition to a vigorous boiling phase where larger bubbles dominate. This shift from smaller to larger bubbles as heat flux increases reflects the boiling process’s progression from nucleate to potentially film boiling, where larger vapor bubbles coalesce and cover the surface more extensively. This observation has significant implications for thermal management and cooling efficiency, marking a crucial transition in heat transfer characteristics. The U-Net CNN's more granular perspective is invaluable for designing and controlling cooling systems in various industrial applications.

\subsubsection{Comparison of Bubble Size Distribution Using Segmentation and Thresholding}
\label{sec:bubble_sizes}

\begin{figure}[ht!]
    \centering
    \includegraphics[width=0.8\textwidth]{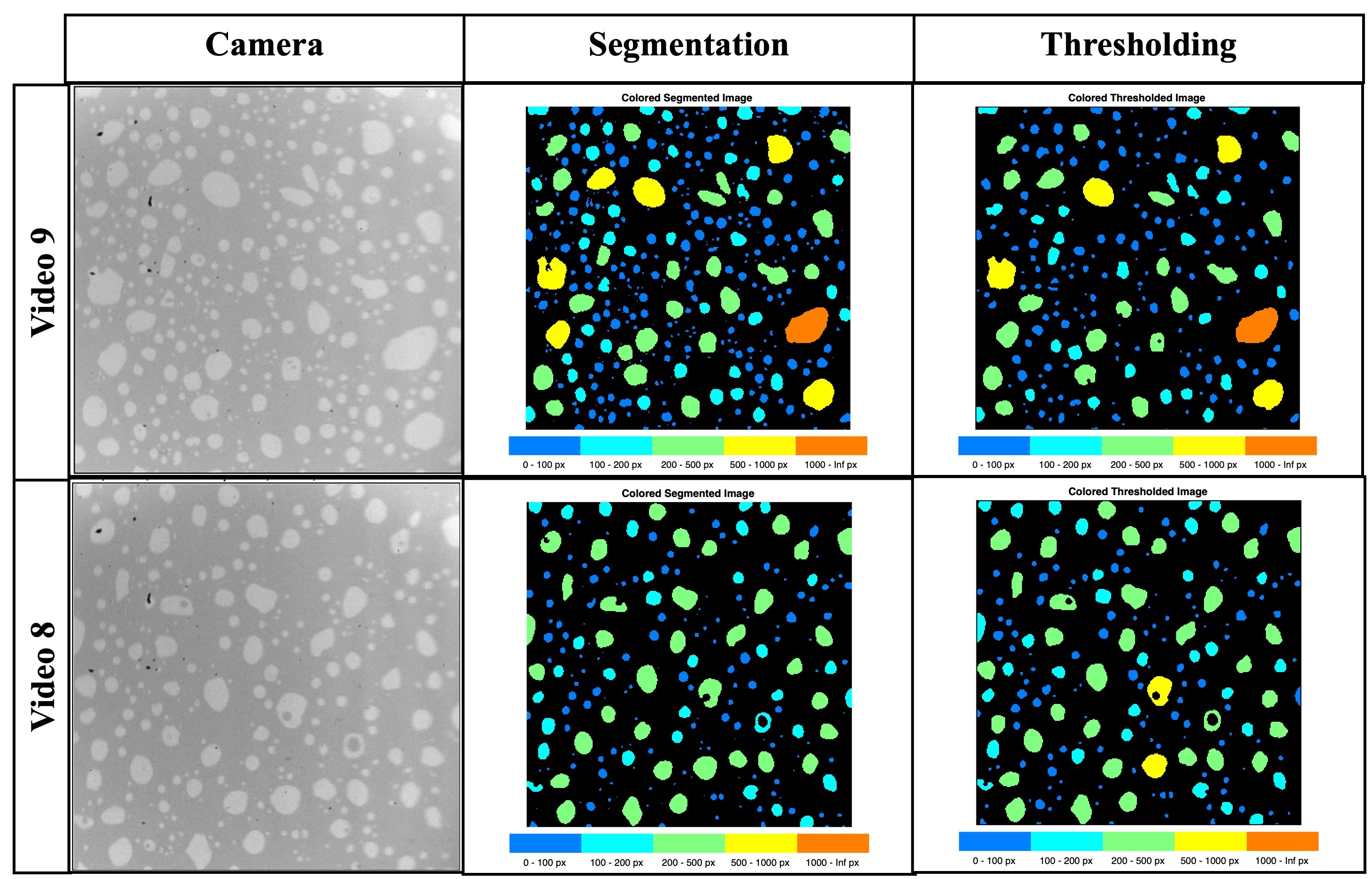}
    \caption{Visualization of Colored Bubble Size Distribution in Video 8 \& 9 Frames: Raw Camera Output vs. Post-Processing with Segmentation and Thresholding Methods}
    \label{fig:bubble_size_comparison}
\end{figure}

Figure \ref{fig:bubble_size_comparison} compares bubble size distribution in frames from videos 8 and 9, illustrating the raw camera output alongside post-processed images obtained using segmentation and thresholding methods. The leftmost column presents grayscale images from the camera, capturing a spectrum of bubble sizes in various shades but lacking a quantifiable differentiation between them. This untreated visual data serves as the baseline for understanding the effects of each processing method.

In the center column, the segmentation results are shown, utilizing a color-coded map to represent bubble sizes: blue for smaller bubbles, green for medium, yellow for larger, and orange for the largest. This clear categorization enables an immediate visual assessment of the distribution and relative size of bubbles within each frame. Both videos 8 and 9 frames reveal a wide range of bubble sizes, with a notable presence of medium to large bubbles. This distribution suggests an active boiling phase, as indicated by the prevalence of larger bubbles. Notably, the segmentation method’s precision is evident in its ability to distinguish individual bubbles even when they are closely packed, emphasizing its robustness in capturing the complete spectrum of bubble sizes.

The third column illustrates the thresholding results using color coding to differentiate sizes. However, this technique highlights larger bubbles more prominently while potentially merging or overlooking smaller bubbles, particularly those nearby. This difference suggests that while thresholding effectively identifies bubbles, it may lack the fine-grained sensitivity of the segmentation approach for capturing smaller or closely situated bubbles.

A comparative analysis of videos 8 and 9 highlights the consistency of each technique across different frames, underscoring their reliability in a series of experiments. The segmentation method’s detailed categorization is particularly valuable for applications requiring a precise understanding of bubble dynamics, which is essential for optimizing thermal processes. In contrast, while providing a broader view focused on larger bubbles, the thresholding method may be more suited for contexts with limited computational resources or where only general bubble characteristics are needed.

The integration of these methods offers a promising pathway for refined analysis. Initially, the thresholding technique could generate annotated samples, which can be fine-tuned through manual adjustments to include smaller bubbles or overlooked features. These refined samples can serve as training data to enhance the accuracy of the U-Net CNN model for improved segmentation outcomes. By visualizing bubble size distribution through these methods, the figure enriches the qualitative understanding of boiling patterns. It lays a foundation for more in-depth quantitative analyses, with potential applications extending to other fluid and boiling conditions in future research.

\subsubsection{Ground Truth Benchmark}
\label{sec:ground_benchmark}

Standardizing ground truth is critical for fine-tuning the U-Net CNN model and validating its segmentation masks, as it provides a benchmark for comparison. This study established ground truth by sending the same frames to five independent users with domain expertise in HSV PD analysis. Each user manually annotated the images, delineating bubbles from the background based on their interpretation. These annotations were then compared for dry area fraction and contact line density to understand potential variations.

\begin{figure}[ht!]
    \centering
    \includegraphics[width=0.8\textwidth]{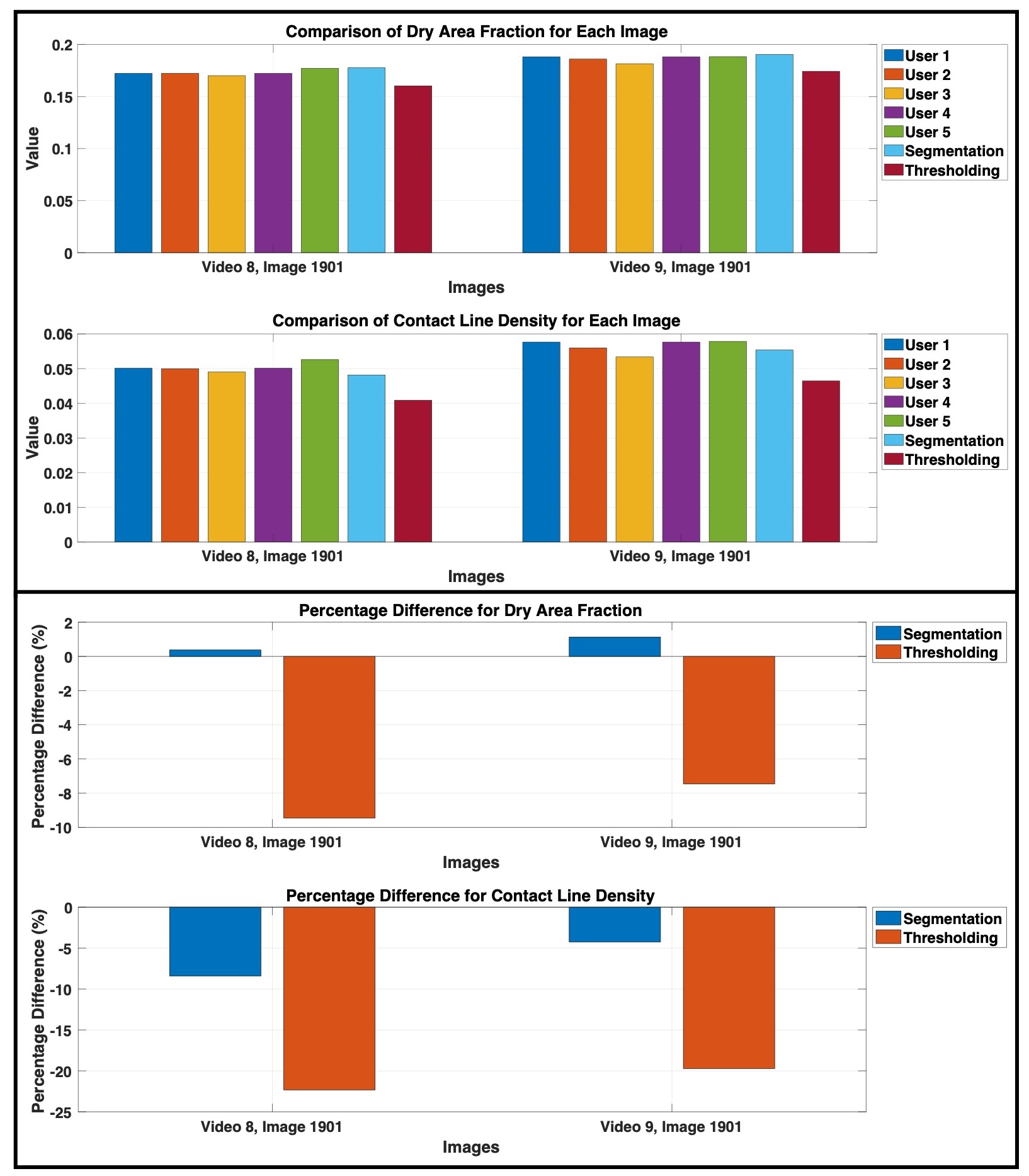}
    \caption{Comparison of Dry Area Fraction and Contact Line Density Across Different Analysts and Image Processing Techniques for Target Frames in Videos 8 \& 9.}
    \label{fig:ground_truth_benchmark}
\end{figure}

The top plot in Figure \ref{fig:ground_truth_benchmark} illustrates the variation in dry area fraction and contact line density as measured by different users and image processing techniques for frames selected from videos 8 and 9. This comparison is vital to gauge the effect of subjective human analysis on the selection of ground truth, especially in image processing contexts where objectivity is often challenging. The bar chart segments data for the two selected frames, showing values recorded by each of the five users (User 1 through User 5) alongside the results from two computational methods: U-Net-based segmentation and thresholding (as implemented by Chavagnat et al. \cite{CHAVAGNAT2021121294}). The low variability among the user-derived values highlights minor perceptual inconsistencies or interpretational differences. However, all user annotations cluster around an average of approximately 0.17 for dry area and 0.05 for contact line density. Despite subtle individual variations, this consistency underscores a general alignment in human perception of these metrics.

The computational methods, however, reveal distinctions. The segmentation method consistently yields higher values for dry area fraction than thresholding, suggesting greater sensitivity in detecting smaller dry areas that the thresholding approach may overlook. Thresholding tends to return lower values, potentially indicating a more conservative estimation that might overlook fine details in the image. This observation supports previous findings that segmentation provides a more comprehensive assessment of bubble characteristics.

Meanwhile, the bottom plot in Figure \ref{fig:ground_truth_benchmark} further quantifies the percentage difference between the computational methods and user-provided ground truth. In selected samples, the thresholding method shows a considerable underestimation, with errors reaching up to -20\% for contact line density and -10\% for dry area fraction. This error likely arises from thresholding’s tendency to miss smaller bubbles, leading to an overall underestimation of key metrics. In contrast, segmentation demonstrates much lower error rates, with deviations of less than 2\% for dry area fraction and under 8\% for contact line density.

Additionally, we compared the mean and standard deviation of user annotations with the computational results to validate the reliability of segmentation. For instance, in the case of contact line density in Video 9, Image 1901, where the highest discrepancy with thresholding was observed, the user-measured mean was 0.05649 with a standard deviation of 0.00171. The segmentation result of 0.05537 falls within one standard deviation of the user measurements, highlighting its close alignment with human observations and supporting its robustness as a reliable image-processing tool.

These results affirm that segmentation offers a closer approximation to human annotation, making it an ideal candidate for applications requiring precision in bubble dynamics analysis. Future work will focus on quantifying uncertainties in pixel-based dry area and contact line density calculations to further refine these measurements.

\subsubsection{Perimeter Visualization for Multimodal Segmentation Accuracy}
\label{sec:perimeter_visuals}

\begin{figure}[ht!]
    \centering
    \includegraphics[width=\textwidth]{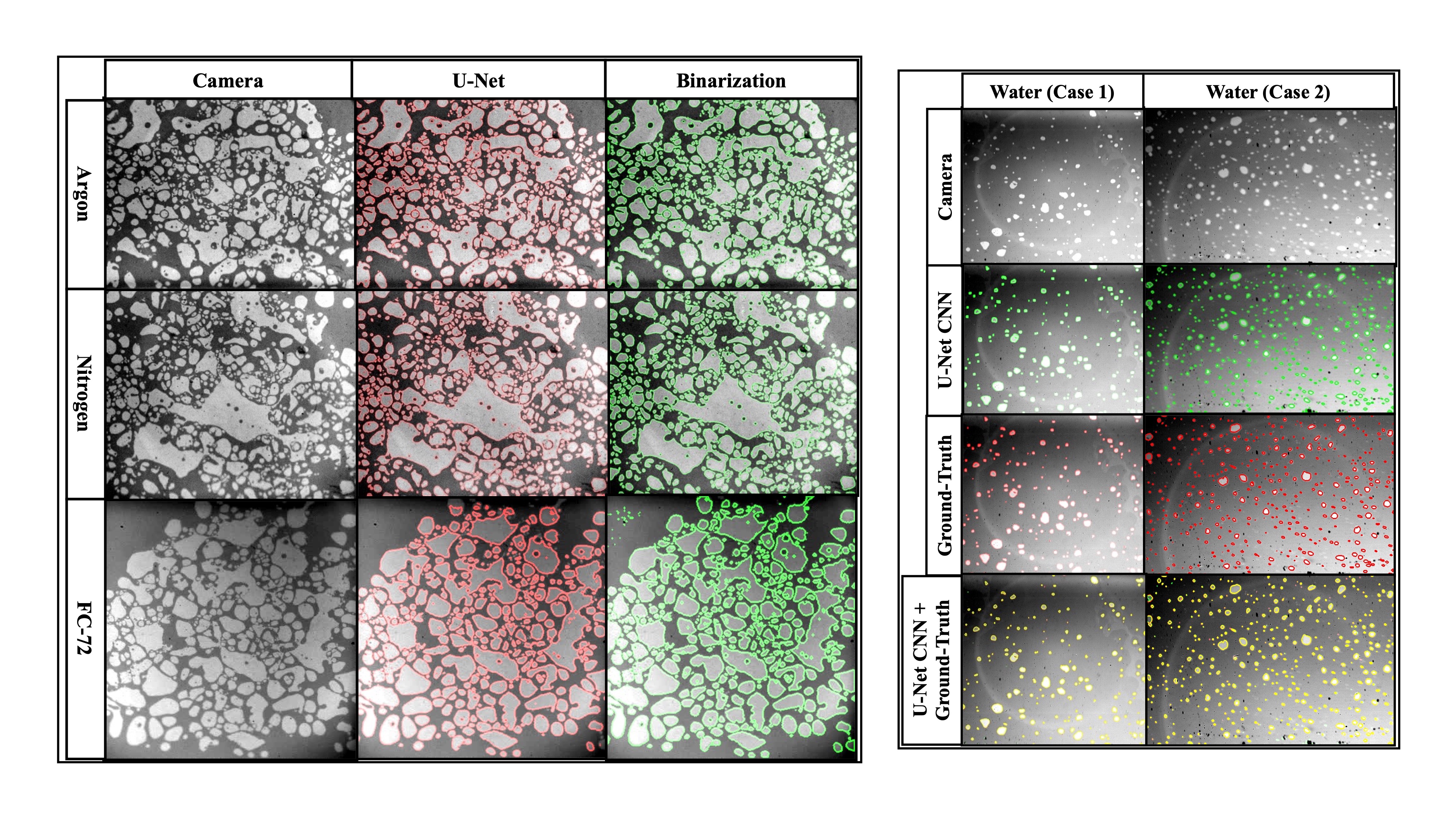}
    \caption{Perimeter Visualization in Nitrogen, Argon, FC-72, and High-Pressure Water Modalities: Raw Camera Images and Results from U-Net and Binarization Processing.}
    \label{fig:perimeter_visualization}
\end{figure}

In the left plot of Figure \ref{fig:perimeter_visualization}, we analyze segmentation results across three distinct modalities—argon, nitrogen, and FC-72—by comparing the U-Net CNN and the binarization-based segmentation methods against the original camera images. The U-Net segmentation results are highlighted with red perimeters, while the binarization (thresholding) algorithm results, developed by the MIT Red Lab, are outlined in green. Argon and nitrogen exhibit similar bubble distribution patterns, which informed the strategy to combine data from both fluids to fine-tune the U-Net model, enhancing its adaptability. This joint model performs well, with close alignment between the U-Net and binarization perimeters, especially due to the relatively lower prevalence of small bubbles in these fluids. This finding reinforces the robustness of both U-Net and binarization algorithms in segmenting fluids with a simpler bubble profile.

The FC-72 modality, however, presents a more challenging case for segmentation, with a noticeably higher concentration of small bubbles. Here, the binarization method struggles, particularly in the upper-left corner, resulting in increased false positives and missed bubble boundaries. This highlights the U-Net model’s superior sensitivity in detecting finer details in complex boiling patterns. It makes it more effective for applications involving fluids with high bubble density and varied bubble sizes. Such differences between segmentation methods could impact subsequent dry area fraction and contact line density measurements, as the segmentation accuracy directly influences these metrics.

Meanwhile, plots on the right-hand side of Figure \ref{fig:perimeter_visualization} illustrate the segmentation outcomes under two high-pressure water modalities: Case 1 at 10 bar with a heat flux of 3000 kW/m² and Case 2 at 40 bar with 3400 kW/m². The images show distinct bubble size and count variations between cases, with Case 2 exhibiting a denser and more intricate bubble distribution due to its elevated pressure and heat flux.

Initially, U-Net CNN models pre-trained on liquid nitrogen data were applied to these water-boiling images. While the model adequately identified smaller bubbles, it faced challenges with larger bubbles, often leaving hollow centers or producing false positives. This issue underscores the importance of fine-tuning U-Net models with data that closely matches the specific properties of the target dataset. Despite these limitations, the preliminary results provided a valuable foundation for generating training data through manual correction, further improving the model’s adaptability to high-pressure water boiling conditions.

After fine-tuning, the enhanced U-Net model was compared to expert-verified ground truth perimeters, as shown in the combined visualization. The yellow outlines indicate regions of overlap between the model’s predictions and expert annotations, demonstrating that the fine-tuned U-Net model achieved high accuracy and closely matched human-generated segmentation. This iterative process underscores the adaptability of U-Net CNNs in refining segmentation quality for complex fluids like high-pressure water, highlighting their potential for reliable and detailed analysis in industrial applications.

\subsubsection{Statistical Comparison of Segmentation Metrics Across Modalities}
\label{sec:stat_comparison}

\begin{figure}[ht!]
    \centering
    \includegraphics[width=\textwidth]{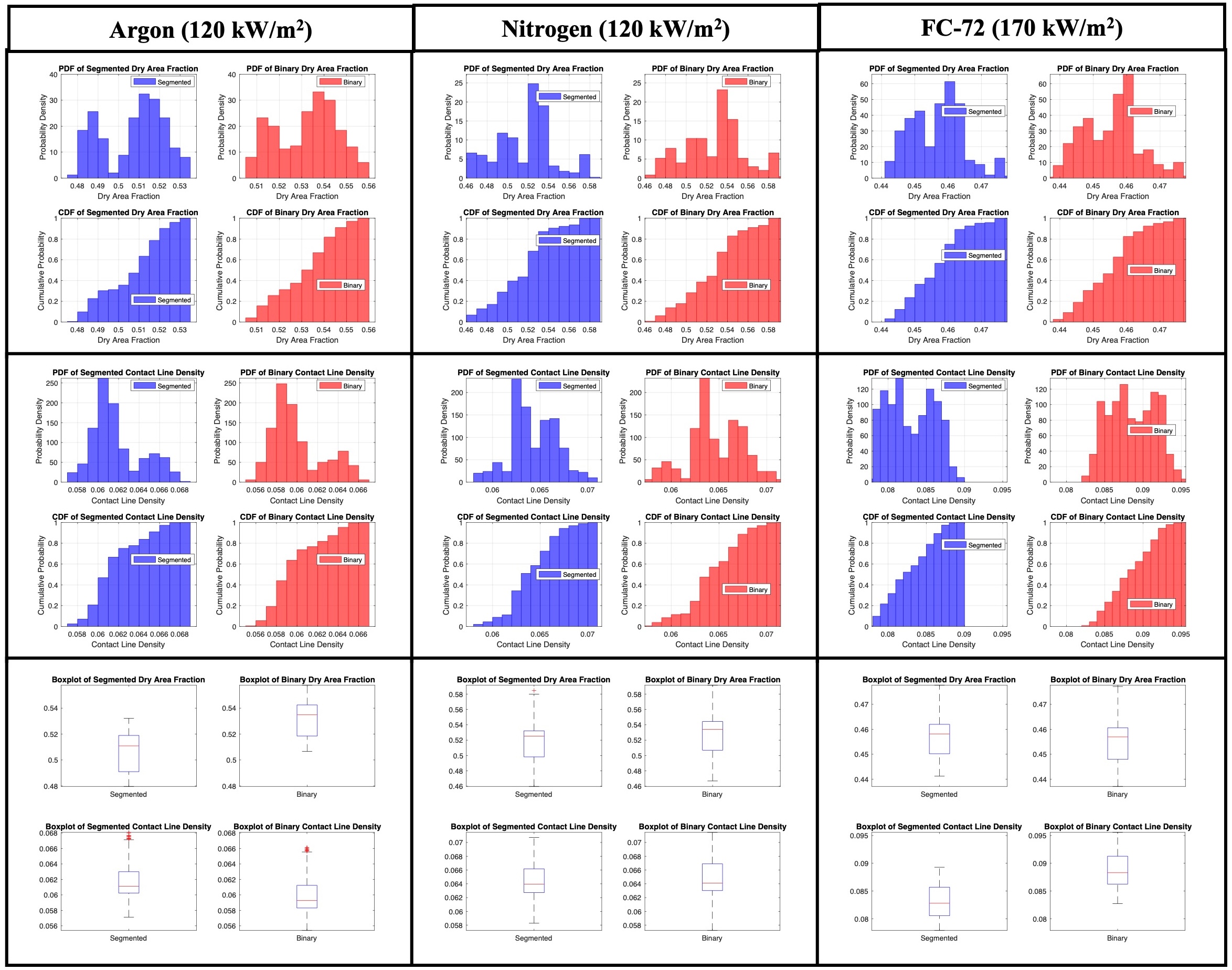}
    \caption{Statistical Analysis of Dry Area Fraction and Contact Line Density: Probability Density Functions (PDFs), Cumulative Distribution Functions (CDFs), and Box Plots for Argon, Nitrogen, and FC-72 Modalities.}
    \label{fig:statistical_analysis}
\end{figure}

Figure \ref{fig:statistical_analysis} illustrates the statistical comparison of dry area fraction and contact line density metrics for argon, nitrogen, and FC-72 modalities, derived from U-Net CNN segmentation and a binarization method. This comparison is visualized using probability density functions (PDFs), cumulative distribution functions (CDFs), and box plots, comprehensively evaluating each modality's segmentation characteristics.

In the argon and nitrogen modalities, a high consistency is observed between the U-Net and binarization methods regarding dry area fraction and contact line density. Specifically, nitrogen shows minimal deviation, with dry area fractions from segmentation ranging between 0.46 and 0.48 and contact line densities from 0.06 to 0.07. Similarly, the PDFs and CDFs for nitrogen suggest close alignment, with both segmentation techniques yielding values within a narrow band. Argon exhibits slightly more variation, with the segmented dry area fraction spanning 0.48 to 0.53 and binarization results showing a broader range from 0.51 to 0.56. Box plots further highlight this similarity, with mean values for dry area and contact line density metrics from both segmentation methods agreeing within ±5\%, indicating consistent performance across these metrics.

In contrast, the FC-72 modality reveals more pronounced discrepancies between the two segmentation methods. Although the x-axis range for dry area and contact line density remains relatively similar in the PDF and CDF plots, the probability densities and cumulative distributions indicate divergence, particularly for contact line density. The U-Net method's box plots for FC-72 suggest closer clustering around the median values, whereas the binarization method displays a broader spread with higher incidences of outliers. This difference is consistent with previous findings, where the binarization method produced more false positives, especially in complex modalities where smaller bubbles are prevalent.

Overall, this statistical analysis demonstrates the robustness of the U-Net CNN segmentation approach across different modalities. It also highlights the limitations of binarization, particularly in more intricate modalities such as FC-72, where fine details are crucial for accurate contact line and dry area measurements. This comparison reinforces the value of U-Net segmentation for high-fidelity multimodal segmentation in HSV PD research.

\subsubsection{Performance of Segmentation Metrics Across Modalities}
\label{sec:quantitative_results}

\begin{figure}[ht!]
    \centering
    \includegraphics[width=\textwidth]{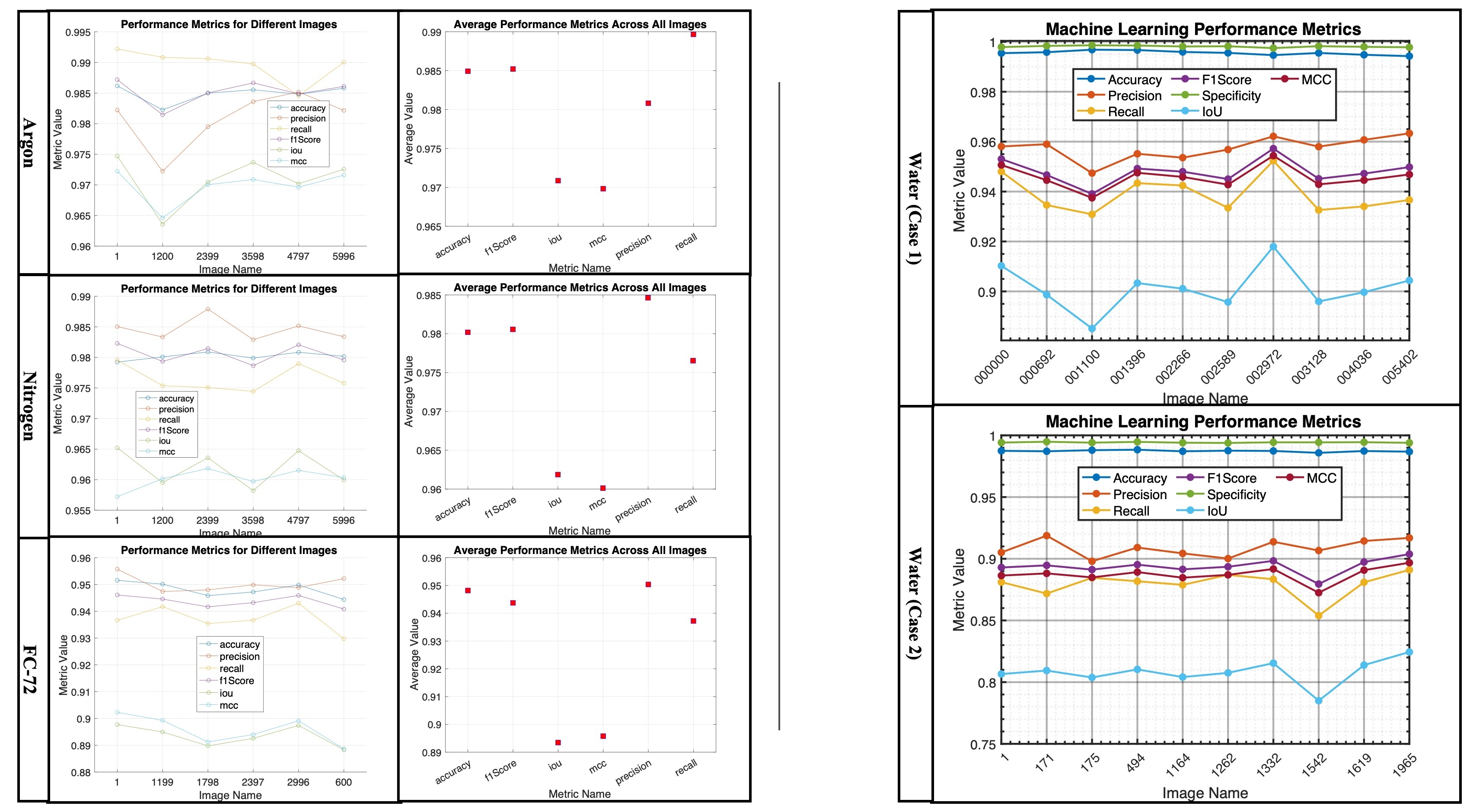}
    \caption{Evaluation of Image Segmentation Performance Metrics Across Different Modalities: Detailed and Aggregate Analysis.}
    \label{fig:segmentation_metrics}
\end{figure}

\begin{figure}[ht!]
    \centering
    \includegraphics[width=\textwidth]{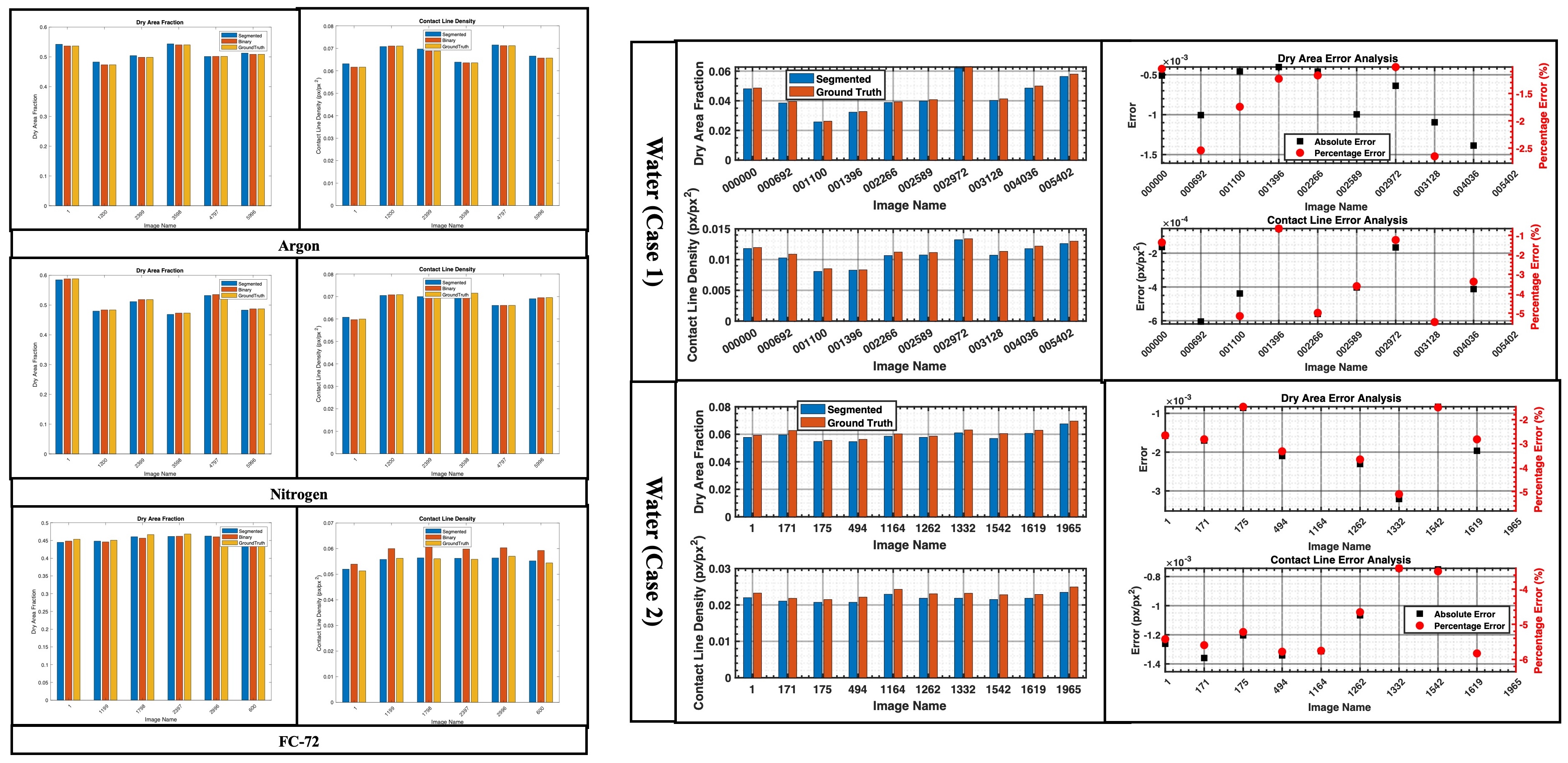}
    \caption{Comparative Boiling Metrics of Dry Area Fraction and Contact Line Density in Argon, Nitrogen, FC-72, and High-Pressure Water Modalities: Segmented vs. Ground Truth Analysis.}
    \label{fig:boiling_metrics}
\end{figure}

\begin{figure}[ht!]
    \centering
    \includegraphics[width=\textwidth]{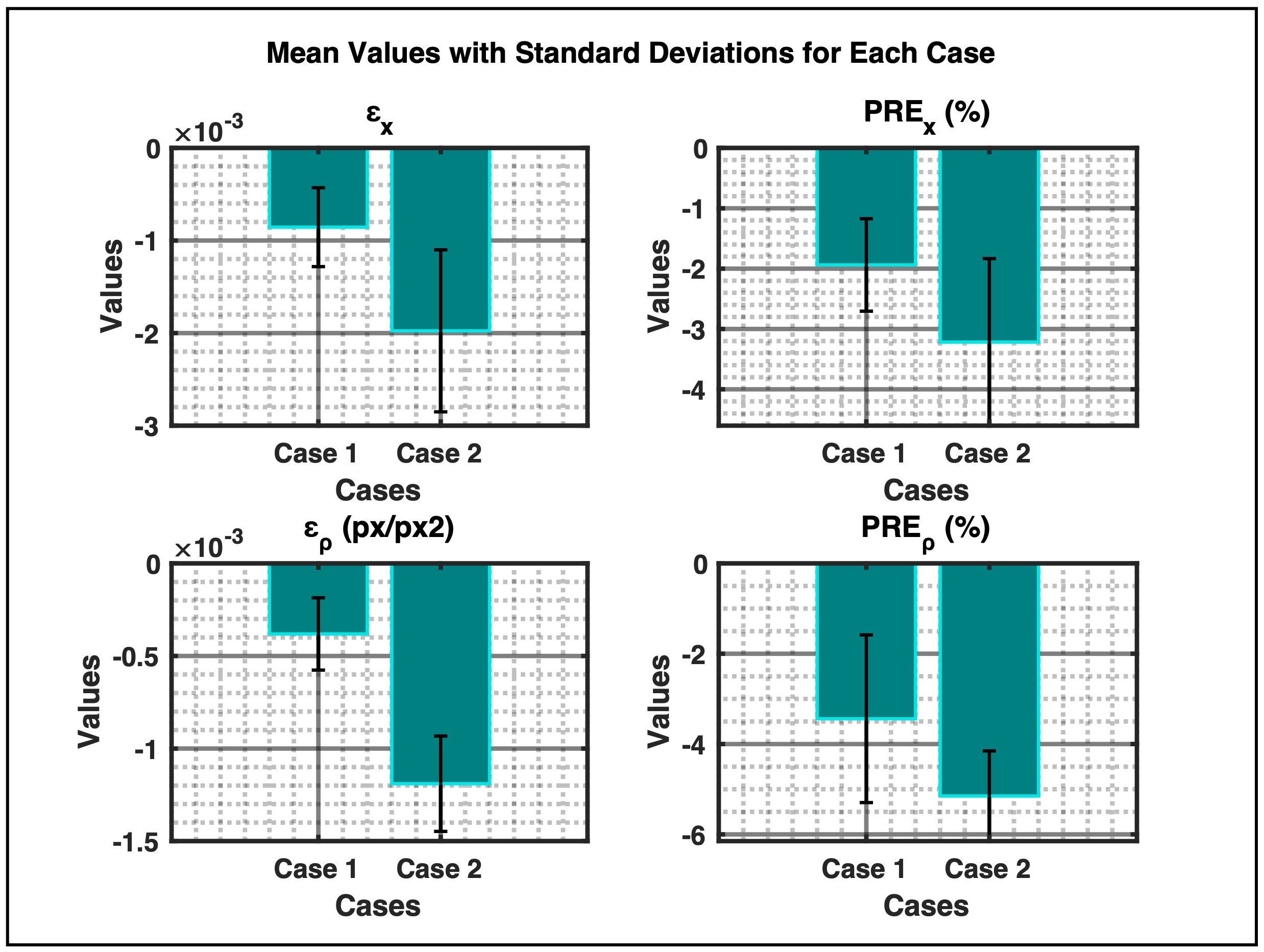}
    \caption{Statistical Analysis of the Absolute Error ($\epsilon$) and Percentage Relative Error (PRE) on the Dry Area Fraction ($x$) and Contact Line Density ($\rho$) for Various Cases.}
    \label{fig:error_distribution}
\end{figure}

Figures \ref{fig:segmentation_metrics} to \ref{fig:error_distribution} illustrate the U-Net CNN model’s segmentation performance and efficacy in capturing boiling dynamics across various experimental conditions and modalities, including argon, nitrogen, FC-72, and high-pressure water (Case 1 and Case 2). The comparative analysis leverages standard segmentation metrics (accuracy, F1 score, IoU, MCC) alongside specific boiling metrics (dry area fraction and contact line density) to understand model performance under diverse conditions.

\paragraph{Segmentation Metrics Across Modalities (Figure \ref{fig:segmentation_metrics})}

Across argon, nitrogen, and FC-72, the U-Net CNN segmentation model demonstrates high accuracy, F1-score, and MCC, especially with argon and nitrogen, where metric values are notably consistent. This suggests that the model is effective in these simpler boiling cases, where bubbles are larger, and fewer small bubbles are present. The FC-72 data, however, exhibits more variability across these metrics, likely due to the greater presence of smaller bubbles and complex boiling patterns that introduce segmentation challenges. The IoU, sensitive to false positives and negatives, shows the greatest variability, particularly in FC-72. This indicates that while the model effectively identifies true positives, the smaller and more complex bubble patterns in FC-72 pose challenges in achieving precise overlap with the ground truth.

For high-pressure water cases, the U-Net model was fine-tuned on a subset of annotated frames to adapt to the challenging conditions of increased pressure and heat flux (10 bar, 3000 kW/m² for Case 1; 40 bar, 3400 kW/m² for Case 2). The model performed robustly, achieving high specificity and accuracy across frames in both cases, indicating strong predictive capabilities for detecting true positives and negatives. However, metrics like precision, recall, and F1-score slightly declined in Case 2, reflecting the increased segmentation complexity introduced by the higher pressure and flux conditions. The IoU, which provides a stringent measure of segmentation overlap, was lowest in Case 2 (around 80\%), underscoring the segmentation challenges posed by high-pressure conditions.

\paragraph{Boiling Metrics: Dry Area Fraction and Contact Line Density (Figure \ref{fig:boiling_metrics})}

For argon and nitrogen, the boiling metrics (dry area fraction and contact line density) derived from U-Net CNN segmentation align closely with the ground-truth annotations, indicating accurate bubble detection across frames. These metrics are consistent across both modalities, highlighting the model’s robust performance in scenarios with relatively straightforward bubble structures. In contrast, FC-72 presents a marked discrepancy in contact line density, especially when using binarization (adaptive thresholding), which tends to overestimate this metric due to the high number of false positives. This overestimation aligns with the challenges discussed in earlier sections, where the binarization method struggled with the smaller, denser bubble patterns of FC-72. When appropriately tuned, the statistical similarity between the U-Net results and ground truth further validates the model’s suitability for complex boiling scenarios.

Regarding boiling metrics for the high-pressure water cases, Figure \ref{fig:boiling_metrics} reveals slight discrepancies between the segmented and ground-truth dry area fractions and contact line densities. These differences, more pronounced in Case 2, likely arise from missed bubble pixels in the segmentation process, where finer details are harder to capture under high-pressure conditions. Absolute error analysis shows values within $10^{-3}$ range, indicating close alignment with ground truth. Yet, percentage errors, especially in contact line density, reach up to 5-6\% in some frames, suggesting room for further refinement in challenging conditions.

\paragraph{Error Analysis and Statistical Insights (Figures \ref{fig:error_distribution})}

Error analysis provides additional insights into model performance consistency. For instance, the absolute and percentage errors for dry area fraction and contact line density reveal that while the U-Net CNN closely follows the ground truth, some underestimation persists across frames, particularly in Case 2. Statistical analysis in Figure \ref{fig:error_distribution} (displaying mean, standard deviation, and error distribution) shows that Case 2 consistently has higher average percentage errors for both metrics than Case 1, reflecting the increased segmentation challenges. The variability observed across these metrics highlights how experimental conditions impact segmentation accuracy and emphasizes the need for fine-tuning to minimize these discrepancies.

\subsection{Uncertainty Quantification}
\label{sec:uncertainty_quantification_results}
This section examines the effects of discretizing binarized bubble images on key boiling parameters, specifically the dry area fraction and contact line density. These parameters are crucial for understanding bubble dynamics in HSV PD segmentation. Since perimeter, area, and radius measurements are fundamental to calculating these parameters, this analysis evaluates the uncertainties introduced by discretization, which depends on factors like bubble size and grid resolution.

To capture this variability, numerical experiments were conducted where grid resolution and bubble size (approximated by bubble radius) were systematically varied to observe their effects on the discretized perimeter and area measurements. Using a theoretical circular bubble as a baseline, its ideal perimeter and area were computed based on the radius. These theoretical values were then compared with discretized measurements, which vary with the grid cell size and the radius. To quantify discrepancies, Percentage Relative Error (PRE) and Mean Error (ME) metrics were introduced, defined as:

\begin{equation}
\mathrm{PRE} = \frac{\psi_{\mathrm{theo}} - \overline{\psi}_{\mathrm{disc}}}{\psi_{\mathrm{theo}}} \times 100
\end{equation}

\begin{equation}
\mathrm{ME} = \psi_{\mathrm{theo}} - \overline{\psi}_{\mathrm{disc}}
\end{equation}

where $\psi_{\mathrm{disc}}$ represents the discretized measurements, and $\psi_{\mathrm{theo}}$, the theoretical value ($2\pi R$ for perimeter and $\pi R^2$ for area), provides a standardized metric for quantifying errors.

\subsubsection{Experimental Data Grounding}

\begin{figure}[ht!]
    \centering
    \includegraphics[width=\textwidth]{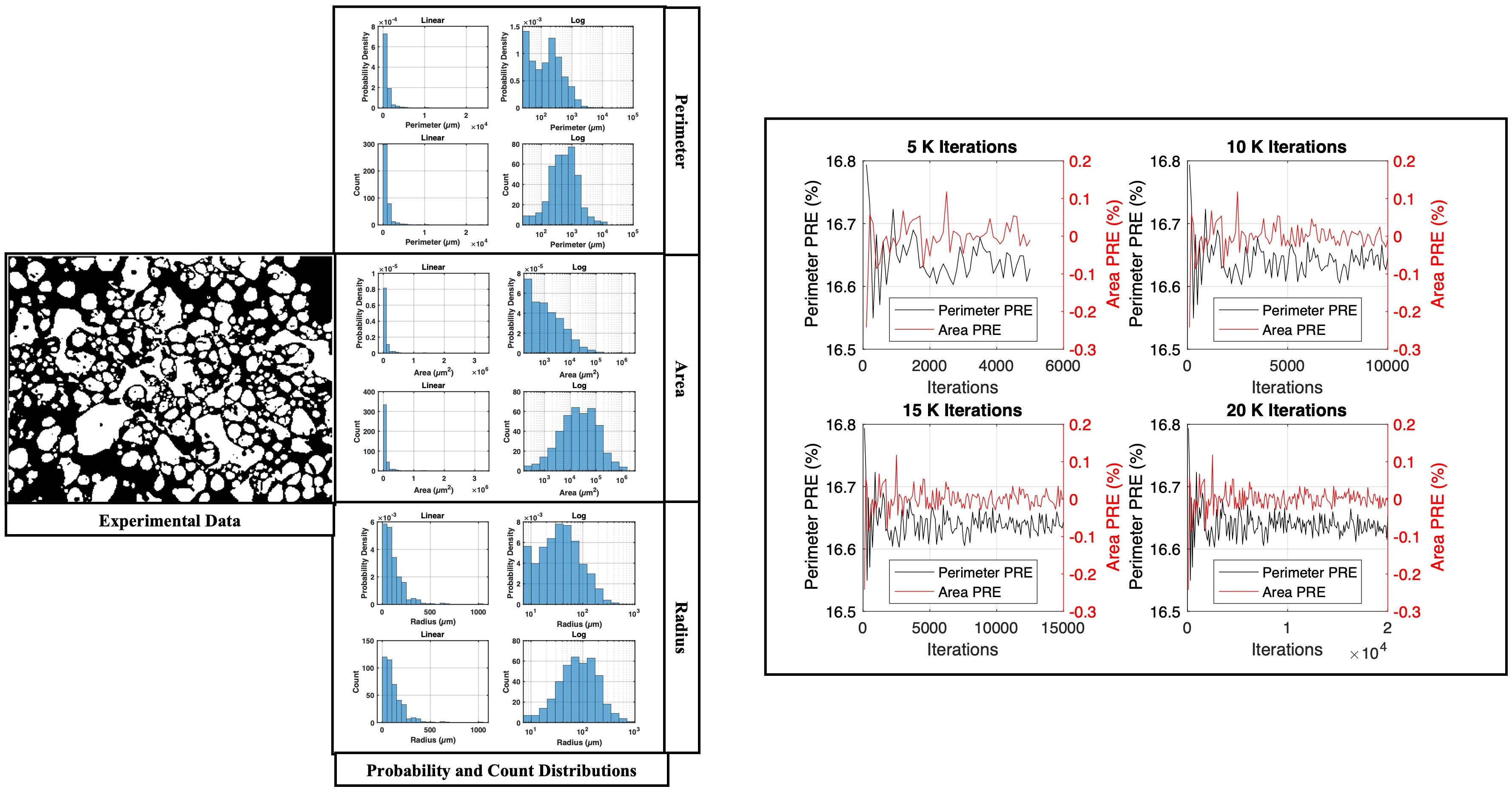}
    \caption{\textbf{Left}: Probability and Count Distributions of the Perimeter, Area, and Radius Distributions of Segmented Experimental Data (Liquid Argon at 1 bar, 120 kW/m\textsuperscript{2}, 9.5 K). \textbf{Right}: Convergence Test Results: Percentage Relative Error (PRE) for Perimeter and Area across Different Iteration Milestones (5K, 10K, 15K, and 20K iterations). Black: Perimeter PRE, Red: Area PRE.}
    \label{fig:argondistributions_convergence}
\end{figure}

To ground this theoretical analysis in experimental data, a representative frame was selected from the liquid argon dataset, captured through HSV imaging at a resolution of 12.6 $\mu$m/px under saturated pool boiling conditions (1 bar pressure, 120 kW/m\textsuperscript{2} heat flux, and 9.5 K wall superheat). The U-Net CNN model was employed to segment this image, distinguishing bubbles (represented by 1) from the liquid background (represented by 0). Figure~\ref{fig:argondistributions_convergence} displays the segmented results and provides probability and count distributions of bubble perimeter, area, and radius—three critical metrics that offer insights into bubble size distribution and spatial dynamics.

The perimeter distribution histogram, presented on both linear and logarithmic scales, reveals a high density of smaller perimeters with a steep drop for larger bubbles. This suggests that smaller bubbles predominate within the argon sample, a common trait in boiling dynamics due to high nucleation rates. The logarithmic representation elongates the tail of the distribution, making the occurrence of larger perimeters more visible. This format provides a clearer view of the distribution's shape across the entire range of bubble sizes, highlighting the occasional presence of larger bubbles, even though they are less frequent.

A similar trend is observed for bubble areas. The linear scale histogram shows a high concentration of smaller bubbles, while larger areas are sparse. However, larger bubbles become more noticeable when viewed on a logarithmic scale, indicating that although smaller bubbles dominate, significant occurrences of larger bubbles persist. This bimodal view implies a dynamic bubble environment where rapid nucleation leads to small bubbles, and sporadic coalescence or slower growth results in larger ones.

The radius distribution further supports these observations. On a linear scale, the histogram confirms that smaller bubbles (with lower radii) are more frequent, indicating a preponderance of minor nucleation events. However, the logarithmic scale reveals a more diverse array of bubble sizes, suggesting that bubble dynamics encompass a range of behaviors, including both rapid nucleation and coalescence processes. This diversity likely reflects factors influencing bubble size, such as surface roughness, wall temperature, and fluid properties affecting nucleation, coalescence, and fragmentation.

\subsubsection{Convergence Test of Discretization Errors in Bubble Measurements}
\label{sec:convergence_test}

A convergence test was conducted across different grid resolutions and bubble sizes to analyze the impact of discretization errors on bubble measurements in experimental data. The analysis was performed over a 0.001 m domain ($L$), varying grid cell sizes ($N$) from 5 to 50 microns, and bubble radii ($R$) ranging from 0 to 200 microns. This range aligns with observed bubble distributions, as highlighted in Figure~\ref{fig:argondistributions_convergence}, where bubbles frequently fall within this radius spectrum. In the simulation, a bubble was randomly placed within the domain, and the discretized area and perimeter values were calculated for each combination of $N$ and $R$.

The right-handed plot in Figure~\ref{fig:argondistributions_convergence} presents the convergence behavior of the Percentage Relative Error (PRE) for both perimeter and area across different iteration milestones: 5,000 (5K), 10,000 (10K), 15,000 (15K), and 20,000 (20K) iterations. Each plot shows the PRE for perimeter (black) and area (red), illustrating how stability and convergence evolve with iteration count:

\paragraph{5K Iterations:} At this early stage, the perimeter PRE demonstrates relative stability, fluctuating within a narrow range, which suggests an initial quasi-convergence. However, the area PRE shows higher variability, indicating that area calculations require more iterations to stabilize than perimeter measurements.

\paragraph{10K Iterations:} Doubling the iteration count significantly reduces perimeter and area PRE fluctuations, suggesting enhanced stability in the numerical estimates. The reduction in variability at this stage indicates that the simulation is progressively converging towards more accurate values.

\paragraph{15K Iterations:} Perimeter and area PRE fluctuations decrease as iterations increase. The oscillation amplitude of area PRE narrows further, suggesting improved convergence towards the true area values.

\paragraph{20K Iterations:} At this final milestone, perimeter and area PRE show minimal oscillations. The area PRE is particularly stable, oscillating close to zero, indicating strong convergence. The perimeter PRE also shows minimal variability, supporting the conclusion that the simulation has reached a stable numerical solution.

These results underscore that the simulation achieves greater stability as iterations increase, with a progressively diminishing PRE for both perimeter and area. This aligns with the expected behavior of a converging numerical solution, where higher iteration counts yield reduced errors. The perimeter measurement converges faster than the area, possibly due to the inherent sensitivity differences in discretization for linear (perimeter) versus areal (area) calculations.

Based on this analysis, 20K iterations were selected as the optimal iteration count for the simulation, ensuring accuracy and computational efficiency. This configuration results in a total runtime of approximately 987.74 seconds (16.46 minutes), balancing the need for convergence with practical computational limits.

\subsubsection{Impact of Bubble Radius and Grid Resolution on Measurement Accuracy}
\label{sec:grid_resolution}

In analyzing the discretization effects on bubble perimeter and area measurements, Figure~\ref{fig:radii_grid_impacts} provides a comprehensive visualization of how bubble radius and grid cell size impact the Mean Error (ME) and Percentage Relative Error (PRE) under various grid resolutions. This analysis is instrumental for quantifying uncertainties in measurements such as dry area fraction and contact line density during bubble segmentation in boiling experiments.

The ME plots for perimeter and area measurements demonstrate complex interactions between grid cell size and bubble radius, revealing trends that underscore the challenges of digital discretization. In the ME plot for the perimeter, we observe a valley-shaped distribution, where perimeter errors transition from overestimation (negative ME) at finer grid resolutions to underestimation (positive ME) at coarser resolutions. This distribution highlights an optimal range where perimeter measurements are most accurate, likely due to finer grids capturing edges more accurately. As grid sizes increase, errors are introduced due to either omission or misalignment of boundary pixels, resulting in under- or overestimation. Interestingly, error magnitude increases at extreme radii, indicating that very small and large bubbles are more challenging to measure accurately within this grid framework.

The ME plot for area displays a similarly nuanced relationship but with greater variability. Unlike perimeter, area measurements are less sensitive to boundary pixel adjustments, as the area is a bulk measure. The squared nature of area calculation amplifies miscounted pixels, leading to abrupt transitions from negative to positive errors. Area measurements tend to stabilize with larger bubbles, but the overall ME is lower compared to perimeter errors. This suggests that while perimeter measurements are highly influenced by grid size, area calculations show resilience, especially in larger bubble regions.

Examining the PRE plots for perimeter and area yields additional insights into measurement fidelity. The PRE plot for the perimeter reveals an overestimation (negative PRE) at finer grid sizes for smaller bubbles, likely due to diagonal pixel inclusion, which artificially extends perimeter length. As grid sizes increase, perimeter measurements shift to underestimation (positive PRE), particularly with larger bubbles, where the stair-stepping approximation of the curved boundary becomes more pronounced, reducing measurement accuracy. This shift from over- to underestimation as the bubble radius grows suggests a non-linear relationship between bubble size and pixel resolution in perimeter measurements.

However, the PRE plot for the area presents a more consistent trend. Overestimation (negative PRE) is predominant across most grid sizes and bubble radii, indicating that area measurements are generally inflated due to the inclusion of partial pixels at the edges. Unlike perimeter, area does not exhibit a clear transition from negative to positive PRE values, suggesting that area calculations are less susceptible to pixel resolution changes relative to bubble size. At larger bubble sizes with coarser grids, a minor increase in positive PRE hints at a threshold where pixelation affects area measurements more significantly, albeit less so than the perimeter.

\subsubsection{Effects of Erosion and Dilation on Measurement Accuracy}
\label{sec:erosion_dilation}

We applied error metrics—Mean Error (ME) and Percentage Relative Error (PRE)—to the dry area fraction and contact line density calculations to quantify the uncertainties in experimental HSV data measurements. The dry area fraction and contact line density values derived from the experimental data (Figure~\ref{fig:argondistributions_convergence}) were 0.48283 and 0.005363~$\mu$m$^{-1}$, respectively, with a resolution of 12.6~$\mu$m. We extracted corresponding error values from Figure~\ref{fig:radii_grid_impacts} using the observed bubble radius distribution, weighted by frequency. These values, reflecting area and perimeter measurements, were compiled into an uncertainty table (Table~\ref{tab:uncertainty_values}). Weighted averages were computed using Equation~\ref{eq:weighted_avg}, with frequency-based weights emphasizing the most common bubble sizes.

\begin{equation}
    W = \frac{\sum_{i=1}^k \left(v_i \cdot w_i\right)}{\sum_{i=1}^k w_i}
    \label{eq:weighted_avg}
\end{equation}

where \( v_i \) represents each value from the set (e.g., PRE or ME for contact line density or dry area fraction, as shown in Table~\ref{tab:uncertainty_values}). In this context, the corresponding \( w_i \) is the weight for each \( v_i \), defined as the frequency of occurrence for each radius bin. The parameter \( k \) denotes the number of values the calculation considers. This weighted average allows for a more accurate representation of the overall error by accounting for the frequency of each bubble size in the dataset.

\begin{table}[ht!]
    \centering
    \caption{Uncertainty Table}
    \label{tab:uncertainty_values}
    \begin{tabular}{cccccc}
        \hline
        S/N & Frequency & Area PRE (\%) & Area ME ($\times10^{-12}$) & Perimeter PRE (\%) & Perimeter ME ($\times10^{-5}$) \\
        \hline
        1 & 184 & -0.5 & -1.6 & -16.3 & -1.0 \\
        2 & 110 & 0.2 & 5.6 & -1.6 & -0.3 \\
        3 & 59 & 0.03 & 2.4 & 2.8 & 0.9 \\
        4 & 31 & 0.01 & 1.8 & 4.8 & 2.2 \\
        5 & 11 & -0.01 & -2.2 & 5.9 & 3.5 \\
        6 & 7 & 0.003 & 1.6 & 7.1 & 6.1 \\
        7 & 3 & 0.003 & 1.6 & 8.0 & 6.1 \\
        8 & 2 & 0.006 & 8.1 & 8.0 & 10.1 \\
        \hline
    \end{tabular}
\end{table}

\begin{figure}[ht!]
    \centering
    \includegraphics[width=0.8\textwidth]{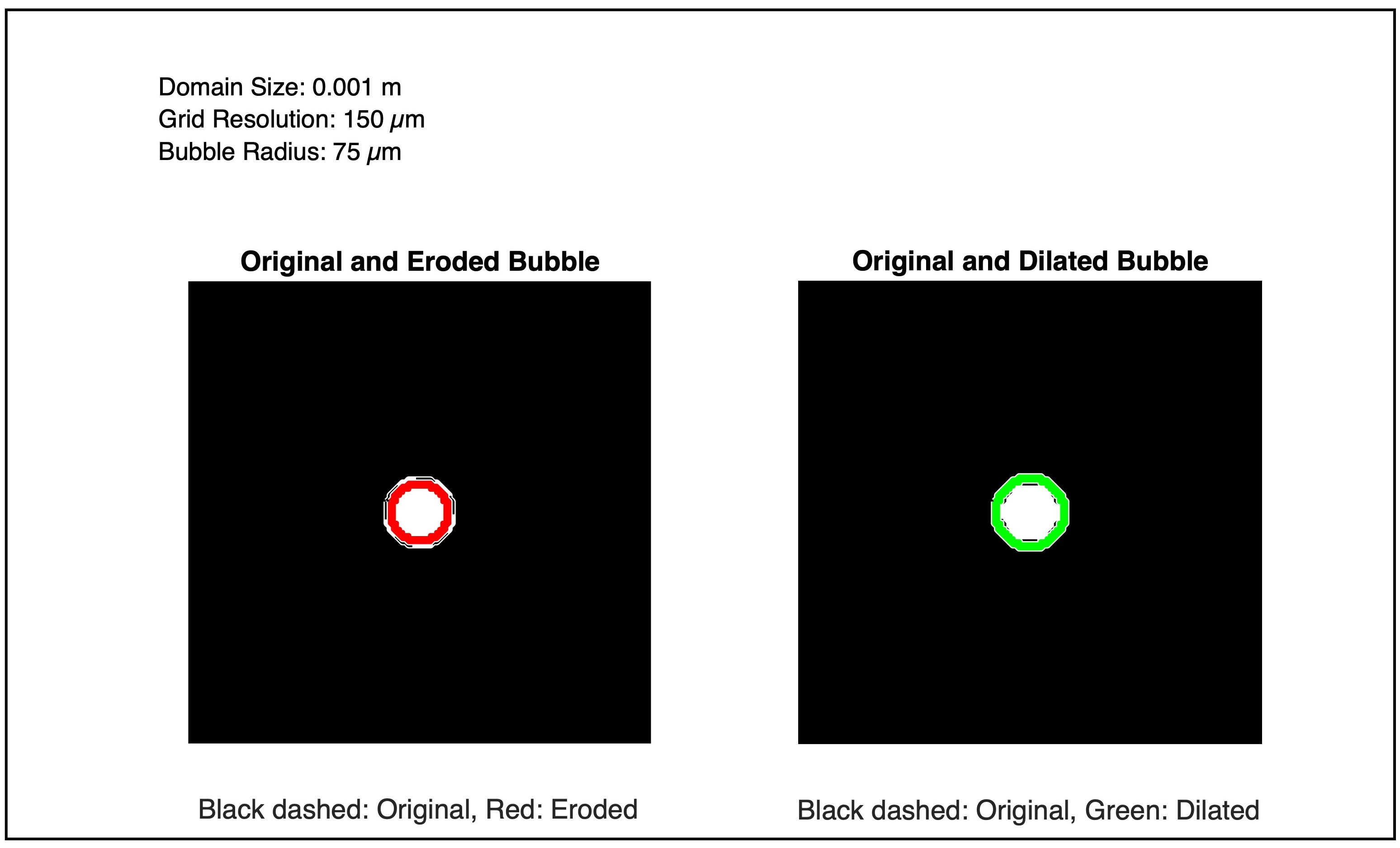}
    \caption{Difference Between Erosion and Dilation}
    \label{fig:erosion_dilation}
\end{figure}

Figure~\ref{fig:erosion_dilation} illustrates the visual impact of erosion and dilation on bubble boundaries, helping clarify their effects on measurement accuracy. Erosion reduces the boundary by one pixel, while dilation expands it, potentially affecting perimeter estimates. 

Table~\ref{tab:boundary_modifications} summarizes how these boundary modifications influence uncertainties. The analysis revealed that while the area PRE remained constant, perimeter PRE decreased from 13.8\% to 6.6\% under dilation. This suggests that dilation smooths boundaries, reducing edge artifacts that cause perimeter overestimation during erosion. Consequently, dilation results in a closer approximation to the actual bubble perimeter, underscoring the importance of selecting appropriate boundary conditions in digital image processing.

\begin{table}[ht!]
    \centering
    \caption{Effects of Dilation on the Uncertainty Quantification}
    \label{tab:boundary_modifications}
    \begin{tabular}{cccc}
        \hline
        Case Type & Weighted Avg PRE Area (\%) & Weighted Avg PRE Perimeter (\%) & Weighted Avg ME Perimeter ($\mu$m) \\
        \hline
        Erosion & -0.4 & 13.8 & 1.2e-4 \\
        Dilation & -0.4 & 6.6 & 7.1e-5 \\
        \hline
    \end{tabular}
\end{table}

\subsubsection{Extended Analysis of Uncertainty Across Modalities}
\label{sec:modality_uq}

We applied the same uncertainty quantification process to datasets for FC-72, liquid argon (LAr), liquid nitrogen (LN2), and high-pressure water to further examine the impact across various modalities. The results are presented in Table~\ref{tab:uncertainty_modalities}. PRE values for the area were consistent across modalities, ranging from -0.03\% for water to -0.06\% for LN2, indicating minor deviations from the true location. However, perimeter uncertainties varied more substantially, with LN2 showing the highest PRE at -8.4\% and water the lowest at -1\%. This disparity highlights perimeter measurements’ greater sensitivity to pixel resolution, likely due to the reliance on edge details that are more susceptible to pixelization effects.

\paragraph{Implications of Measurement Uncertainty}
The higher PRE for perimeter measurements across all modalities implies that perimeter estimations are more prone to uncertainty than area measurements, particularly in LN2, where boundary irregularities are amplified. Such uncertainties could influence heat flux calculations that depend on precise measurements of bubble boundaries and areas. When comparing uncertainties from pixelization errors to segmentation errors, we find that, for instance, the pixelization uncertainty in FC-72 (6.8\%) aligns closely with the mean segmentation error of 6.5\%. This suggests that pixelization and segmentation contribute comparably to the overall uncertainty, identifying them as critical factors in the accuracy of heat flux reconstructions.

\subsection{VideoSAM Results}
\subsubsection{Zero-Shot Generalization Across Modalities}

\begin{figure}[ht!]
    \centering
    \includegraphics[width=\textwidth]{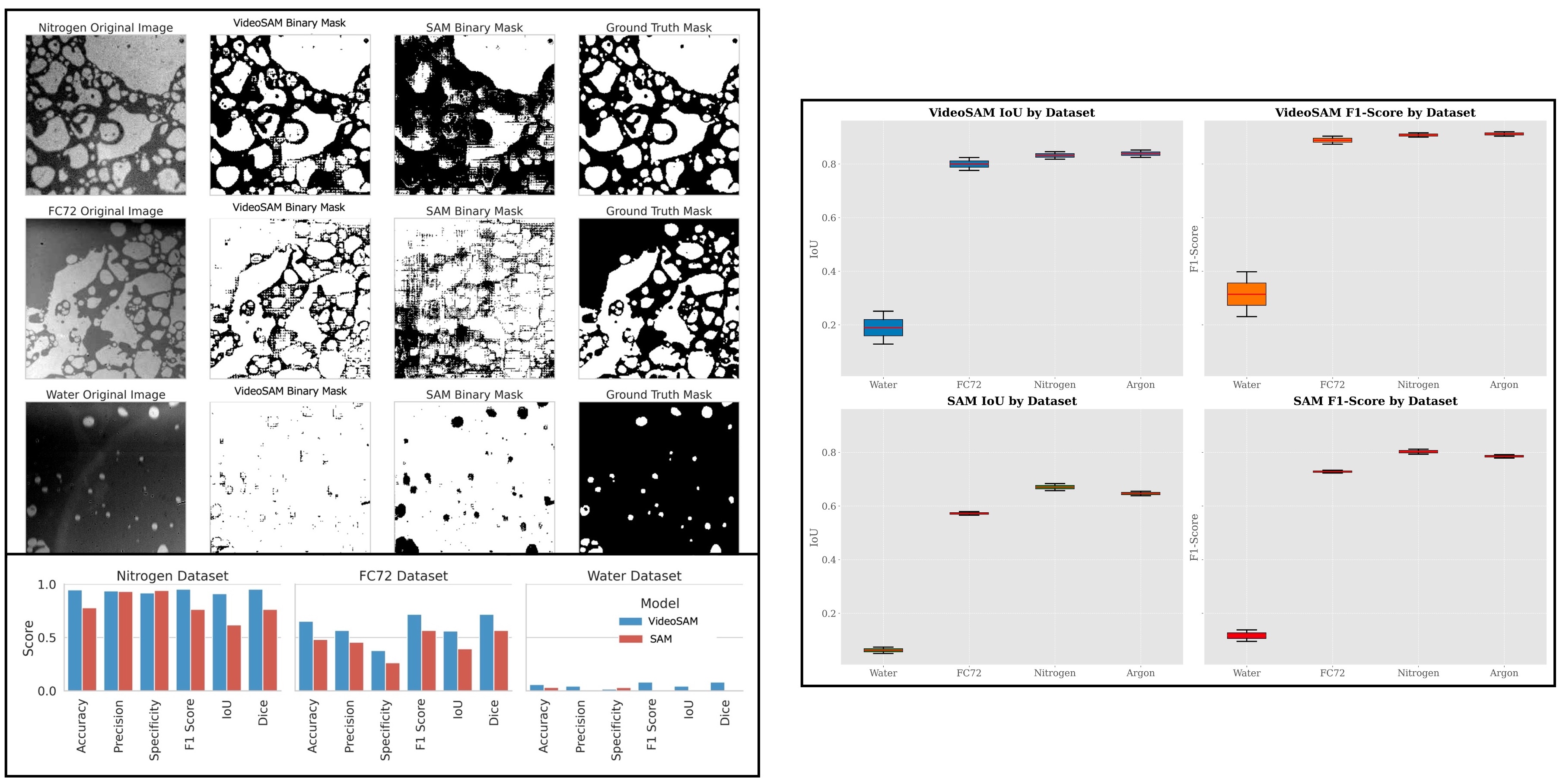}
    \caption{Qualitative and Quantitative Comparison of Binary Masks for Nitrogen, FC-72, and Water Datasets between VideoSAM, SAM, and Ground Truth Masks.}
    \label{fig:videosam_results}
\end{figure}

This experiment evaluates the zero-shot performance of VideoSAM on unseen data modalities, including the Nitrogen, FC-72, and Water datasets, with both qualitative and quantitative analyses depicted on the left-handed plots of Figure~\ref{fig:videosam_results}. The segmentation results portrayed in the top-left plots of Figure~\ref{fig:videosam_results} highlight VideoSAM’s impressive ability to generalize across different datasets, especially in complex environments like Nitrogen and FC-72. VideoSAM’s binary masks closely align with the ground truth, maintaining precise contours and boundaries of bubbles, which the SAM model struggles with. The SAM model’s segmentation displays inconsistencies, particularly noticeable in high-density regions where VideoSAM excels at capturing intricate bubble structures. This robustness in VideoSAM’s segmentation highlights its suitability for environments with diverse and irregular bubble formations, as seen in the FC-72 dataset.

However, VideoSAM’s performance declines with the Water dataset. Here, the simpler structure, fewer objects of interest, and prevalent background noise undermine its segmentation accuracy, resulting in poorly defined binary masks that fail to distinguish bubbles clearly. This degradation in performance suggests a challenge for VideoSAM when applied to datasets with low contrast and minimal object variability.

The quantitative metrics in the bottom-left bar chart in Figure~\ref{fig:videosam_results} reinforce these observations. VideoSAM consistently outperforms SAM in the Nitrogen and FC-72 datasets, achieving higher scores in accuracy, precision, IoU, and Dice coefficients. Particularly in the Nitrogen dataset, VideoSAM demonstrates substantial improvements in boundary detection and reduced false positives, as reflected in its enhanced specificity. In contrast, the Water dataset’s metrics reveal a significant drop, with notably low IoU and Dice scores, indicating that VideoSAM may be over-tuned to handle dense and complex bubble structures. Its architecture appears less effective in simpler scenarios with fewer segmentation cues.

\subsubsection{Multi-Modal Segmentation Performance Analysis}
\label{sec:videosam_multimodal}

This section presents an integrated evaluation of the segmentation performance of VideoSAM, SAM, and U-Net across multiple modalities—Water, FC-72, Nitrogen, and Argon—using IoU and F1 Score as the primary metrics, summarized in Table~\ref{tab:performance_metrics} and the right-handed box-plot in Figure~\ref{fig:videosam_results}. The analysis encompasses single-frame (Table~\ref{tab:performance_metrics}) and composite-frame (right-handed plot in Figure~\ref{fig:videosam_results}) evaluations, providing insights into the models' robustness in complex and simpler environments.

In datasets with complex environments, such as FC-72, Nitrogen, and Argon, VideoSAM consistently outperforms SAM and U-Net. These datasets are characterized by intricate bubble dynamics, overlapping structures, and high-density regions, which present challenges for precise segmentation. For instance, in the Nitrogen dataset, VideoSAM achieves an IoU of 0.8317 and an F1 Score of 0.9080, considerably higher than SAM’s IoU of 0.6702 and F1 Score of 0.8025. Similarly, in the FC-72 dataset, VideoSAM achieves an IoU of 0.7997 and an F1 Score of 0.8885, highlighting its superior ability to capture boundary details and segment overlapping bubble structures. While U-Net performs competitively in these environments, it does not match VideoSAM’s robustness, as reflected in lower IoU and F1 scores. This suggests that VideoSAM’s architecture is better suited for handling complex segmentation tasks, likely due to its enhanced capability for boundary preservation and detailed feature capture.

Conversely, U-Net demonstrates the best segmentation performance in the simpler Water dataset with fewer objects and a more uniform background, achieving an IoU of 0.5619 and an F1 Score of 0.7191. VideoSAM performs less effectively in this dataset, with an IoU of 0.1894 and an F1 Score of 0.3143. In contrast, SAM performs even lower, indicating significant difficulty distinguishing bubbles from the background in the Water dataset. The reduced performance of VideoSAM in this environment underscores its limitations when faced with datasets lacking structural complexity. The model, optimized for dynamic structures, struggles with the simplicity and low contrast of the Water dataset. On the other hand, U-Net appears to adapt well to these simpler environments, suggesting that its cellular-level segmentation design aligns better with the characteristics of datasets with more uniform structures and minimal background noise.

\begin{thebibliography}{54}
\providecommand{\natexlab}[1]{#1}
\providecommand{\url}[1]{\texttt{#1}}
\expandafter\ifx\csname urlstyle\endcsname\relax
  \providecommand{\doi}[1]{doi: #1}\else
  \providecommand{\doi}{doi: \begingroup \urlstyle{rm}\Url}\fi

\bibitem[Brzozowski et~al.(2022)Brzozowski, Matuszyk, Pieczara, Firlej, Nowakowska, and Baranska]{BRZOZOWSKI2022108003}
Brzozowski, K., Matuszyk, E., Pieczara, A., Firlej, J., Nowakowska, A., and Baranska, M.
\newblock Stimulated raman scattering microscopy in chemistry and life science – development, innovation, perspectives.
\newblock \emph{Biotechnology Advances}, 60:\penalty0 108003, 2022.
\newblock ISSN 0734-9750.
\newblock \doi{https://doi.org/10.1016/j.biotechadv.2022.108003}.
\newblock URL \url{https://www.sciencedirect.com/science/article/pii/S0734975022000994}.

\bibitem[Bucci et~al.(2016)Bucci, Richenderfer, Su, McKrell, and Buongiorno]{BUCCI2016115}
Bucci, M., Richenderfer, A., Su, G.-Y., McKrell, T., and Buongiorno, J.
\newblock A mechanistic ir calibration technique for boiling heat transfer investigations.
\newblock \emph{International Journal of Multiphase Flow}, 83:\penalty0 115--127, 2016.
\newblock ISSN 0301-9322.
\newblock \doi{https://doi.org/10.1016/j.ijmultiphaseflow.2016.03.007}.
\newblock URL \url{https://www.sciencedirect.com/science/article/pii/S030193221530183X}.

\bibitem[Chavagnat et~al.(2021)Chavagnat, Nop, Dorville, Phillips, and Bucci]{CHAVAGNAT2021121294}
Chavagnat, F., Nop, R., Dorville, N., Phillips, B., and Bucci, M.
\newblock Single-phase heat transfer regimes in forced flow conditions under exponential heat inputs.
\newblock \emph{International Journal of Heat and Mass Transfer}, 174:\penalty0 121294, 2021.
\newblock ISSN 0017-9310.
\newblock \doi{https://doi.org/10.1016/j.ijheatmasstransfer.2021.121294}.
\newblock URL \url{https://www.sciencedirect.com/science/article/pii/S0017931021003975}.

\bibitem[Cheng et~al.(2022)Cheng, Misra, Schwing, Kirillov, and Girdhar]{9878483}
Cheng, B., Misra, I., Schwing, A.~G., Kirillov, A., and Girdhar, R.
\newblock Masked-attention mask transformer for universal image segmentation.
\newblock In \emph{Proceedings of the IEEE/CVF Conference on Computer Vision and Pattern Recognition (CVPR)}, pp.\  1290--1299, June 2022.
\newblock URL \url{https://ieeexplore.ieee.org/document/9878483}.

\bibitem[Cheng et~al.(2023)Cheng, Li, Xu, Li, Yang, Wang, and Yang]{cheng2023segmenttrack}
Cheng, Y., Li, L., Xu, Y., Li, X., Yang, Z., Wang, W., and Yang, Y.
\newblock Segment and track anything, 2023.
\newblock URL \url{https://arxiv.org/abs/2305.06558}.

\bibitem[{Chi Wang} et~al.(2024){Chi Wang}, {Guanyu Su}, {Olorunsola Akinsulire}, {Limiao Zhang}, {Md Mahamudur Rahman}, and {Matteo Bucci}]{doi:10.1080/01457632.2023.2191441}
{Chi Wang}, {Guanyu Su}, {Olorunsola Akinsulire}, {Limiao Zhang}, {Md Mahamudur Rahman}, and {Matteo Bucci}.
\newblock Investigation of critical heat flux enhancement on nanoengineered surfaces in pressurized subcooled flow boiling using infrared thermometry.
\newblock \emph{Heat Transfer Engineering}, 45\penalty0 (4-5):\penalty0 417--432, 2024.
\newblock \doi{10.1080/01457632.2023.2191441}.
\newblock URL \url{https://doi.org/10.1080/01457632.2023.2191441}.

\bibitem[Choi et~al.(2022)Choi, Kim, and Park]{Choi2022}
Choi, D., Kim, H., and Park, H.
\newblock Bubble velocimetry using the conventional and cnn-based optical flow algorithms.
\newblock \emph{Scientific Reports}, 12\penalty0 (1):\penalty0 11879, 2022.
\newblock ISSN 2045-2322.
\newblock \doi{10.1038/s41598-022-16145-y}.
\newblock URL \url{https://doi.org/10.1038/s41598-022-16145-y}.

\bibitem[Dunlap et~al.(2022)Dunlap, Pandey, and Hu]{10.1115/HT2022-85582}
Dunlap, C., Pandey, H., and Hu, H.
\newblock Supervised and unsupervised learning models for detection of critical heat flux during pool boiling.
\newblock In \emph{ASME 2022 Heat Transfer Summer Conference collocated with the ASME 2022 16th International Conference on Energy Sustainability}, volume ASME 2022 Heat Transfer Summer Conference of \emph{Heat Transfer Summer Conference}, pp.\  V001T08A004, 07 2022.
\newblock \doi{10.1115/HT2022-85582}.
\newblock URL \url{https://doi.org/10.1115/HT2022-85582}.

\bibitem[Eppel et~al.(2020)Eppel, Xu, Bismuth, and Aspuru-Guzik]{Eppel2020}
Eppel, S., Xu, H., Bismuth, M., and Aspuru-Guzik, A.
\newblock Computer vision for recognition of materials and vessels in chemistry lab settings and the vector-labpics data set.
\newblock \emph{ACS Central Science}, 6\penalty0 (10):\penalty0 1743--1752, 2020.
\newblock ISSN 2374-7943.
\newblock \doi{10.1021/acscentsci.0c00460}.
\newblock URL \url{https://doi.org/10.1021/acscentsci.0c00460}.

\bibitem[Hanafizadeh et~al.(2011)Hanafizadeh, Ghanbarzadeh, and Saidi]{HANAFIZADEH2011327}
Hanafizadeh, P., Ghanbarzadeh, S., and Saidi, M.~H.
\newblock Visual technique for detection of gas–liquid two-phase flow regime in the airlift pump.
\newblock \emph{Journal of Petroleum Science and Engineering}, 75\penalty0 (3):\penalty0 327--335, 2011.
\newblock ISSN 0920-4105.
\newblock \doi{https://doi.org/10.1016/j.petrol.2010.11.028}.
\newblock URL \url{https://www.sciencedirect.com/science/article/pii/S092041051000272X}.

\bibitem[Hassan et~al.(2023)Hassan, Feeney, Dhruv, Kim, Suh, Ryu, Won, and Chandramowlishwaran]{NEURIPS2023_01726ae0}
Hassan, S. M.~S., Feeney, A., Dhruv, A., Kim, J., Suh, Y., Ryu, J., Won, Y., and Chandramowlishwaran, A.
\newblock Bubbleml: A multiphase multiphysics dataset and benchmarks for machine learning.
\newblock In Oh, A., Naumann, T., Globerson, A., Saenko, K., Hardt, M., and Levine, S. (eds.), \emph{Advances in Neural Information Processing Systems}, volume~36, pp.\  418--449. Curran Associates, Inc., 2023.
\newblock URL \url{https://proceedings.neurips.cc/paper_files/paper/2023/file/01726ae05d72ddba3ac784a5944fa1ef-Paper-Datasets_and_Benchmarks.pdf}.

\bibitem[He et~al.(2017)He, Gkioxari, Dollár, and Girshick]{8237584}
He, K., Gkioxari, G., Dollár, P., and Girshick, R.
\newblock Mask r-cnn.
\newblock In \emph{2017 IEEE International Conference on Computer Vision (ICCV)}, pp.\  2980--2988, 2017.
\newblock \doi{10.1109/ICCV.2017.322}.
\newblock URL \url{https://ieeexplore.ieee.org/document/8237584?ref=labellerr.com}.

\bibitem[Hu(2023)]{Hu2023}
Hu, H.
\newblock Boiling image dataset: Chf vs. pre-chf - water, polished copper, ambient pressure, 2023.
\newblock URL \url{https://doi.org/10.17632/g53k5w6ptb.1}.

\bibitem[Huang et~al.(2024)Huang, Yang, Liu, Zhou, Chang, Zhou, Chen, Yu, Chen, Chen, Liu, Chi, Hu, Yue, Li, Grau, Fan, Dong, and Ni]{HUANG2024103061}
Huang, Y., Yang, X., Liu, L., Zhou, H., Chang, A., Zhou, X., Chen, R., Yu, J., Chen, J., Chen, C., Liu, S., Chi, H., Hu, X., Yue, K., Li, L., Grau, V., Fan, D.-P., Dong, F., and Ni, D.
\newblock Segment anything model for medical images?
\newblock \emph{Medical Image Analysis}, 92:\penalty0 103061, 2024.
\newblock ISSN 1361-8415.
\newblock \doi{https://doi.org/10.1016/j.media.2023.103061}.
\newblock URL \url{https://www.sciencedirect.com/science/article/pii/S1361841523003213}.

\bibitem[Ji et~al.(2024)Ji, Li, Bi, Liu, Li, and Cheng]{Ji2024}
Ji, W., Li, J., Bi, Q., Liu, T., Li, W., and Cheng, L.
\newblock Segment anything is not always perfect: An investigation of sam on different real-world applications.
\newblock \emph{Machine Intelligence Research}, 21\penalty0 (4):\penalty0 617--630, 2024.
\newblock ISSN 2731-5398.
\newblock \doi{10.1007/s11633-023-1385-0}.
\newblock URL \url{https://doi.org/10.1007/s11633-023-1385-0}.

\bibitem[Jin \& Shirvan(2021)Jin and Shirvan]{JIN2021121517}
Jin, Y. and Shirvan, K.
\newblock Study of the film boiling heat transfer and two-phase flow interface behavior using image processing.
\newblock \emph{International Journal of Heat and Mass Transfer}, 177:\penalty0 121517, 2021.
\newblock ISSN 0017-9310.
\newblock \doi{https://doi.org/10.1016/j.ijheatmasstransfer.2021.121517}.
\newblock URL \url{https://www.sciencedirect.com/science/article/pii/S0017931021006207}.

\bibitem[Jüngst et~al.(2024)Jüngst, Ersoy, Smallwood, and Kaiser]{JUNGST2024106314}
Jüngst, N., Ersoy, V., Smallwood, G.~J., and Kaiser, S.~A.
\newblock Neural networks for classification and segmentation of thermally-induced droplet breakup in spray-flame synthesis.
\newblock \emph{Journal of Aerosol Science}, 176:\penalty0 106314, 2024.
\newblock ISSN 0021-8502.
\newblock \doi{https://doi.org/10.1016/j.jaerosci.2023.106314}.
\newblock URL \url{https://www.sciencedirect.com/science/article/pii/S0021850223001799}.

\bibitem[Kim \& Park(2021)Kim and Park]{Kim2021}
Kim, Y. and Park, H.
\newblock Deep learning-based automated and universal bubble detection and mask extraction in complex two-phase flows.
\newblock \emph{Scientific Reports}, 11\penalty0 (1):\penalty0 8940, 2021.
\newblock ISSN 2045-2322.
\newblock \doi{10.1038/s41598-021-88334-0}.
\newblock URL \url{https://doi.org/10.1038/s41598-021-88334-0}.

\bibitem[Kirillov et~al.(2023)Kirillov, Mintun, Ravi, Mao, Rolland, Gustafson, Xiao, Whitehead, Berg, Lo, Dollar, and Girshick]{Kirillov_2023_ICCV}
Kirillov, A., Mintun, E., Ravi, N., Mao, H., Rolland, C., Gustafson, L., Xiao, T., Whitehead, S., Berg, A.~C., Lo, W.-Y., Dollar, P., and Girshick, R.
\newblock Segment anything.
\newblock In \emph{Proceedings of the IEEE/CVF International Conference on Computer Vision (ICCV)}, pp.\  4015--4026, October 2023.
\newblock URL \url{https://openaccess.thecvf.com/content/ICCV2023/html/Kirillov_Segment_Anything_ICCV_2023_paper.html}.

\bibitem[Kossolapov et~al.(2021)Kossolapov, Phillips, and Bucci]{KOSSOLAPOV2021103522}
Kossolapov, A., Phillips, B., and Bucci, M.
\newblock Can led lights replace lasers for detailed investigations of boiling phenomena?
\newblock \emph{International Journal of Multiphase Flow}, 135:\penalty0 103522, 2021.
\newblock ISSN 0301-9322.
\newblock \doi{https://doi.org/10.1016/j.ijmultiphaseflow.2020.103522}.
\newblock URL \url{https://www.sciencedirect.com/science/article/pii/S0301932220306339}.

\bibitem[Kossolapov et~al.(2024)Kossolapov, Hughes, Phillips, and Bucci]{Kossolapov_2024}
Kossolapov, A., Hughes, M.~T., Phillips, B., and Bucci, M.
\newblock Bubble departure and sliding in high-pressure flow boiling of water.
\newblock \emph{Journal of Fluid Mechanics}, 987, May 2024.
\newblock ISSN 1469-7645.
\newblock \doi{10.1017/jfm.2024.405}.
\newblock URL \url{http://dx.doi.org/10.1017/jfm.2024.405}.

\bibitem[Lin et~al.(2014)Lin, Maire, Belongie, Hays, Perona, Ramanan, Doll{\'a}r, and Zitnick]{10.1007/978-3-319-10602-1_48}
Lin, T.-Y., Maire, M., Belongie, S., Hays, J., Perona, P., Ramanan, D., Doll{\'a}r, P., and Zitnick, C.~L.
\newblock Microsoft coco: Common objects in context.
\newblock In Fleet, D., Pajdla, T., Schiele, B., and Tuytelaars, T. (eds.), \emph{Computer Vision -- ECCV 2014}, pp.\  740--755, Cham, 2014. Springer International Publishing.
\newblock ISBN 978-3-319-10602-1.
\newblock URL \url{https://link.springer.com/chapter/10.1007/978-3-319-10602-1_48}.

\bibitem[Liu et~al.(2019)Liu, Dinh, Smith, and Sun]{LIU20191096}
Liu, Y., Dinh, N.~T., Smith, R.~C., and Sun, X.
\newblock Uncertainty quantification of two-phase flow and boiling heat transfer simulations through a data-driven modular bayesian approach.
\newblock \emph{International Journal of Heat and Mass Transfer}, 138:\penalty0 1096--1116, 2019.
\newblock ISSN 0017-9310.
\newblock \doi{https://doi.org/10.1016/j.ijheatmasstransfer.2019.04.075}.
\newblock URL \url{https://www.sciencedirect.com/science/article/pii/S0017931019304922}.

\bibitem[Liu et~al.(2021)Liu, Lin, Cao, Hu, Wei, Zhang, Lin, and Guo]{9710580}
Liu, Z., Lin, Y., Cao, Y., Hu, H., Wei, Y., Zhang, Z., Lin, S., and Guo, B.
\newblock Swin transformer: Hierarchical vision transformer using shifted windows.
\newblock In \emph{2021 IEEE/CVF International Conference on Computer Vision (ICCV)}, pp.\  9992--10002, 2021.
\newblock \doi{10.1109/ICCV48922.2021.00986}.
\newblock URL \url{https://ieeexplore.ieee.org/document/9710580}.

\bibitem[Ma et~al.(2024)Ma, He, Li, Han, You, and Wang]{Ma2024}
Ma, J., He, Y., Li, F., Han, L., You, C., and Wang, B.
\newblock Segment anything in medical images.
\newblock \emph{Nature Communications}, 15\penalty0 (1):\penalty0 654, 2024.
\newblock ISSN 2041-1723.
\newblock \doi{10.1038/s41467-024-44824-z}.
\newblock URL \url{https://doi.org/10.1038/s41467-024-44824-z}.

\bibitem[Maalouf et~al.(2024)Maalouf, Jadhav, Jatavallabhula, Chahine, Vogt, Wood, Torralba, and Rus]{10436161}
Maalouf, A., Jadhav, N., Jatavallabhula, K.~M., Chahine, M., Vogt, D.~M., Wood, R.~J., Torralba, A., and Rus, D.
\newblock Follow anything: Open-set detection, tracking, and following in real-time.
\newblock \emph{IEEE Robotics and Automation Letters}, 9\penalty0 (4):\penalty0 3283--3290, 2024.
\newblock \doi{10.1109/LRA.2024.3366013}.
\newblock URL \url{https://ieeexplore.ieee.org/stamp/stamp.jsp?arnumber=10436161}.

\bibitem[Maduabuchi(2024)]{Maduabuchi2024}
Maduabuchi, C.
\newblock Automated segmentation and analysis of high-speed video phase-detection data for boiling heat transfer characterization using u-net convolutional neural networks and uncertainty quantification.
\newblock Master's thesis, Massachusetts Institute of Technology, 2024.
\newblock URL \url{https://dspace.mit.edu/handle/1721.1/155645}.
\newblock Accessed: 2024-09-15.

\bibitem[Maduabuchi et~al.(2024)Maduabuchi, Jossou, and Bucci]{maduabuchi2024videosamlargevisionfoundation}
Maduabuchi, C., Jossou, E., and Bucci, M.
\newblock Videosam: A large vision foundation model for high-speed video segmentation, 2024.
\newblock URL \url{https://arxiv.org/abs/2410.21304}.

\bibitem[Malakhov et~al.(2023)Malakhov, Seredkin, Chernyavskiy, Serdyukov, Mullyadzanov, and Surtaev]{MALAKHOV2023104402}
Malakhov, I., Seredkin, A., Chernyavskiy, A., Serdyukov, V., Mullyadzanov, R., and Surtaev, A.
\newblock Deep learning segmentation to analyze bubble dynamics and heat transfer during boiling at various pressures.
\newblock \emph{International Journal of Multiphase Flow}, 162:\penalty0 104402, 2023.
\newblock ISSN 0301-9322.
\newblock \doi{https://doi.org/10.1016/j.ijmultiphaseflow.2023.104402}.
\newblock URL \url{https://www.sciencedirect.com/science/article/pii/S0301932223000253}.

\bibitem[Mazurowski et~al.(2023)Mazurowski, Dong, Gu, Yang, Konz, and Zhang]{MAZUROWSKI2023102918}
Mazurowski, M.~A., Dong, H., Gu, H., Yang, J., Konz, N., and Zhang, Y.
\newblock Segment anything model for medical image analysis: An experimental study.
\newblock \emph{Medical Image Analysis}, 89:\penalty0 102918, 2023.
\newblock ISSN 1361-8415.
\newblock \doi{https://doi.org/10.1016/j.media.2023.102918}.
\newblock URL \url{https://www.sciencedirect.com/science/article/pii/S1361841523001780}.

\bibitem[Munia et~al.(2025)Munia, Abdar, Hasan, Jalali, Banerjee, Khosravi, Hossain, Fu, and Frangi]{MUNIA2025102719}
Munia, A.~A., Abdar, M., Hasan, M., Jalali, M.~S., Banerjee, B., Khosravi, A., Hossain, I., Fu, H., and Frangi, A.~F.
\newblock Attention-guided hierarchical fusion u-net for uncertainty-driven medical image segmentation.
\newblock \emph{Information Fusion}, 115:\penalty0 102719, 2025.
\newblock ISSN 1566-2535.
\newblock \doi{https://doi.org/10.1016/j.inffus.2024.102719}.
\newblock URL \url{https://www.sciencedirect.com/science/article/pii/S1566253524004974}.

\bibitem[Osco et~al.(2023)Osco, Wu, {de Lemos}, Gonçalves, Ramos, Li, and Marcato]{OSCO2023103540}
Osco, L.~P., Wu, Q., {de Lemos}, E.~L., Gonçalves, W.~N., Ramos, A. P.~M., Li, J., and Marcato, J.
\newblock The segment anything model (sam) for remote sensing applications: From zero to one shot.
\newblock \emph{International Journal of Applied Earth Observation and Geoinformation}, 124:\penalty0 103540, 2023.
\newblock ISSN 1569-8432.
\newblock \doi{https://doi.org/10.1016/j.jag.2023.103540}.
\newblock URL \url{https://www.sciencedirect.com/science/article/pii/S1569843223003643}.

\bibitem[Passoni et~al.(2024)Passoni, Mereu, and Bucci]{PASSONI2024104871}
Passoni, S., Mereu, R., and Bucci, M.
\newblock Integrating machine learning and image processing for void fraction estimation in two-phase flow through corrugated channels.
\newblock \emph{International Journal of Multiphase Flow}, 177:\penalty0 104871, 2024.
\newblock ISSN 0301-9322.
\newblock \doi{https://doi.org/10.1016/j.ijmultiphaseflow.2024.104871}.
\newblock URL \url{https://www.sciencedirect.com/science/article/pii/S0301932224001484}.

\bibitem[Paz et~al.(2017)Paz, Conde, Porteiro, and Concheiro]{s17061448}
Paz, C., Conde, M., Porteiro, J., and Concheiro, M.
\newblock On the application of image processing methods for bubble recognition to the study of subcooled flow boiling of water in rectangular channels.
\newblock \emph{Sensors}, 17\penalty0 (6), 2017.
\newblock ISSN 1424-8220.
\newblock \doi{10.3390/s17061448}.
\newblock URL \url{https://www.mdpi.com/1424-8220/17/6/1448}.

\bibitem[Portillo-Portillo et~al.(2022)Portillo-Portillo, Sanchez-Perez, Toscano-Medina, Hernandez-Suarez, Olivares-Mercado, Perez-Meana, Velarde-Alvarado, Orozco, and García~Villalba]{e24070942}
Portillo-Portillo, J., Sanchez-Perez, G., Toscano-Medina, L.~K., Hernandez-Suarez, A., Olivares-Mercado, J., Perez-Meana, H., Velarde-Alvarado, P., Orozco, A. L.~S., and García~Villalba, L.~J.
\newblock Fassvid: Fast and accurate semantic segmentation for video sequences.
\newblock \emph{Entropy}, 24\penalty0 (7), 2022.
\newblock ISSN 1099-4300.
\newblock \doi{10.3390/e24070942}.
\newblock URL \url{https://www.mdpi.com/1099-4300/24/7/942}.

\bibitem[Ravichandran et~al.(2023)Ravichandran, Kossolapov, Aguiar, Phillips, and Bucci]{RAVICHANDRAN2023110879}
Ravichandran, M., Kossolapov, A., Aguiar, G.~M., Phillips, B., and Bucci, M.
\newblock Autonomous and online detection of dry areas on a boiling surface using deep learning and infrared thermometry.
\newblock \emph{Experimental Thermal and Fluid Science}, 145:\penalty0 110879, 2023.
\newblock ISSN 0894-1777.
\newblock \doi{https://doi.org/10.1016/j.expthermflusci.2023.110879}.
\newblock URL \url{https://www.sciencedirect.com/science/article/pii/S0894177723000353}.

\bibitem[Richenderfer et~al.(2018)Richenderfer, Kossolapov, Seong, Saccone, Demarly, Kommajosyula, Baglietto, Buongiorno, and Bucci]{RICHENDERFER201835}
Richenderfer, A., Kossolapov, A., Seong, J.~H., Saccone, G., Demarly, E., Kommajosyula, R., Baglietto, E., Buongiorno, J., and Bucci, M.
\newblock Investigation of subcooled flow boiling and chf using high-resolution diagnostics.
\newblock \emph{Experimental Thermal and Fluid Science}, 99:\penalty0 35--58, 2018.
\newblock ISSN 0894-1777.
\newblock \doi{https://doi.org/10.1016/j.expthermflusci.2018.07.017}.
\newblock URL \url{https://www.sciencedirect.com/science/article/pii/S089417771831255X}.

\bibitem[Ronneberger et~al.(2015)Ronneberger, Fischer, and Brox]{10.1007/978-3-319-24574-4_28}
Ronneberger, O., Fischer, P., and Brox, T.
\newblock U-net: Convolutional networks for biomedical image segmentation.
\newblock In Navab, N., Hornegger, J., Wells, W.~M., and Frangi, A.~F. (eds.), \emph{Medical Image Computing and Computer-Assisted Intervention -- MICCAI 2015}, pp.\  234--241, Cham, 2015. Springer International Publishing.
\newblock ISBN 978-3-319-24574-4.
\newblock URL \url{https://link.springer.com/chapter/10.1007/978-3-319-24574-4_28}.

\bibitem[Schindelin et~al.(2012)Schindelin, Arganda-Carreras, Frise, Kaynig, Longair, Pietzsch, Preibisch, Rueden, Saalfeld, Schmid, Tinevez, White, Hartenstein, Eliceiri, Tomancak, and Cardona]{Schindelin2012}
Schindelin, J., Arganda-Carreras, I., Frise, E., Kaynig, V., Longair, M., Pietzsch, T., Preibisch, S., Rueden, C., Saalfeld, S., Schmid, B., Tinevez, J.-Y., White, D.~J., Hartenstein, V., Eliceiri, K., Tomancak, P., and Cardona, A.
\newblock Fiji: an open-source platform for biological-image analysis.
\newblock \emph{Nature Methods}, 9\penalty0 (7):\penalty0 676--682, 2012.
\newblock ISSN 1548-7105.
\newblock \doi{10.1038/nmeth.2019}.
\newblock URL \url{https://doi.org/10.1038/nmeth.2019}.

\bibitem[Selvan et~al.(2020)Selvan, Faye, Middleton, and Pai]{10.1007/978-3-030-59861-7_9}
Selvan, R., Faye, F., Middleton, J., and Pai, A.
\newblock Uncertainty quantification in medical image segmentation with normalizing flows.
\newblock In \emph{Machine Learning in Medical Imaging: 11th International Workshop, MLMI 2020, Held in Conjunction with MICCAI 2020, Lima, Peru, October 4, 2020, Proceedings}, pp.\  80–90, Berlin, Heidelberg, 2020. Springer-Verlag.
\newblock ISBN 978-3-030-59860-0.
\newblock \doi{10.1007/978-3-030-59861-7_9}.
\newblock URL \url{https://doi.org/10.1007/978-3-030-59861-7_9}.

\bibitem[Seong et~al.(2023)Seong, Ravichandran, Su, Phillips, and Bucci]{SEONG2023104336}
Seong, J.~H., Ravichandran, M., Su, G., Phillips, B., and Bucci, M.
\newblock Automated bubble analysis of high-speed subcooled flow boiling images using u-net transfer learning and global optical flow.
\newblock \emph{International Journal of Multiphase Flow}, 159:\penalty0 104336, 2023.
\newblock ISSN 0301-9322.
\newblock \doi{https://doi.org/10.1016/j.ijmultiphaseflow.2022.104336}.
\newblock URL \url{https://www.sciencedirect.com/science/article/pii/S0301932222002956}.

\bibitem[Singh et~al.(2009)Singh, Jain, Sridharan, Duttagupta, and Agrawal]{Singh_2009}
Singh, S.~G., Jain, A., Sridharan, A., Duttagupta, S.~P., and Agrawal, A.
\newblock Flow map and measurement of void fraction and heat transfer coefficient using an image analysis technique for flow boiling of water in a silicon microchannel.
\newblock \emph{Journal of Micromechanics and Microengineering}, 19\penalty0 (7):\penalty0 075004, jun 2009.
\newblock \doi{10.1088/0960-1317/19/7/075004}.
\newblock URL \url{https://dx.doi.org/10.1088/0960-1317/19/7/075004}.

\bibitem[Soibam et~al.(2023)Soibam, Scheiff, Aslanidou, Kyprianidis, and {Bel Fdhila}]{SOIBAM2023104589}
Soibam, J., Scheiff, V., Aslanidou, I., Kyprianidis, K., and {Bel Fdhila}, R.
\newblock Application of deep learning for segmentation of bubble dynamics in subcooled boiling.
\newblock \emph{International Journal of Multiphase Flow}, 169:\penalty0 104589, 2023.
\newblock ISSN 0301-9322.
\newblock \doi{https://doi.org/10.1016/j.ijmultiphaseflow.2023.104589}.
\newblock URL \url{https://www.sciencedirect.com/science/article/pii/S0301932223002094}.

\bibitem[Suh et~al.(2024)Suh, Chang, Simadiris, Inouye, Hoque, Khodakarami, Kharangate, Miljkovic, and Won]{SUH2024100309}
Suh, Y., Chang, S., Simadiris, P., Inouye, T.~B., Hoque, M.~J., Khodakarami, S., Kharangate, C., Miljkovic, N., and Won, Y.
\newblock Vision-it: A framework for digitizing bubbles and droplets.
\newblock \emph{Energy and AI}, 15:\penalty0 100309, 2024.
\newblock ISSN 2666-5468.
\newblock \doi{https://doi.org/10.1016/j.egyai.2023.100309}.
\newblock URL \url{https://www.sciencedirect.com/science/article/pii/S2666546823000812}.

\bibitem[Wahid et~al.(2024)Wahid, Kaffey, Farris, Humbert-Vidan, Moreno, Rasmussen, Ren, Naser, Netherton, Korreman, Balakrishnan, Fuller, Fuentes, and Dohopolski]{WAHID2024110542}
Wahid, K.~A., Kaffey, Z.~Y., Farris, D.~P., Humbert-Vidan, L., Moreno, A.~C., Rasmussen, M., Ren, J., Naser, M.~A., Netherton, T.~J., Korreman, S., Balakrishnan, G., Fuller, C.~D., Fuentes, D., and Dohopolski, M.~J.
\newblock Artificial intelligence uncertainty quantification in radiotherapy applications − a scoping review.
\newblock \emph{Radiotherapy and Oncology}, 201:\penalty0 110542, 2024.
\newblock ISSN 0167-8140.
\newblock \doi{https://doi.org/10.1016/j.radonc.2024.110542}.
\newblock URL \url{https://www.sciencedirect.com/science/article/pii/S0167814024035205}.

\bibitem[Wang et~al.(2023)Wang, Rahman, and Bucci]{10.1063/5.0135110}
Wang, C., Rahman, M.~M., and Bucci, M.
\newblock {Decrypting the mechanisms of wicking and evaporation heat transfer on micro-pillars during the pool boiling of water using high-resolution infrared thermometry}.
\newblock \emph{Physics of Fluids}, 35\penalty0 (3):\penalty0 037112, 03 2023.
\newblock ISSN 1070-6631.
\newblock \doi{10.1063/5.0135110}.
\newblock URL \url{https://doi.org/10.1063/5.0135110}.

\bibitem[Wang et~al.(2021)Wang, Sun, Cheng, Jiang, Deng, Zhao, Liu, Mu, Tan, Wang, Liu, and Xiao]{9052469}
Wang, J., Sun, K., Cheng, T., Jiang, B., Deng, C., Zhao, Y., Liu, D., Mu, Y., Tan, M., Wang, X., Liu, W., and Xiao, B.
\newblock Deep high-resolution representation learning for visual recognition.
\newblock \emph{IEEE Transactions on Pattern Analysis and Machine Intelligence}, 43\penalty0 (10):\penalty0 3349--3364, 2021.
\newblock \doi{10.1109/TPAMI.2020.2983686}.
\newblock URL \url{https://ieeexplore.ieee.org/document/9052469}.

\bibitem[Wenyin et~al.(2008)Wenyin, Ningde, Xia, and Zhiqiang]{4721870}
Wenyin, Z., Ningde, J., Xia, L., and Zhiqiang, N.
\newblock Bubble image segmentation of gas/liquid two-phase flow based on improved canny operator.
\newblock In \emph{2008 International Conference on Computer Science and Software Engineering}, volume~1, pp.\  799--801, 2008.
\newblock \doi{10.1109/CSSE.2008.1396}.
\newblock URL \url{https://ieeexplore.ieee.org/document/4721870}.

\bibitem[Yang et~al.(2023)Yang, Gao, Li, Gao, Wang, and Zheng]{yang2023track}
Yang, J., Gao, M., Li, Z., Gao, S., Wang, F., and Zheng, F.
\newblock Track anything: Segment anything meets videos, 2023.
\newblock URL \url{https://arxiv.org/abs/2304.11968}.

\bibitem[Zhang et~al.(2020)Zhang, Zhao, Lin, Tan, and Cheng]{Zhang2020}
Zhang, J., Zhao, J., Lin, H., Tan, Y., and Cheng, J.-X.
\newblock High-speed chemical imaging by dense-net learning of femtosecond stimulated raman scattering.
\newblock \emph{The Journal of Physical Chemistry Letters}, 11\penalty0 (20):\penalty0 8573--8578, 2020.
\newblock \doi{10.1021/acs.jpclett.0c01598}.
\newblock URL \url{https://doi.org/10.1021/acs.jpclett.0c01598}.

\bibitem[Zhang et~al.(2023)Zhang, Wang, Su, Kossolapov, Matana~Aguiar, Seong, Chavagnat, Phillips, Rahman, and Bucci]{Zhang2023}
Zhang, L., Wang, C., Su, G., Kossolapov, A., Matana~Aguiar, G., Seong, J.~H., Chavagnat, F., Phillips, B., Rahman, M.~M., and Bucci, M.
\newblock A unifying criterion of the boiling crisis.
\newblock \emph{Nature Communications}, 14\penalty0 (1):\penalty0 2321, 2023.
\newblock ISSN 2041-1723.
\newblock \doi{10.1038/s41467-023-37899-7}.
\newblock URL \url{https://doi.org/10.1038/s41467-023-37899-7}.

\bibitem[Zhou \& Niu(2020)Zhou and Niu]{ZHOU2020103277}
Zhou, H. and Niu, X.
\newblock An image processing algorithm for the measurement of multiphase bubbly flow using predictor-corrector method.
\newblock \emph{International Journal of Multiphase Flow}, 128:\penalty0 103277, 2020.
\newblock ISSN 0301-9322.
\newblock \doi{https://doi.org/10.1016/j.ijmultiphaseflow.2020.103277}.
\newblock URL \url{https://www.sciencedirect.com/science/article/pii/S0301932219309826}.

\bibitem[Zhou et~al.(2024)Zhou, Liang, Chen, Liu, Song, Vivone, and Chanussot]{10522788}
Zhou, X., Liang, F., Chen, L., Liu, H., Song, Q., Vivone, G., and Chanussot, J.
\newblock Mesam: Multiscale enhanced segment anything model for optical remote sensing images.
\newblock \emph{IEEE Transactions on Geoscience and Remote Sensing}, 62, 2024.
\newblock \doi{10.1109/TGRS.2024.3398038}.
\newblock URL \url{https://ieeexplore.ieee.org/document/10522788}.

\bibitem[Zou et~al.(2023)Zou, Yang, Zhang, Li, Li, Wang, Wang, Gao, and Lee]{NEURIPS2023_3ef61f7e}
Zou, X., Yang, J., Zhang, H., Li, F., Li, L., Wang, J., Wang, L., Gao, J., and Lee, Y.~J.
\newblock Segment everything everywhere all at once.
\newblock In Oh, A., Naumann, T., Globerson, A., Saenko, K., Hardt, M., and Levine, S. (eds.), \emph{Advances in Neural Information Processing Systems}, volume~36, pp.\  19769--19782. Curran Associates, Inc., 2023.
\newblock URL \url{https://proceedings.neurips.cc/paper_files/paper/2023/file/3ef61f7e4afacf9a2c5b71c726172b86-Paper-Conference.pdf}.

\end{thebibliography}
\end{document}